\documentclass{article}

\newif\ifbook
\bookfalse

\usepackage[utf8]{inputenc}
\usepackage{stmaryrd}
\usepackage{amsmath,amsthm, amssymb, amsfonts}
\usepackage{fullpage}
\usepackage{footmisc}
\usepackage[figurewithin = section]{caption}
\usepackage{paralist}
\usepackage[ruled,vlined]{algorithm2e}
\usepackage{wasysym}
\usepackage{verbatim}
\usepackage{authblk}
\usepackage{ltablex}
\usepackage{xcolor}
\usepackage{shadethm}
\usepackage[pdfencoding=auto]{hyperref}
\usepackage{graphicx}
\usepackage{enumitem}
\usepackage{dsfont}
\usepackage{microtype}
\usepackage{bbm}
\usepackage{csquotes}
\usepackage[export]{adjustbox}

\usepackage{tikz}
\usetikzlibrary{calc,shapes,matrix,chains,positioning,decorations.pathreplacing,arrows}

\usepackage{mathtools}
\mathtoolsset{showonlyrefs}

\usepackage{wrapfig}
\definecolor{darkred}{rgb}{0.55, 0.0, 0.0}

\newcommand{\gr}{{\ensuremath{{G}}}}
\newcommand{\arch}{{\ensuremath{{a}}}}
\newcommand{\E}{{\ensuremath{\mathbbm{E}}}}
\newcommand{\V}{{\ensuremath{\mathbbm{V}}}}
\newcommand{\cS}{{\ensuremath{S}}}
\newcommand{\cM}{{\ensuremath{\mathcal{M}}}}
\newcommand{\cL}{{\ensuremath{\mathcal{L}}}}
\newcommand{\cX}{{\ensuremath{\mathcal{X}}}}
\newcommand{\cY}{{\ensuremath{\mathcal{Y}}}}
\newcommand{\cZ}{{\ensuremath{\mathcal{Z}}}}
\newcommand{\cR}{{\ensuremath{\mathcal{R}}}}
\newcommand{\cF}{{\ensuremath{\mathcal{F}}}}
\newcommand{\cG}{{\ensuremath{\mathcal{G}}}}
\newcommand{\cO}{{\ensuremath{\mathcal{O}}}}
\newcommand{\cP}{{\ensuremath{\mathcal{P}}}}
\newcommand{\cA}{{\ensuremath{\mathcal{A}}}}
\newcommand{\cU}{{\ensuremath{\mathcal{U}}}}
\newcommand{\cN}{{\ensuremath{\mathcal{N}}}}
\newcommand{\cs}{{\ensuremath{s}}}
\newcommand{\fR}{{\ensuremath{\mathfrak{R}}}}
\newcommand{\sgn}{{\ensuremath{\operatorname{sgn}}}}
\newcommand{\trace}{{\ensuremath{\operatorname{Tr}}}}
\newcommand{\prisk}{{\ensuremath{r}}}
\newcommand{\ReLU}{{\ensuremath{\varrho_{R}}}}
\newcommand{\Z}{{\ensuremath{\mathbb{Z}}}}
\newcommand{\N}{\ensuremath{{\mathbb{N}}}}
\newcommand{\R}{{\ensuremath{\mathbb{R}}}}
\newcommand{\eps}{{\ensuremath{\varepsilon}}}
\newcommand{\risk}{\ensuremath{{\mathcal{R}}}}
\newcommand{\erisk}[1]{\ensuremath{\widehat{\mathcal{R}}_{#1}}}
\newcommand{\batch}{{\ensuremath{b}}}

\renewcommand{\P}{{\ensuremath{\mathbbm{P}}}}

\DeclareMathOperator*{\supp}{supp}
\DeclareMathOperator{\spann}{span}
\DeclareMathOperator*{\argmin}{arg\,min}
\DeclareMathOperator*{\limp}{lim\vphantom{p}}

\newshadetheorem{theorem}{Theorem}[section]
\newshadetheorem{lemma}{Lemma}[section]
\newshadetheorem{proposition}{Proposition}[section]

\newtheorem{remark}[theorem]{Remark}
\newtheorem{definition}[theorem]{Definition}
\newtheorem{assumption}[theorem]{Assumption}
\newtheorem{example}[theorem]{Example}

\numberwithin{equation}{section}

\title{The Modern Mathematics of Deep Learning%
\ifbook%
\else%
\thanks{A version of this review paper appears as a chapter in the book \enquote{Mathematical Aspects of Deep Learning} by Cambridge University Press.}
\fi%
}
\author{Julius Berner\thanks{Faculty of Mathematics, University of Vienna.}\qquad Philipp Grohs\thanks{Faculty of Mathematics and Research Network DataScience@UniVienna, University of Vienna.}\qquad Gitta Kutyniok\thanks{Department of Mathematics, Ludwig Maximilian University of Munich, and Department of Physics and Technology, University of Tromsø.}\qquad Philipp Petersen%
\ifbook%
$^\dag$
\else%
$^\ddag$
\fi%
}
\date{}

\begin{document}

\maketitle

\begin{abstract}
We describe the new field of mathematical analysis of deep learning. This field emerged around a list of research questions that were not answered within the classical framework of learning theory. These questions concern: the outstanding generalization power of overparametrized neural networks, the role of depth in deep architectures, the apparent absence of the curse of dimensionality, the surprisingly successful optimization performance despite the non-convexity of the problem, understanding what features are learned, why deep architectures perform exceptionally well in physical problems, and which fine aspects of an architecture affect the behavior of a learning task in which way. We present an overview of modern approaches that yield partial answers to these questions. For selected approaches, we describe the main ideas in more detail.
\end{abstract}

\setcounter{tocdepth}{2}
\tableofcontents

\section{Introduction}
\label{sec:intro}
Deep learning has undoubtedly established itself as the outstanding machine learning technique of recent times. This dominant position was claimed through a series of overwhelming successes in widely different application areas. 

Perhaps the most famous application of deep learning and certainly one of the first where these techniques became state-of-the-art is image classification~\cite{lecun1998gradient, krizhevsky2012imagenet, szegedy2015going, he2016deep}. In this area, deep learning is nowadays the only method that is seriously considered. The prowess of deep learning classifiers goes so far that they often outperform humans in image labelling tasks~\cite{he2015delving}.

A second famous application area is the training of deep-learning-based agents to play board games or computer games, such as Atari games~\cite{mnih2013playing}. In this context, probably the most prominent achievement yet is the development of an algorithm that beat the best human player in the game of Go~\cite{silver2016mastering, silver2017mastering}---a feat that was previously unthinkable owing to the extreme complexity of this game. Besides, even in multiplayer, team-based games with incomplete information deep-learning-based agents nowadays outperform world-class human teams~\cite{berner2019dota, vinyals2019grandmaster}.

In addition to playing games, deep learning has also led to impressive breakthroughs in the natural sciences. For example, it is used in the development of drugs~\cite{ma2015deep}, molecular dynamics~\cite{faber2017prediction}, or in high-energy physics~\cite{baldi2014searching}. One of the most astounding recent breakthroughs in scientific applications is the development of a deep-learning-based predictor for the folding behavior of proteins~\cite{senior2020improved}. This predictor is the first method to match the accuracy of lab-based methods. 

Finally, in the vast field of natural language processing, which includes the subtasks of understanding, summarizing, or generating text, 
impressive advances were made based on deep learning. Here, we refer to~\cite{young2018recent} for an overview. One technique that recently stood out is based on a so-called transformer neural network~\cite{bahdanau2014neural, vaswani2017attention}. This network structure gave rise to the impressive GPT-3 model~\cite{brown2020language} which not only creates coherent and compelling texts but can also produce code, such as, for the layout of a webpage according to some instructions that a user inputs in plain English. Transformer neural networks have also been successfully employed in the field of symbolic mathematics~\cite{saxton2018analysing,lample2019deep}.

In this
\ifbook
\else%
article,
\fi%
we present and discuss the mathematical foundations of the success story outlined above. More precisely, our goal is to outline the newly emerging field of \emph{mathematical analysis of deep learning}. To accurately describe this field, a necessary preparatory step is to sharpen our definition of the term deep learning. For the purposes of this 
\ifbook
\else%
article,
\fi%
we will use the term in the following narrow sense: \emph{Deep learning refers to techniques where deep neural networks\footnote{We will define the term \emph{neural network} later but, for this definition,
one can view it as a parametrized family of functions with a differentiable parametrization.} are trained with gradient-based methods}. This narrow definition is helpful to make this 
\ifbook
\else%
article
\fi%
more concise. We would like to stress, however, that we do not claim in any way that this is the \emph{best} or the \emph{right} definition of deep learning.

Having fixed a definition of deep learning, three questions arise concerning the aforementioned emerging field of mathematical analysis of deep learning: To what extent is a mathematical theory necessary? Is it truly a new field? What are the questions studied in this area?

Let us start by explaining the necessity of a theoretical analysis of the tools described above. From a scientific perspective, the primary reason why deep learning should be studied mathematically is simple curiosity. As we will see throughout this
\ifbook
\else%
article,
\fi%
many practically observed phenomena in this context are not explained theoretically. Moreover, theoretical insights and the development of a comprehensive theory are often the driving force underlying the development of new and improved methods. Prominent examples of mathematical theories with such an effect are the theory of fluid mechanics which is an invaluable asset to the design of aircraft or cars and the theory of information that affects and shapes all modern digital communication. In the words of Vladimir Vapnik\footnote{This claim can be found earlier in a non-mathematical context in the works of Kurt Lewin~\cite{lewin1943psychology}.}: \enquote{Nothing is more practical than a good theory},~\cite[Preface]{vapnik2013nature}. In addition to being interesting and practical, theoretical insight may also be necessary. Indeed, in many applications of machine learning, such as medical diagnosis, self-driving cars, and robotics, a significant level of control and predictability of deep learning methods is mandatory. Also, in services, such as banking or insurance, the technology should be controllable to guarantee fair and explainable decisions.

Let us next address the claim that the field of mathematical analysis of deep learning is a newly emerging area. In fact, under the aforementioned definition of deep learning, there are two main ingredients of the technology: deep neural networks and gradient-based optimization. The first artificial neuron was already introduced in 1943 in~\cite{mcculloch1943logical}.
This neuron was not trained but instead used to explain a biological neuron. The first multi-layered network of such artificial neurons that was also trained can be found in~\cite{rosenblatt1958perceptron}. Since then, various neural network architectures have been developed. We will discuss these architectures in detail in the following sections. The second ingredient, gradient-based optimization, is made possible by the observation that due to the graph-based structure of neural networks the gradient of an objective function with respect to the parameters
of the neural network can be computed efficiently. This has been observed in various ways, see~\cite{kelley1960gradient, dreyfus1962numerical, linnainmaa1970representation, rumelhart1986learning}. Again, these techniques will be discussed in the upcoming sections. 
Since then, techniques have been improved and extended. As the rest of the
\ifbook
\else%
manuscript
\fi%
is spent reviewing these methods, we will keep the discussion of literature at this point brief. Instead, we refer to some overviews of the history of deep learning from various perspectives:~\cite{lecun2015deep, schmidhuber2015deep, goodfellow2016deep, higham2019deep}.

Given the fact that the two main ingredients of deep neural networks have been around for a long time, one would expect that a comprehensive mathematical theory has been developed that describes why and when deep-learning-based methods will perform well or when they will fail. 
Statistical learning theory~\cite{anthony1999neural, vapnik1999overview, cucker2002mathematical, bousquet2003introduction, vapnik2013nature} describes multiple aspects of the performance of general learning methods and in particular deep learning. We will review this theory in the context of deep learning in Subsection~\ref{subsec:foundations_learning} below. Hereby, we focus on classical, deep learning-related results that we consider well-known in the machine learning community. Nonetheless, the choice of these results is guaranteed to be subjective. We will find that the presented, classical theory is too general to explain the performance of deep learning adequately. In this context, we will identify the following questions that appear to be difficult to answer within the classical framework of learning theory:
\emph{Why do trained deep neural networks not overfit on the training data despite the enormous power of the architecture? What is the advantage of deep compared to shallow architectures? Why do these methods seemingly not suffer from the curse of dimensionality? Why does the optimization routine often succeed in finding good solutions despite the non-convexity, non-linearity, and often non-smoothness of the problem? Which aspects of an architecture affect the performance of the associated models and how? Which features of data are learned by deep architectures? Why do these methods perform as well as or better than specialized numerical tools in natural sciences?}

The new field of mathematical analysis of deep learning has emerged around questions like the ones listed above. In the remainder of this 
\ifbook
\else%
article,
\fi%
we will collect some of the main recent advances to answer these questions. Because this field of mathematical analysis of deep learning is incredibly active and new material is added at breathtaking speeds, a brief survey on recent advances in this area is guaranteed to miss not only a couple of references but also many of the most essential ones. Therefore, we do not strive for a complete overview, but instead, showcase several fundamental ideas on a mostly intuitive level. In this way, we hope to allow the reader to familiarize themselves with some exciting concepts and provide a convenient entry-point for further studies.
\subsection{Notation}
\label{subsec:notation}
We denote by $\N$ the set of natural numbers, by $\Z$ the set of integers and by $\R$ the field of real numbers. For $N \in \N$, we denote by $[N]$ the set $\{1, \dots, N\}$. For two functions $f,g\colon \mathcal{X} \to [0,\infty)$, we write $f \lesssim g$, if there exists a universal constant $c$ such that $f(x) \le c g(x)$ for all $x \in \mathcal{X}$. 
In a pseudometric space $(\cX,d_\cX)$, we define the ball of radius $r \in (0,\infty)$ around a point $x\in \cX$ by $B^{d_\cX}_{r}(x)$ or $B_{r}(x)$ if the pseudometric $d_\cX$ is clear from the context.
By $\|\cdot\|_p$, 
$p\in[1,\infty]$, we denote the $\ell^p$-norm, and by 
$\langle \cdot, \cdot \rangle$
the Euclidean inner product of given vectors.
By 
$\|\cdot\|_{\mathrm{op}}$ we denote
the operator norm induced by the Euclidean norm and by $\|\cdot\|_F$
the Frobenius norm of given matrices.
For $p \in [1, \infty]$, $s\in [0,\infty)$, $d\in\N$, and $\cX \subset \R^d$, we denote by $W^{s,p}(\cX)$ the Sobolev-Slobodeckij space, which for $s=0$ is just a Lebesgue space, i.e., $W^{0,p}(\cX)=L^p(\cX)$. For measurable spaces $\cX$ and $\cY$, we define $\cM(\cX,\cY)$ to be the set of measurable functions from $\cX$ to $\cY$. 
We denote by $\hat{g}$ the Fourier transform\footnote{Respecting common notation, we will also use the hat symbol to denote the minimizer of the empirical risk $\widehat{f}_\cs$ in Definition~\ref{def:ERM} but this clash of notation does not cause any ambiguity.} of a tempered distribution $g$. 
For probabilistic statements, we will assume a suitable underlying probability space with probability measure $\P$. For an $\cX$-valued random variable $X$, we denote by $\E[X]$ and $\V[X]$ its expectation and variance and by $\P_X$ the image measure of $X$ on $\cX$, i.e., $\P_X(A) = \P(X \in A)$ for every measurable set $A\subset \cX$. If possible, we use the corresponding lowercase letter to denote the realization $x\in\cX$ of the random variable $X$ for a given outcome. 
We write $\mathrm{I}_d$ for the $d$-dimensional identity matrix and, for a set $A$, we write $\mathds{1}_A$ for the indicator function of $A$, i.e., $\mathds{1}_A(x) = 1$ if $x \in A$ and $\mathds{1}_A(x) = 0$ else.

\subsection{Foundations of learning theory}
\label{subsec:foundations_learning}
Before we continue to describe recent developments in the mathematical analysis of deep learning methods, we start by providing a concise overview of the classical mathematical and statistical theory underlying machine learning tasks and algorithms which, in their most general form,
can be formulated as follows.
\begin{definition}[Learning - informal]
\label{def:learning}
Let $\cX,\cY$, and $\cZ$ be measurable spaces. In a learning task, one is given data in $\cZ$ and a loss function $\cL\colon \cM(\cX,\cY)\times\cZ\to \R$. The goal is to choose a hypothesis set $\cF\subset \cM(\cX,\cY)$ and construct a learning algorithm, i.e., a mapping
\begin{equation*}
    \cA\colon \bigcup_{m\in\N} \cZ^m \to \cF,
\end{equation*}
that uses training data $\cs=(z^{(i)})_{i=1}^m\in \cZ^m$ to find a model $f_\cs=\mathcal{A}(\cs)\in\cF$ that performs well on the training data
$\cs$ and also generalizes to unseen data $z\in\cZ$. Here, performance is measured via the loss function $\cL$ and the corresponding loss $\cL(f_\cs,z)$ and, informally speaking, generalization means that the out-of-sample performance of $f_\cs$ at $z$ behaves similar to the in-sample performance on $\cs$.
\end{definition}
Definition~\ref{def:learning} is deliberately vague on how to measure generalization performance. Later, we will often study the \emph{expected} out-of-sample performance. To talk about expected performance, a data distribution needs to be specified. We will revisit this point in Assumption~\ref{ass:iid} and Definition~\ref{def:risk}.

For simplicity, we focus on one-dimensional, supervised prediction tasks with input features in Euclidean space as defined in the following.
\begin{definition}[Prediction task]
\label{def:pred_task}
In a prediction task, we have that $\cZ\coloneqq\cX\times \cY$, i.e., we are given training data $\cs= ((x^{(i)},y^{(i)}))_{i=1}^m$ that consist of input features $x^{(i)} \in\cX$ and corresponding labels $y^{(i)} \in\cY$. For one-dimensional regression tasks with $\cY\subset\R$, we consider the quadratic loss $\cL(f,(x,y))=(f(x)-y)^2$
and, for binary classification tasks with $\cY=\{-1,1\}$, we consider the $0$-$1$ loss $\cL(f,(x,y))=\mathds{1}_{(-\infty,0)}(y f(x))$. We assume that our input features are in Euclidean space, i.e., $\cX\subset\R^d$ with input dimension $d\in\N$.
\end{definition}
In a prediction task, we aim for a model $f_\cs\colon\cX\to\cY$, such that, for unseen pairs $(x,y)\in\cX\times \cY$, $f_\cs(x)$ is a good prediction of the true label $y$. However, note that large parts of the presented theory can be applied to more general settings.
\begin{remark}[Learning tasks]
\label{rem:further_settings}
Apart from straightforward extensions to multi-dimensional prediction tasks and other loss functions, we want to mention that unsupervised and semi-supervised learning tasks are often treated as prediction tasks. More precisely, one transforms unlabeled training data $z^{(i)}$ into features $x^{(i)}=T_1(z^{(i)})\in\cX$ and labels $y^{(i)}=T_2(z^{(i)})\in\cY$ using suitable transformations $T_1\colon\cZ\to\cX$, $T_2\colon\cZ\to\cY$.
In doing so, one asks for a model $f_\cs$ approximating the transformation $T_2\circ T_1^{-1}\colon\cX\to\cY$ which is, e.g., done in order to learn feature representations or invariances.

Furthermore, one can consider density estimation tasks, where $\cX=\cZ$, $\cY\coloneqq[0,\infty]$, and $\cF$ consists of probability densities with respect to some $\sigma$-finite reference measure $\mu$ on $\cZ$. One then aims for a probability density $f_\cs$ that approximates the density of the unseen data $z$ with respect to $\mu$. One can perform $L^2(\mu)$-approximation based on the discretization $\cL(f, z)=-2f(z)+\|f\|^2_{L^2(\mu)}$ or maximum likelihood estimation based on the surprisal $\cL(f, z)=-\log(f(z))$.
\end{remark}

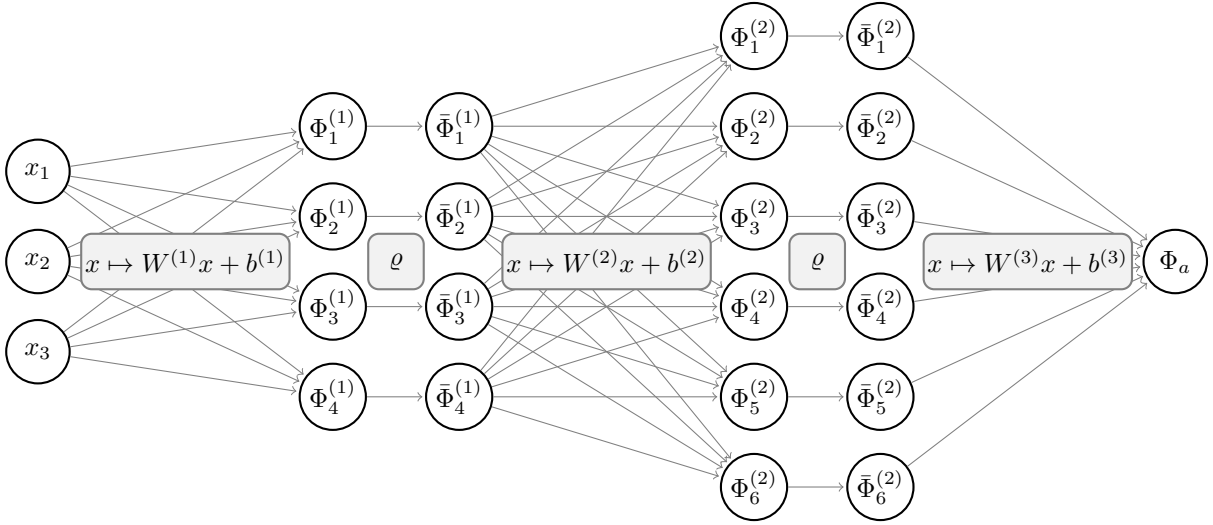
\begin{figure}[t]
\centering
 \def\layersep{4.9}
 \def\actsep{2.1}
\begin{tikzpicture}[shorten >=1pt,->,draw=black!50, node distance=\layersep, scale=0.4]
    \tikzstyle{every pin edge}=[<-,shorten <=1pt]
    \tikzstyle{neuron}=[circle,draw=black,thick,minimum size=24pt,inner sep=1pt]
    \tikzstyle{box}=[rectangle,rounded corners,fill=black!5,draw=black!50,thick,minimum size=21pt,inner sep=1.5pt]
    \tikzstyle{input neuron}=[neuron];
    \tikzstyle{output neuron}=[neuron];
    \tikzstyle{hidden neuron}=[neuron];

    \foreach \name / \y in {1,2,3}
        \node[input neuron] (I-\name) at (-\layersep,-3*\y+6) {$x_{\y}$};
    
    \foreach \name / \y in {1,...,4}
        \path
            node[hidden neuron] (H-\name) at (\layersep,-3*\y+7.5) {$\Phi^{(1)}_{\y}$};
    
    \foreach \source in {1,2,3}
        \foreach \dest in {1,...,4}
            \path (I-\source) edge (H-\dest);

    \node[box] (G-1) at (0,0) {$x\mapsto W^{(1)}x+b^{(1)}$};
    
    \foreach \name / \y in {1,...,4}
        \path
            node[hidden neuron] (Ha-\name) at (\layersep+2*\actsep,-3*\y+7.5) {$\bar{\Phi}^{(1)}_{\y}$};

    \foreach \source in {1,2,3,4}
        \path (H-\source) edge (Ha-\source);
        
    \node[box] (G-2) at (\layersep+\actsep,0) {$\varrho$};
    
    \foreach \name / \y in {1,...,6}
        \path
            node[hidden neuron] (L-\name) at (3*\layersep+2*\actsep,-3*\y+10.5) {$\Phi^{(2)}_{\y}$};
    
    \foreach \source in {1,...,4}
        \foreach \dest in {1,...,6}
            \path (Ha-\source) edge (L-\dest);

    \node[box] (G-3) at (2*\layersep+2*\actsep,0) {$x\mapsto W^{(2)}x+b^{(2)}$};
        
    \foreach \name / \y in {1,...,6}
        \path
            node[hidden neuron] (La-\name) at (3*\layersep+4*\actsep,-3*\y+10.5) {$\bar{\Phi}^{(2)}_{\y}$};

    \foreach \source in {1,...,6}
        \path (L-\source) edge (La-\source);
            
    \node[box] (G-4) at (3*\layersep+3*\actsep,0) {$\varrho$};
        
    \node[output neuron] (out) at (5*\layersep+4*\actsep,0){$\Phi_\arch$};

    \foreach \source in {1,...,6}
        \path (La-\source) edge (out);

    \node[box] (G-5) at (4*\layersep+4*\actsep,0) {$x\mapsto W^{(3)}x+b^{(3)}$};
\end{tikzpicture}
\caption{Graph (grey) and (pre-)activations of the neurons (white) of a deep fully connected feedforward neural network $\Phi_\arch\colon\R^3\times \R^{53}\mapsto\R$ with architecture $\arch=((3,4,6,1),\varrho)$ and parameters $\theta=((W^{(\ell)}, b^{(\ell)})_{\ell=1}^3$.} 
\label{fig:neural_net} 
\end{figure}

In deep learning the hypothesis set $\cF$ consists of \emph{realizations of neural networks} $\Phi_\arch(\cdot, \theta)$, $\theta \in \cP$, with a given \emph{architecture} $\arch$ and \emph{parameter set} $\cP$.
In practice, one uses the term neural network for a range of functions that can be represented by directed acyclic graphs, where the vertices correspond to elementary almost everywhere differentiable functions parametrizable by $\theta\in\cP$ and the edges symbolize compositions of these functions. In Section~\ref{sec:architectures}, we will review some frequently used architectures, in the other sections, however, we will mostly focus on \emph{fully connected feedforward} (FC) neural networks as defined below.
\begin{definition}[FC neural network]
\label{def:classical_nns}
A fully connected feedforward neural network is given by its architecture $\arch=(N,\varrho)$, where $L\in\N$, $N\in \N^{L+1}$, and $\varrho\colon\R\to\R$. 
We refer to $\varrho$ as the activation function, to $L$ as the number of layers, and to $N_0$, $N_L$, and $N_\ell$, $\ell \in [L-1]$, as the number of neurons in the input, output, and $\ell$-th hidden layer, respectively.
We denote the number of parameters by 
\begin{equation*}
    P(N)\coloneqq\sum_{\ell=1}^L N_{\ell}N_{\ell-1} +N_{\ell}
\end{equation*}
and define the corresponding realization function $\Phi_{\arch}\colon \R^{N_0} \times \R^{P(N)}\to \R^{N_L}$ which satisfies for every input $x\in\R^{N_0}$ and parameters
\begin{equation*}
    \theta = (\theta^{(\ell)})_{\ell=1}^L=((W^{(\ell)},b^{(\ell)}))_{\ell=1}^L \in \bigtimes_{\ell=1}^L (\R^{N_{\ell} \times N_{\ell-1} } \times \R^{N_{\ell}}) \cong \R^{P(N)}
\end{equation*}
that $\Phi_\arch(x,\theta)=\Phi^{(L)}(x,\theta)$, where
\begin{equation}
\begin{split}
    \Phi^{(1)}(x,\theta)&= W^{(1)}x+b^{(1)}, \\ 
    \bar{\Phi}^{(\ell)}(x,\theta)&= \varrho \big( \Phi^{(\ell)}(x,\theta)\big), \quad \ell\in [L-1], \quad  \text{and} \label{eq:def_nn_classical}\\
    \Phi^{(\ell+1)}(x,\theta) &= W^{(\ell+1)} \bar{\Phi}^{(\ell)}(x,\theta)+b^{(\ell+1)},  \quad \ell\in [L-1],
\end{split}
\end{equation}
and $\varrho$ is applied componentwise.
We refer to $W^{(\ell)}\in\R^{N_{\ell} \times N_{\ell-1}}$ and $b^{(\ell)}\in\R^{N_\ell}$ as the weight matrices and bias vectors, and to $\bar{\Phi}^{(\ell)}$ and $\Phi^{(\ell)}$ as the activations and pre-activations of the $N_\ell$ neurons in the $\ell$-th layer. 
The width and depth of the architecture are given by $\| N\|_\infty$ and $L$ and we call the architecture deep if $L> 2$ and shallow if $L=2$.
\end{definition}
The underlying directed acyclic graph of FC networks is given by compositions of the affine linear maps $x\mapsto W^{(\ell)}x +b^{(\ell)}$, $\ell\in[L]$, with the activation function $\varrho$ intertwined, see Figure~\ref{fig:neural_net}.
Typical activation functions used in practice are variants of the \emph{rectified linear unit} (ReLU) given by $\ReLU(x)\coloneqq\max\{0,x\}$ and \emph{sigmoidal functions} $\varrho\in C(\R)$ satisfying $\varrho(x) \to 1$ for $x \to \infty$ and $\varrho(x) \to 0$ for $x \to -\infty$, such as the logistic function $\varrho_\sigma(x)\coloneqq 1/(1+e^{-x})$ (often referred to as \emph{the} sigmoid function). See also Table~\ref{tbl:activationfunctions} for a comprehensive list of widely used activation functions. 

\renewcommand{\arraystretch}{1.6}
\begin{table}
\centering
\begin{tabular}{|l|c|c|}
\hline
  \textbf{Name}
& \textbf{Given as a function of $x\in\R$ by}
& \textbf{Plot}
\\ \hline
linear
& $x$
& \raisebox{-.42\height}{\includegraphics[width = 2.5cm]{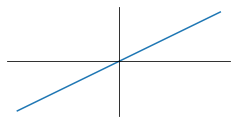}}
\\ \hline
Heaviside / step function
& $\mathds{1}_{(0, \infty)}(x)$
& \raisebox{-.42\height}{\includegraphics[width = 2.5cm]{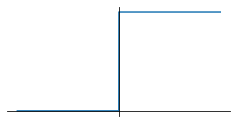}}
\\ \hline
logistic / sigmoid
& $\frac{1}{1+e^{-x}}$
& \raisebox{-.42\height}{\includegraphics[width = 2.5cm]{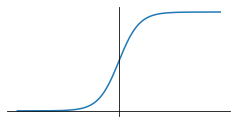}}
\\ \hline
  rectified linear unit (ReLU)
& $\max\{0,x\}$
& \raisebox{-.42\height}{\includegraphics[width = 2.5cm]{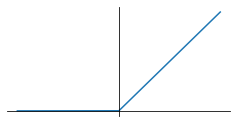}}
\\ \hline
  power rectified linear unit
& $\max\{0,x\}^k\ $ for $k\in\N$
& \raisebox{-.42\height}{\includegraphics[width = 2.5cm]{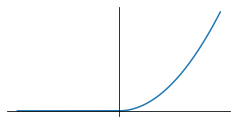}}
\\ \hline
  parametric ReLU (PReLU)
& $\max\{ax,x\}\ $ for $a \geq 0$, $a \neq 1$
& \raisebox{-.42\height}{\includegraphics[width = 2.5cm]{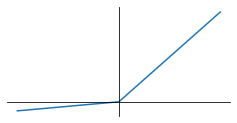}}
\\ \hline
  exponential linear unit (ELU)
& $x\cdot \mathds{1}_{[0,\infty)}(x) + (e^x-1)\cdot \mathds{1}_{(-\infty,0)}(x)$
& \raisebox{-.42\height}{\includegraphics[width = 2.5cm]{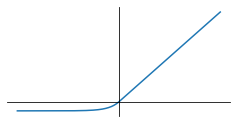}}
\\ \hline
  softsign
& $\frac{x}{1+|x|}$
& \raisebox{-.42\height}{\includegraphics[width = 2.5cm]{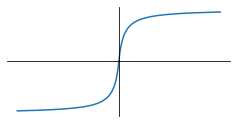}}
\\ \hline
  inverse square root linear unit \hspace{0.5em}  
&  \hspace{0.5em}  $x\cdot \mathds{1}_{[0,\infty)}(x) + \frac{x}{\sqrt{1+ax^2}}\cdot \mathds{1}_{(-\infty,0)}(x)\ $ for $a>0 $ \hspace{0.5em} 
 & \raisebox{-.42\height}{\includegraphics[width = 2.5cm]{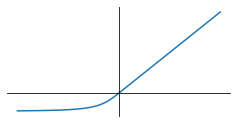}}
\\ \hline
  inverse square root unit
& $\frac{x}{\sqrt{1 + a x^2}}\ $ for $a>0$
& \raisebox{-.42\height}{\includegraphics[width = 2.5cm]{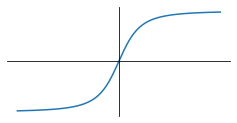}}
\\ \hline
  tanh
& $\frac{e^x-e^{-x}}{e^x+e^{-x}}$
& \raisebox{-.42\height}{\includegraphics[width = 2.5cm]{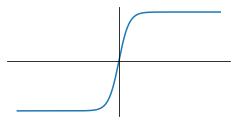}}
\\ \hline
  arctan
& $\arctan(x)$
& \raisebox{-.42\height}{\includegraphics[width = 2.5cm]{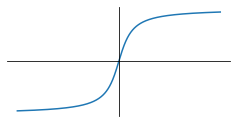}}
\\ \hline
  softplus
& $\ln(1+e^x)$
& \raisebox{-.42\height}{\includegraphics[width = 2.5cm]{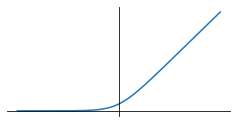}}
\\ \hline
  Gaussian
& $e^{-x^2/2}$
& \raisebox{-.42\height}{\includegraphics[width = 2.5cm]{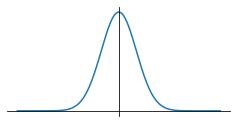}}
\\ \hline
\end{tabular}
\caption{List of commonly used activation functions.}
\label{tbl:activationfunctions}
\end{table}
\renewcommand{\arraystretch}{1}

\begin{remark}[Neural networks]
\label{rem:nn_notation}
If not further specified, we will use the term (neural) network, or the abbreviation NN, to refer to FC neural networks. 
Note that many of the architectures used in practice (see Section~\ref{sec:architectures}) can be written as special cases of Definition~\ref{def:classical_nns} where, e.g., specific parameters are prescribed by constants or shared with other parameters. Furthermore, note that affine linear functions are NNs with depth $L=1$. We will also consider biasless NNs given by linear mappings without bias vector, i.e., $b^{(\ell)}=0$, $\ell \in [L]$. In particular, any NN can always be written without bias vectors by redefining
\begin{equation*}
    x \to \begin{bmatrix}x \\  1 \end{bmatrix}, \quad (W^{(\ell)}, b^{(\ell)}) \to \begin{bmatrix} W^{(\ell)} & b^{(\ell)} \\ 0 & 1 \end{bmatrix}, \quad \ell\in[L-1], \quad \text{and} \quad (W^{(L)}, b^{(L)}) \to \begin{bmatrix} W^{(L)} & b^{(L)} \end{bmatrix}.
\end{equation*}
To enhance readability we will often not specify the underlying architecture $\arch=(N,\varrho)$ or the parameters $\theta\in\R^{P(N)}$ and use the term NN to refer to the architecture as well as the realization functions $\Phi_\arch({{\cdot}},\theta)\colon\R^{N_0} \to \R^{N_L}$ or $\Phi_\arch\colon\R^{N_0} \times \R^{P(N)} \to \R^{N_L}$.
However, we want to emphasize that one cannot infer the underlying architecture or properties like magnitude of parameters solely from these functions as the mapping $(\arch,\theta) \mapsto \Phi_\arch({{\cdot}},\theta)$ is highly non-injective. As an example, we can set $W^{(L)}=0$ which implies $\Phi_\arch({{\cdot}},\theta)=b^{(L)}$ for all architectures $\arch=(N,\varrho)$ and all values of $(W^{(\ell)},b^{(\ell)})_{\ell=1}^{L-1}$. 
\end{remark}
In view of our considered prediction tasks in Definition~\ref{def:pred_task}, this naturally leads to the following hypothesis sets of neural networks.
\begin{definition}[Hypothesis sets of neural networks]
\label{def:nn_hypothesis}
Let $\arch=(N,\varrho)$ be a NN architecture with input dimension $N_0=d$, output dimension $N_L=1$, and measurable activation function $\varrho$.
For regression tasks the corresponding hypothesis set is given by
\begin{equation*}
    \cF_{\arch}=\big\{
    \Phi_\arch(\cdot,\theta)
    \colon \theta\in\R^{P(N)}\big\}
\end{equation*}
and for classification tasks by
\begin{equation*}
    \cF_{\arch, \sgn} =\big\{\sgn(\Phi_\arch(\cdot,\theta))
    \colon \theta\in \R^{P(N)}\big\}, \quad \text{where} \quad 
    \sgn(x)\coloneqq\begin{cases} \phantom{-}1, &\textup{if }x\ge 0, \\ -1, &\textup{if }x<0. \end{cases}
\end{equation*}
\end{definition}
Note that we compose the output of the NN with the sign function in order to obtain functions mapping to $\cY=\{-1,1\}$. This can be generalized to multi-dimensional classification tasks by replacing the sign by an argmax function.
Given a hypothesis set, a popular learning algorithm is \emph{empirical risk minimization} (ERM), which minimizes the average loss on the given training data, as described in the next definitions. 
\begin{definition}[Empirical risk] \label{def:emp_risk}
For training data $\cs=(z^{(i)})_{i=1}^m \in\cZ^m$ and a function $f\in \cM(\cX,\cY)$, we define the empirical risk by 
\begin{equation} \label{eq:emp_risk}
    \widehat \cR_{\cs} (f) \coloneqq\frac{1}{m}\sum_{i=1}^m \cL(f, z^{(i)}).
\end{equation}
\end{definition}
\begin{definition}[ERM learning algorithm]\label{def:ERM}
Given a hypothesis set $\cF$, an empirical risk minimization algorithm $\mathcal{A}^{\mathrm{erm}}$
chooses\footnote{For simplicity, we assume that the minimum is attained which, for instance, is the case if $\cF$ is a compact topological space on which $\widehat \cR_\cs$ is continuous. Hypothesis sets of NNs $\cF_{(N,\varrho)}$ constitute a compact space if, e.g., one chooses a compact parameter set $\cP\subset\R^{P(N)}$ and a continuous activation function $\varrho$. One could also work with approximate minimizers, see~\cite{anthony1999neural}.} for training data $\cs\in\cZ^m$ a minimizer $\widehat f_{\cs}\in \cF$ of the empirical risk in $\cF$, i.e.,
\begin{equation} 
\label{eq:emp_risk_minimizer}
    \cA^{\mathrm{erm}}(\cs) \in \argmin_{f\in\cF} \widehat\cR_\cs(f).
\end{equation}
\end{definition}
\begin{remark}[Surrogate loss and regularization]
\label{rem:surrogate_loss}
Note that, for classification tasks, one needs to optimize over non-differentiable functions with discrete outputs in~\eqref{eq:emp_risk_minimizer}.
For NN hypothesis sets $\cF_{\arch,\sgn}$ one typically uses the corresponding hypothesis set
for regression tasks $\cF_\arch$ to find an approximate minimizer $\widehat f_\cs^{\mathrm{surr}}\in \cF_{\arch}$ of
\begin{equation*}
    \frac{1}{m}\sum_{i=1}^m\cL^{\mathrm{surr}}(f,z^{(i)}),
\end{equation*}
where $\cL^{\mathrm{surr}}\colon \cM(\cX,\R) \times \cZ \to \R$ is a surrogate loss 
guaranteeing that 
$\sgn(\widehat f_\cs^{\mathrm{surr}}) \in \argmin_{f\in \cF_{\arch,\sgn}} \widehat\cR_\cs(f)$.
A frequently used surrogate loss is the logistic loss\footnote{This can be viewed as cross-entropy between the label $y$ and the output of $f$ composed with a logistic function $\varrho_\sigma$. In a multi-dimensional setting one can replace the logistic function with a softmax function.} given by
\begin{equation*}
    \cL^{\mathrm{surr}}(f,z) =   \log \left( 1+e^{-yf(x)}\right).
\end{equation*} 
In various learning tasks one also adds regularization terms to the minimization problem in~\eqref{eq:emp_risk_minimizer}, such as penalties on the norm of the parameters of the NN, i.e.,
\begin{equation*}
    \min_{\theta\in\R^{P(N)}} \widehat\cR_\cs(\Phi_\arch(\cdot,\theta))+\alpha\|\theta\|^2_2,
\end{equation*}
where $\alpha\in(0,\infty)$ is a regularization parameter.
Note that in this case the minimizer depends on the chosen parameters $\theta$ and not only on the realization function $\Phi_\arch(\cdot,\theta)$, see also Remark~\ref{rem:nn_notation}.
\end{remark}
Coming back to our initial, informal description of learning in Definition~\ref{def:learning}, we have now outlined potential learning tasks in Definition~\ref{def:pred_task}, NN hypothesis sets in Definition~\ref{def:nn_hypothesis}, a metric for the in-sample performance in Definition~\ref{def:emp_risk}, and a corresponding learning algorithm in Definition~\ref{def:ERM}. However, we are still lacking a mathematical concept to describe the out-of-sample (generalization) performance of our learning algorithm. This question has been intensively studied in the field of statistical learning theory, see Section~\ref{sec:intro} for various references.

In this field one usually establishes a connection between unseen data $z$ and the training data $\cs=(z^{(i)})_{i=1}^m$ by imposing that $z$ and $z^{(i)}$, $i\in[m]$, are realizations of independent samples drawn from the same distribution. 
\begin{assumption}[Independent and identically distributed data]\label{ass:iid}
We assume that $z^{(1)},\dots,z^{(m)}, z$ are realizations of i.i.d.\@ random variables
$Z^{(1)},\dots,Z^{(m)},Z$.
\end{assumption}
In this formal setting, we can compute the average out-of-sample performance of 
a model. Recall from our notation in Section~\ref{subsec:notation} that we denote by $\P_Z$ the image measure of $Z$ on $\cZ$, which is the underlying distribution of our training data $\cS=(Z^{(i)})_{i=1}^m\sim\P_Z^m$ and unknown data $Z\sim\P_Z$.
\begin{definition}[Risk]\label{def:risk}
For a function $f\in \cM(\cX,\cY)$, we define\footnote{Note that this requires $z\mapsto \cL(f,z)$ to be measurable for every $f\in \cM(\cX,\cY)$, which is the case for our considered prediction tasks.} the risk by
\begin{equation}
    \cR(f)\coloneqq\E\big[\cL(f,Z)\big]    =\int_{\cZ} \cL(f,z) \, \mathrm{d}\P_Z(z).
\end{equation}
Defining $\cS\coloneqq(Z^{(i)})_{i=1}^m$, the risk of a model $f_\cS=\cA(\cS)$ is thus given by
$
    \cR(f_\cS)=\E\big[\cL(f_\cS,Z)|\cS\big]
$.
\end{definition}
For prediction tasks, we can write $Z=(X,Y)$, such that the input features and labels are given by an $\cX$-valued random variable $X$ and a $\cY$-valued random variable $Y$, respectively. Note that for classification tasks the risk equals the probability of misclassification
\begin{equation*} 
\cR(f)=\E[\mathds{1}_{(-\infty,0)}(Y f(X))] = \P[f(X) \neq Y].
\end{equation*}

For noisy data, there might be a positive, lower bound on the risk, i.e., an irreducible error. If the lower bound on the risk is attained, one can also define the notion of an optimal solution to a learning task.
\begin{definition}[Bayes-optimal function]
A function $f^*\in \cM(\cX,\cY)$ achieving the smallest risk, the so-called Bayes risk
\begin{equation*}
    \cR^*\coloneqq\inf_{f\in \cM(\cX,\cY)} \cR(f),
\end{equation*}
is called a Bayes-optimal function.
\end{definition}
For the prediction tasks in Definition~\ref{def:pred_task}, we can represent the risk of a function with respect to the Bayes risk and compute the Bayes-optimal function, see, e.g.,~\cite[Propositions 1.8 and 9.3]{cucker2007learning}.
\begin{lemma}[Regression and classification risk]
\label{lem:risk}
For a regression task with $\V[Y]<\infty$, the risk can be decomposed into
\begin{equation}
\label{eq:regression_risk}
    \cR(f)=\E\big[( f(X) - \E[Y|X] )^2 \big] +  \cR^*, \quad f\in\cM(\cX,\cY),
\end{equation}
which is minimized by the regression function $f^* (x)=\E[Y|X=x]$.
For a classification task, the risk can be decomposed into
\begin{equation}
\label{eq:classification_risk}
\cR(f)=\E\big[| \E[Y|X]|\mathds{1}_{(-\infty,0)}(\E[Y|X]f(X)) \big] +  \cR^*, \quad f\in\cM(\cX,\cY),
\end{equation}
which is minimized by the Bayes classifier
$ f^*(x) = \sgn(\E[Y|X=x])$.
\end{lemma}

As our model $f_\cS$ is depending on the random training data $\cS$, the risk $\cR(f_\cS)$ is a random variable and we might aim\footnote{In order to make probabilistic statements on $\cR(f_\cS)$ we assume that $\cR(f_\cS)$ is a random variable, i.e., measurable. This is, e.g., the case if $\cF$ constitutes a measurable space and $\cs\mapsto\cA(s)$ and $f\to \cR|_\cF$ are measurable.} for $\cR(f_\cS)$ small with high probability or in expectation over the training data.
The challenge for the learning algorithm $\cA$ is to minimize the risk by only using training data but without knowing the underlying distribution. One can even show that for every learning algorithm there exists a distribution where convergence of the expected risk of $f_\cS$ to the Bayes risk is arbitrarily slow with respect to the number of samples $m$~\cite[Theorem 7.2]{devroye1996probabilistic}.
\begin{theorem}[No free lunch] \label{thm:nofree}
    Let $a_m\in(0,\infty)$, $m\in\N$, be a monotonically decreasing sequence with $a_1 \le 1/16$. Then for every learning algorithm $\cA$ of a classification task there exists a distribution $\P_Z$ such that for every $m\in\N$ and training data $\cS\sim\P_Z^m$ it holds that
    \begin{equation*}
        \E\big[\cR(\cA(\cS))\big] \ge  \cR^* + a_m.
    \end{equation*}
\end{theorem}

\begin{figure}[t]
    \centering
    \includegraphics[width = 0.56\textwidth]{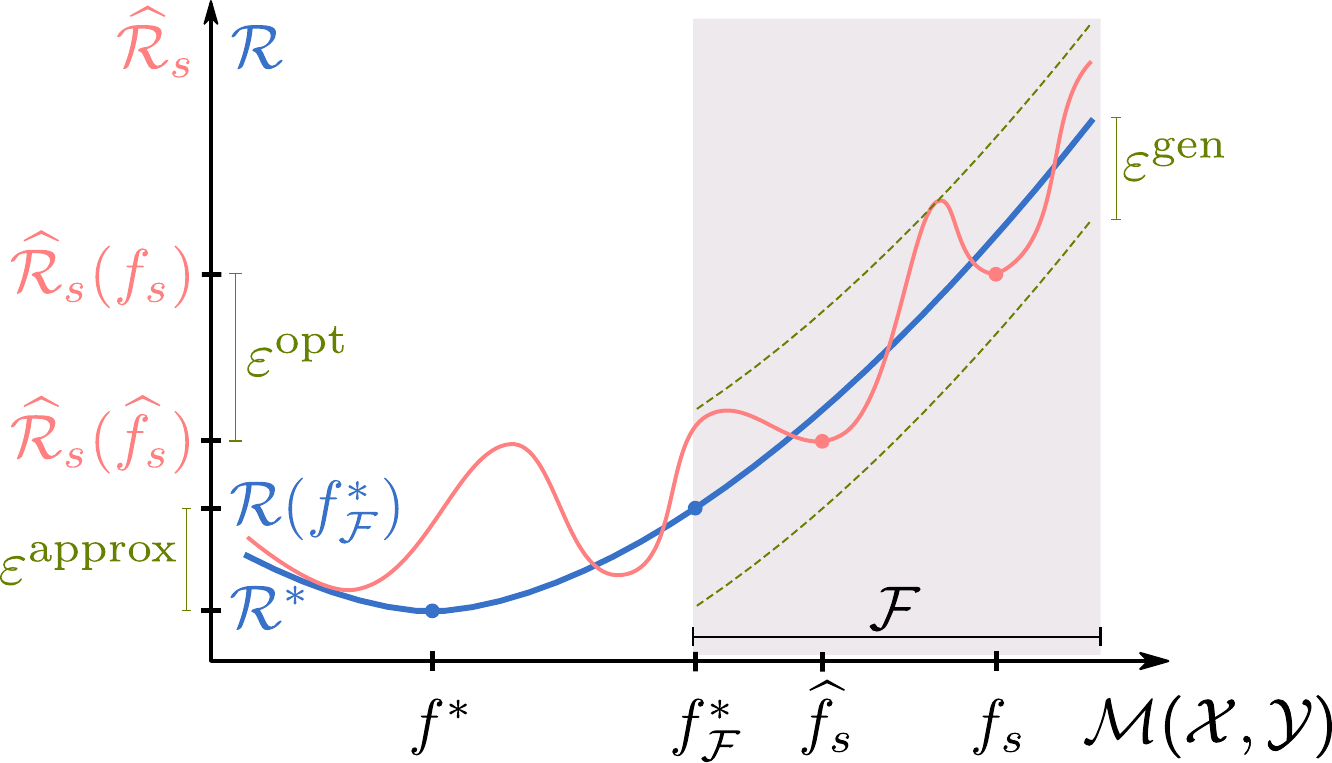}
    \caption{Illustration of the errors~\ref{it:err_1}--\ref{it:err_3} in the decomposition of~\eqref{eq:errorDecompositionClassical}. It shows an exemplary risk $\widehat \cR$ (blue) and empirical risk $\widehat \cR_\cs$ (red) with respect to the projected space of measurable functions $\cM(\cX,\cY)$. Note that the empirical risk and thus $\eps^{\mathrm{gen}}$ and $\eps^{\mathrm{opt}}$ depend on the realization $\cs=(z^{(i)})_{i=1}^m$ of the training data $\cS\sim \P_Z^m$.}
    \label{fig:decomposition}
\end{figure}

Theorem \ref{thm:nofree} shows the non-existence of a universal learning algorithm for every data distribution $\P_Z$ and shows that useful bounds must necessarily be accompanied by a priori regularity conditions on the underlying distribution $\P_Z$. Such prior knowledge can then be incorporated in the choice of the hypothesis set $\cF$. To illustrate this, let $f^*_\cF \in \argmin_{f\in\cF} \cR(f)$ be a best approximation in $\cF$, such that we can bound the error
\begin{equation}
\begin{split}
    \cR(f_\cS) - \cR^* &= \cR(f_\cS) - \widehat \cR_\cS(f_\cS) + \widehat \cR_\cS(f_\cS) - \widehat \cR_\cS(f^*_\cF )  +\widehat \cR_\cS(f^*_\cF ) - \cR( f^*_\cF)+ \cR( f^*_\cF)-\cR^* \\& \leq \eps^{\mathrm{opt}}+
    2\eps^{\mathrm{gen}}+\eps^{\mathrm{approx}} \label{eq:errorDecompositionClassical}
\end{split}
\end{equation}
by
\begin{enumerate}[label=(\Alph*)]
    \item \label{it:err_1} an \emph{optimization error} $\eps^{\mathrm{opt}}\coloneqq\widehat\cR_\cS(f_\cS)-\widehat \cR_\cS( \widehat{f}_\cS ) \geq \widehat\cR_\cS(f_\cS)-\widehat \cR_\cS( f^*_\cF )$, with $\widehat{f}_\cS$ as in Definition~\ref{def:ERM},
    \item \label{it:uniform_generalization} a (uniform\footnote{Although this uniform deviation can be a coarse estimate it is frequently considered to allow for the application of uniform laws of large numbers from the theory of empirical processes.}) \emph{generalization error} $\eps^{\mathrm{gen}}\coloneqq\sup_{f\in\cF} |\cR(f)-\widehat \cR_\cS( f)|\ge \max\{\cR(f_\cS) - \widehat \cR_\cS(f_\cS),\widehat \cR_\cS(f^*_\cF ) - \cR( f^*_\cF)\} $, and
    \item \label{it:err_3} an \emph{approximation error} $\eps^{\mathrm{approx}}\coloneqq\cR(f^*_\cF) - \cR^*$,
\end{enumerate}
see also Figure~\ref{fig:decomposition}.
The approximation error is decreasing when enlarging the hypothesis set, but taking $\cF=\cM(\cX,\cY)$ prevents controlling the generalization error, see also Theorem~\ref{thm:nofree}. This suggests a sweet-spot for the complexity of our hypothesis set $\cF$ and is usually referred to as the \emph{bias-variance trade-off}, see also Figure~\ref{fig:tradeoff} below. In the next sections, we will sketch mathematical ideas to tackle each of the errors in~\ref{it:err_1}--\ref{it:err_3} in the context of deep learning. Observe that we bound the generalization and optimization error with respect to the empirical risk $\widehat \cR_\cS$ and its minimizer $\widehat f_\cS$ which is motivated by the fact that in deep-learning-based applications one typically tries to minimize variants of $\widehat \cR_\cS$.

\subsubsection{Optimization}\label{subsubsec:optimization}

The first error in the decomposition of~\eqref{eq:errorDecompositionClassical} is the optimization error: $\eps^{\mathrm{opt}}$. This error is primarily influenced by the numerical algorithm $\cA$ that is used to find the model $f_\cs$ in a hypothesis set of NNs for given training data $\cs\in\cZ^m$. We will focus on the typical setting where such an algorithm tries to approximately minimize the empirical risk $\widehat \cR_\cs$.
While there are many conceivable methods to solve this minimization problem, by far the most common are gradient-based methods. 
The main reason for the popularity of gradient-based methods is that for FC networks as in Definition~\ref{def:classical_nns}, the accurate and efficient computation of pointwise derivatives $\nabla_\theta \Phi_\arch(x,\theta)$ is possible by means of automatic differentiation, a specific form of which is often referred to as the \emph{backpropagation algorithm}~\cite{kelley1960gradient, dreyfus1962numerical, linnainmaa1970representation, rumelhart1986learning, griewank2008evaluating}. 
This numerical scheme is also applicable in general settings, such as, when the architecture of the NN is given by a general directed acyclic graph.
Using these pointwise derivatives, one usually attempts to minimize the empirical risk $\widehat\cR_\cs$ by updating the parameters $\theta$ according to a variant of \emph{stochastic gradient descent} (SGD), which we shall review below in a general formulation:

\medskip
\begin{algorithm}[H]
\SetAlgoLined
\SetKwInOut{Input}{Input}
\SetKwInOut{Output}{Output}
\Input{\,Differentiable function $r\colon \R^p \to \R$, sequence of step-sizes $\eta_k \in (0,\infty)$, $k\in[K]$, \\ \,$\R^p$-valued random variable $\Theta^{(0)}$.}
\Output{\,Sequence of $\R^p$-valued random variables $(\Theta^{(k)})_{k=1}^K$.}
 \For{$k = 1, \dots, K$}{
  Let $D^{(k)}$ be a random variable such that $\E[D^{(k)}| \Theta^{(k-1)}] = \nabla r(\Theta^{(k-1)})$\;
  Set $\Theta^{(k)} \coloneqq \Theta^{(k-1)} - \eta_k D^{(k)}$\;
 }
 \caption{Stochastic gradient descent}
 \label{alg:1}
\end{algorithm}
\medskip

If $D^{(k)}$ is chosen deterministically in Algorithm~\ref{alg:1}, i.e., $D^{(k)} = \nabla r(\Theta^{(k-1)})$, then the algorithm is known as \emph{gradient descent}. To minimize the empirical loss, we apply SGD with $r \colon \R^{P(N)} \to \R$ set to $r(\theta) = \erisk{\cs}(\Phi_\arch(\cdot,\theta))$.
More concretely, one might choose a \emph{batch-size} $m'\in\N$ with $m' \le m$ and consider the iteration
\begin{equation}\label{eq:SGDForERM}
    \Theta^{(k)} \coloneqq \Theta^{(k-1)} -    \frac{\eta_{k}}{m'}\sum_{z\in\cS'}  \nabla_\theta \cL(\Phi_\arch(\cdot,\Theta^{(k-1)}),z),
\end{equation}
where $\cS'$ is a so-called \emph{mini-batch} of size $|\cS'|=m'$ chosen uniformly\footnote{We remark that in practice one typically picks $\cS'$ by selecting a subset of training data in a way to cover the full training data after one \emph{epoch} of $\lceil m/m' \rceil$ many steps. This, however, does not necessarily yield an unbiased estimator $D^{(k)}$ of $\nabla_\theta r(\Theta^{(k-1)})$ given $\Theta^{(k-1)}$.} at random from the training data $\cs$.
The sequence of step-sizes $(\eta_k)_{k\in \N}$ is often called \emph{learning rate} in this context. Stopping at step $K$, the output of a deep learning algorithm $\cA$ is then given by 
\begin{equation*}
        f_\cs = \cA(\cs) = \Phi_\arch(\cdot,\bar{\theta}),
\end{equation*}
where $\bar{\theta}$ can be chosen to be the realization of the last parameter $\Theta^{(K)}$ of~\eqref{eq:SGDForERM} or a convex combination of $(\Theta^{(k)})_{k=1}^K$ such as the mean.

Algorithm~\ref{alg:1} was originally introduced in~\cite{robbins1951stochastic} in the context of finding the root of a nondecreasing function from noisy measurements. Shortly afterwards this idea was applied to find a unique minimum of a Lipschitz-regular function that has no flat regions away from the global minimum~\cite{kiefer1952stochastic}. 

In some regimes, we can guarantee convergence of SGD at least in expectation, see~\cite{nemirovsky1983problem, nemirovski2009robust, shalev2009stochastic},~\cite[Section 5.9]{shapiro2014lectures},~\cite[Chapter 14]{shalev2014understanding}. One prototypical convergence guarantee that is found in the aforementioned references in various forms is stated below.
\begin{theorem}[Convergence of SGD]\label{thm:convergenceConvexLipschitz}
    Let $p,K\in\N$ and let $r\colon \R^p\supset B_{1}(0) \to \R$ be differentiable and convex.
    Further let $(\Theta^{(k)})_{k=1}^K$ be the output of Algorithm~\ref{alg:1} with initialization $\Theta^{(0)} = 0$, step-sizes $\eta_k = K^{-1/2}$, $k\in[K]$, and random variables $(D^{(k)})_{k=1}^K$ satisfying that $\|D^{(k)}\|_2 \leq 1$ almost surely for all $k \in [K]$. Then 
    \begin{equation*}
    \E[r(\bar{\Theta})] - r(\theta^*) \leq \frac{1}{\sqrt{K}},
    \end{equation*}
    where $\bar{\Theta} \coloneqq \frac{1}{K} \sum_{k=1}^K \Theta^{(k)}$ and $\theta^* \in \argmin_{\theta \in B_{1}(0)} r(\theta)$.
\end{theorem}

\begin{figure}[t]
    \centering
    \includegraphics[width=0.25\textwidth]{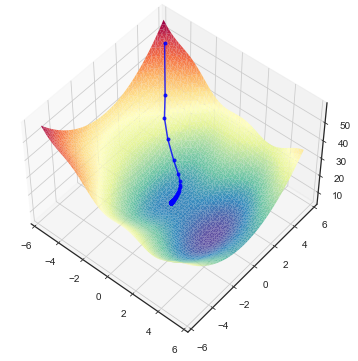} \hspace*{-0.15cm}\includegraphics[width = 0.25\textwidth]{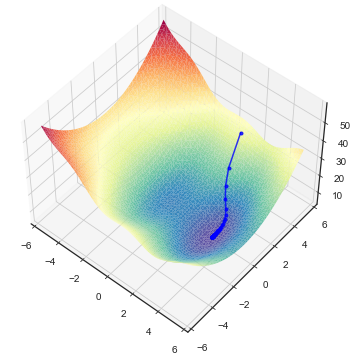}\hspace*{-0.15cm} \includegraphics[width = 0.25\textwidth]{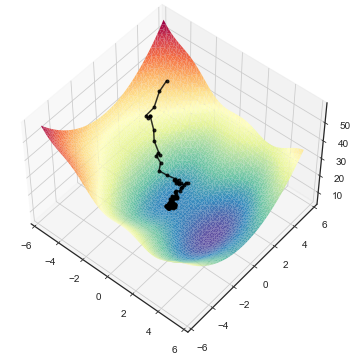}\hspace*{-0.15cm} \includegraphics[width = 0.25\textwidth]{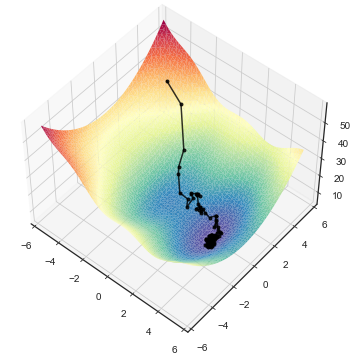}
    \includegraphics[width = 0.24\textwidth]{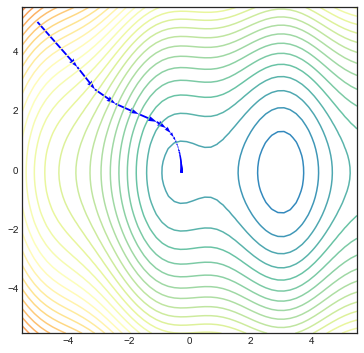} \includegraphics[width = 0.24\textwidth]{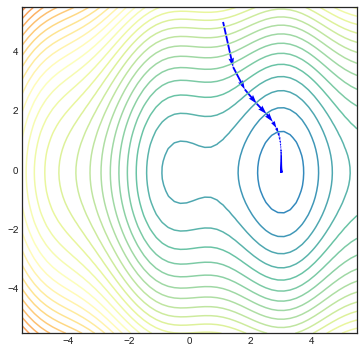} \includegraphics[width = 0.24\textwidth]{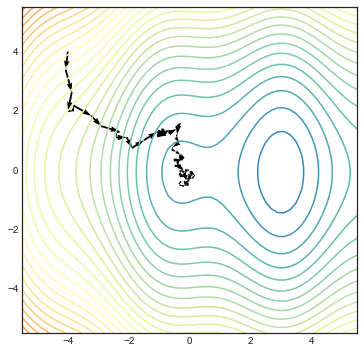} \includegraphics[width = 0.24\textwidth]{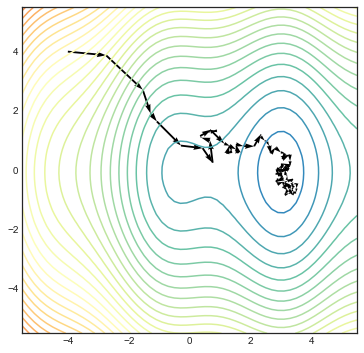}
    \caption{Examples of the dynamics of gradient descent (left) and stochastic gradient descent (right) for an objective function with one non-global minimum next to the global minimum. We see that depending on the initial condition and also on fluctuations in the stochastic part of SGD the algorithm can fail or succeed in finding the global minimum.}
    \label{fig:sgdetcwinornowin}
\end{figure}

Theorem~\ref{thm:convergenceConvexLipschitz} can be strengthened to yield a faster convergence rate if the convexity is replaced by strict convexity. If $r$ is not convex, then convergence to a global minimum can in general not be guaranteed. In fact, in that case, stochastic gradient descent may converge to a local, non-global minimum, see Figure~\ref{fig:sgdetcwinornowin} for an example. 

Moreover, gradient descent, i.e., the deterministic version of Algorithm~\ref{alg:1}, will stop progressing if at any point the gradient of $r$ vanishes. This is the case in every stationary point of $r$. A stationary point is either a local minimum, a local maximum, or a saddle point. One would expect that if the direction of the step $D^{(k)}$ in Algorithm~\ref{alg:1} is not deterministic, then the random fluctuations may allow the iterates to escape saddle points. Indeed, results guaranteeing convergence to local minima exist under various conditions on the type of saddle points that $r$ admits,~\cite{nemirovski2009robust, ghadimi2013stochastic, ge2015escaping, lee2016gradient, jentzenErrorAnalysis2020}. 

In addition, many methods that improve the convergence by, for example, introducing more elaborate step-size rules or a momentum term have been established. We shall not review these methods here, but instead refer to~\cite[Chapter 8]{goodfellow2016deep} for an overview. 

\subsubsection{Approximation}
\label{subsubsec:approximationError}
Generally speaking, NNs, even FC NNs (see Definition~\ref{def:classical_nns}) with only $L=2$ layers, are universal approximators, meaning that under weak conditions on the activation function $\varrho$ they can approximate any continuous function on a compact set up to arbitrary precision~\cite{Cybenko1989, Funahashi1989183, Hornik1989universalApprox, leshno1993}. 

\begin{theorem}[Universal approximation theorem]\label{theo:universal}
    Let $d\in\N$, let $K\subset \R^d$ be compact, and let $\varrho\in L_{\mathrm{loc}}^\infty(\R)$ be an activation function such that the closure of the points of discontinuity of $\varrho$ is a Lebesgue null set. Further let
    \begin{equation*}
    \widetilde{\cF}\coloneqq \bigcup_{n\in \N} \cF_{((d,n,1),\varrho)}
    \end{equation*}
    be the corresponding set of two-layer NN realizations.
    Then it holds that
    $C(K) \subset \operatorname{cl}(\widetilde{\cF})$ (where the closure is taken with respect to the topology induced by the $L^\infty(K)$-norm)
    if and only if there does not exist a polynomial $p\colon\R\to\R$ with $p=\varrho$ almost everywhere.
\end{theorem}

The theorem can be proven by the theorem of Hahn--Banach, which implies that $\widetilde{\cF}$ being dense in some real normed vector space $\mathcal{S}$ is equivalent to the following condition: For all non-trivial functionals $F\in \mathcal{S}'\setminus \{0\}$ from the topological dual space of $\mathcal{S}$ there exist parameters $w\in\R^d$ and $b\in\R$ such that
\begin{equation*}
    F(\varrho(\langle w, \cdot\rangle+b)) \neq 0.
\end{equation*}
In case of $\mathcal{S} = C(K)$ we have by the Riesz--Markov--Kakutani representation theorem that $\mathcal{S}'$ is the space of signed Borel measures on $K$, see~\cite{rudin2006real}. Therefore, Theorem~\ref{theo:universal} holds, if $\varrho$ is such that, for a signed Borel measure $\mu$,
\begin{equation}\label{eq:discriminatory}
    \int_{K} \varrho(\langle w, x\rangle+b) \, \mathrm{d}\mu(x) = 0 
\end{equation}
for all $w \in \R^d$ and $b \in \R$ implies that $\mu = 0$. An activation function $\varrho$ satisfying this condition is called \emph{discriminatory}. It is not hard to see that any sigmoidal $\varrho$ is discriminatory. Indeed, assume that $\varrho$ satisfies~\eqref{eq:discriminatory} for all $w \in \R^d$ and $b \in \R$. Since for every $x\in\R^d$ it holds that $\varrho(ax + b) \to \mathds{1}_{(0,\infty)}(x) + \varrho(b)\mathds{1}_{\{0\}}(x)$ for $a \to \infty$, we conclude by superposition and passing to the limit that for all $c_1, c_2 \in \R$ and $w \in \R^d$, $b \in \R$ 
\begin{equation*}
    \int_{K} \mathds{1}_{[c_1, c_2]}(\langle w, x\rangle+b) \, \mathrm{d}\mu(x) = 0.
\end{equation*} 
Representing the exponential function $x \mapsto e^{-2\pi i x}$ as the limit of sums of elementary functions yields that $\int_{K} e^{-2\pi i (\langle w, x\rangle+b)} \, \mathrm{d}\mu(x) = 0$ for all $w \in \R^d$, $b \in \R$. Hence, the Fourier transform of $\mu$ vanishes which implies that $\mu=0$. 

Theorem~\ref{theo:universal} addresses a uniform approximation problem on a general compact set. If we are given a finite number of points and only care about good approximation at these points, then one can ask if this approximation problem is potentially simpler. Below we see that, if the number of neurons is larger or equal to the number of data points, then one can always interpolate, i.e., exactly fit the data on a given finite number of points. 

\begin{proposition}[Interpolation] 
\label{prop:interpolation}
Let $d,m\in\N$, let $x^{(i)}\in \R^d$, $i\in [m]$, with $x^{(i)}\neq x^{(j)}$ for $i\neq j$, let $\varrho\in C(\R)$, and assume that $\varrho$ is not a polynomial.
Then, there exist parameters $\theta^{(1)}\in \R^{m\times d}\times \R^m$ with the following property: For every $k\in\N$ and every sequence of labels $y^{(i)}\in \R^k$, $i\in[m]$, there exist parameters $\theta^{(2)}=(W^{(2)},0)\in\R^{k\times m}\times \R^k$ for the second layer of the NN architecture $\arch=((d,m,k),\varrho)$ such that
\begin{equation*}
      \Phi_\arch(x^{(i)},(\theta^{(1)},\theta^{(2)})) = y^{(i)}, \quad i\in[m].
\end{equation*}
\end{proposition}

Let us sketch the proof in the following. First, note that Theorem~\ref{theo:universal} also holds for functions $g\in C(K,\R^m)$ with multi-dimensional output by approximating each one-dimensional component $x\mapsto (g(x))_i$ and stacking the resulting networks. Second, one can add an additional row containing only zeros to the weight matrix $W^{(1)}$ of the approximating neural network as well as an additional entry to the vector $b^{(1)}$. The effect of this is that we obtain an additional neuron with constant output. Since $\varrho\neq 0$, we can choose $b^{(1)}$ such that the output of this neuron is not zero. Therefore, we can include the bias vector $b^{(2)}$ of the second layer into the weight matrix $W^{(2)}$, see also Remark~\ref{rem:nn_notation}. Now choose $g\in C(\R^m,\R^m)$ to be a function satisfying $g(x^{(i)})=e^{(i)}$, $i\in[m]$, where $e^{(i)}\in\R^m$ denotes the $i$-th standard basis vector. 
By the discussion before there exists a neural network architecture $\tilde{a}=((d,n,m),\varrho)$ and parameters $\tilde{\theta}=((\widetilde{W}^{(1)},\tilde{b}^{(1)}),(\widetilde{W}^{(2)},0))$ such that 
\begin{equation}
    \label{eq:universal_approx}
    \|\Phi_{\tilde{a}}(\cdot,\tilde{\theta})-g\|_{L^\infty(K)} < \frac{1}{m},
\end{equation}
where $K$ is a compact set with $x^{(i)}\in K$, $i\in [m]$.
Let us abbreviate the output of the activations in the first layer evaluated at the input features by 
\begin{equation}
    \widetilde{A}\coloneqq\begin{bmatrix}\varrho(\widetilde{W}^{(1)}(x^{(1)}) + \tilde{b}^{(1)})) \dots \varrho(\widetilde{W}^{(1)}(x^{(m)}) + \tilde{b}^{(1)})) \end{bmatrix}\in\R^{n \times m}.
\end{equation}
The equivalence of the max and operator norm and~\eqref{eq:universal_approx} establish that
\begin{equation*}
   \| \widetilde{W}^{(2)}\widetilde{A}-\mathrm{I}_m \|_{\mathrm{op}} \le m \max_{i,j \in [m]} \big|( \widetilde{W}^{(2)}\widetilde{A}-\mathrm{I}_m )_{i,j}\big| = m \max_{j\in[m]} \| \Phi_{\tilde{a}}(x^{(j)},\tilde{\theta})-g(x^{(j)}) \|_\infty < 1,
\end{equation*}
where $\mathrm{I}_m$ denotes the $m \times m$ identity matrix. Thus, the matrix $\widetilde{W}^{(2)}\widetilde{A}\in\R^{m\times m}$ needs to have full rank and we can extract $m$ linearly independent rows from $\widetilde{A}$ resulting in an invertible matrix $A\in\R^{m \times m}$. Now, we define the desired parameters $\theta^{(1)}$ for the first layer by extracting the corresponding rows from $\widetilde{W}^{(1)}$ and $\tilde{b}^{(1)}$ and the parameters $\theta^{(2)}$ of the second layer by 
\begin{equation*}
    W^{(2)} \coloneqq  \begin{bmatrix}y^{(1)} \dots y^{(m)} \end{bmatrix} A^{-1} \in \R^{k \times m}.
\end{equation*}
This proves that with any discriminatory activation function we can interpolate arbitrary training data $(x^{(i)},y^{(i)})\in \R^d\times \R^k$, $i\in [m]$, using a two-layer NN with $m$ hidden neurons, i.e., $\cO(m(d+k))$ parameters. 

One can also first project the input features to a one-dimensional line where they are separated and then apply Proposition~\ref{prop:interpolation} with $d=1$. For nearly all activation functions, this can be represented by a three-layer NN using only $\cO(d+mk)$ parameters\footnote{To avoid the $m\times d$ weight matrix (without using shared parameters as in~\cite{zhang2016understanding}) one interjects an approximate one-dimensional identity~\cite[Definition 2.5]{petersen2018optimal}, which can be arbitrarily well approximated by a NN with architecture $a=((1,2,1),\varrho)$ given that $\varrho'(\lambda)\neq 0$ for some $\lambda \in\R$, see~\eqref{eq:representationofMonomials} below.}.

Beyond interpolation results, one can obtain a quantitative version of Theorem~\ref{theo:universal} if one knows additional regularity properties of the Bayes optimal function $f^*$, such as smoothness, compositionality, and symmetries.
For surveys on such results, we refer the reader to
\ifbook
\else%
\cite{devore2020neural, guhring2020expressivity}.
\fi%
For instructive purposes, we review one such result, which can be found in~\cite[Theorem 2.1]{mhaskar1996neural}, below:
\begin{theorem}[Approximation of smooth functions] \label{thm:MhaskarForPresidentOfNeuralNetworkLand}
    Let $d, k \in \N$ and $p \in [1,\infty]$. Further let $\varrho \in C^\infty(\R)$ and assume that $\varrho$ is not a polynomial. Then there exists a constant $c\in(0,\infty)$ with the following property: For every $n\in\N$ there exist parameters $\theta^{(1)}\in \R^{n\times d}\times \R^n$ for the first layer of the NN architecture $\arch=((d,n,1),\varrho)$ 
    such that for every $g \in W^{k,p}((0,1)^d)$ 
    it holds that
    \begin{equation*}
       \inf_{\theta^{(2)}\in\R^{1 \times n}\times \R} \| \Phi_\arch(\cdot,(\theta^{(1)},\theta^{(2)}))-g\|_{L^p((0,1)^d)} \leq c n^{-\frac{d}{k}} \|g \|_{W^{k,p}((0,1)^d)}.
    \end{equation*}
\end{theorem}

Theorem~\ref{thm:MhaskarForPresidentOfNeuralNetworkLand} shows that NNs achieve the same optimal approximation rates that, for example, spline-based approximation yields for smooth functions. The idea behind this theorem is based on a strategy that is employed repeatedly throughout the literature. This is the idea of re-approximating classical approximation methods by NNs and thereby transferring the approximation rates of these methods to NNs. In the example of Theorem~\ref{thm:MhaskarForPresidentOfNeuralNetworkLand}, approximation by polynomials is used. The idea is that due to the non-vanishing derivatives of the activation function\footnote{The Baire category theorem ensures that for a non-polynomial $\varrho\in C^\infty(\R)$ there exists $\lambda\in\R$ with $\varrho^{(p)}(\lambda)\neq 0$ for all $p\in\N$, see, e.g.,~\cite[Chapter 10]{william1969distributions}.}, one can approximate every univariate polynomial via divided differences of the activation function. Specifically, accepting unbounded parameter magnitudes, for any activation function $\varrho\colon\R\to\R$ which is \emph{p-times differentiable} at some point $\lambda\in\R$ with $\varrho^{(p)}(\lambda)\neq 0$, one can approximate the monomial $x\mapsto x^p$ on a compact set $K\subset \R$ up to arbitrary precision by a fixed-size NN via rescaled $p$-th order difference quotients as
\begin{equation}\label{eq:representationofMonomials}
\limp_{h \to 0} \ \sup_{x\in K} \Big|\sum_{i=0}^p \frac{ (-1)^i\binom{p}{i}}{h^p\varrho^{(p)}(\lambda)}\varrho\big((p/2-i)hx+\lambda\big) - x^p \Big| = 0.
\end{equation}

Let us end this subsection by clarifying the connection of the approximation results above to the error decomposition of~\eqref{eq:errorDecompositionClassical}. 
Consider, for simplicity, a regression task with quadratic loss. Then, the approximation error $\eps^{\mathrm{approx}}$ equals a common $L^2$-error
\begin{equation*}
\begin{split}
    \eps^{\mathrm{approx}} = \cR(f^*_\cF) - \cR^* 
    & \overset{(*)}{=} \int_{\cX} (f^*_\cF(x) - f^*(x))^2 \, \mathrm{d}\P_X(x)\\
    & \overset{(*)}{=} \min_{f\in\cF} \|f - f^*\|_{L^2(\P_X)}^2\\
    & \leq \min_{f\in\cF} \|f - f^*\|_{L^\infty(\mathcal{X})}^2,
\end{split}
\end{equation*}
where the identities marked by $(*)$ follow from Lemma~\ref{lem:risk}. Hence, Theorem~\ref{theo:universal} postulates that $\eps^{\mathrm{approx}} \to 0$ for increasing NN sizes, whereas Theorem~\ref{thm:MhaskarForPresidentOfNeuralNetworkLand} additionally explains how fast $\eps^{\mathrm{approx}}$ converges to 0.

\subsubsection{Generalization}
\label{subsubsec:gen_error}

Towards bounding the generalization error
$
\eps^{\mathrm{gen}} = \sup_{f\in\cF}|\cR(f)-\widehat \cR_\cS( f)|
$,
one observes that, for every $f\in\cF$, Assumption~\ref{ass:iid} ensures that $\cL(f, Z^{(i)})$, $i\in[m]$, are i.i.d.\@ random variables. Thus, one can make use of concentration inequalities to bound the deviation of the empirical risk $\widehat \cR_\cS( f)=\frac{1}{m}\sum_{i=1}^m\cL(f, Z^{(i)})$ from its expectation $\cR( f)$. For instance, assuming boundedness\footnote{Note that for our classification tasks in Definition~\ref{def:pred_task} it holds that $\cL(f,Z) \in \{0,1\}$ for every $f\in\cF$. For the regression tasks, one typically assumes boundedness conditions, such as $|Y|\le c$ and $\sup_{f\in\cF} |f(X)|\le c$ almost surely for some $c\in(0,\infty)$, which yields that $\sup_{f\in\cF} |\cL(f,Z)| \le 4c^2$.} of the loss, Hoeffding's inequality~\cite{hoeffding1963probability} and a union bound directly imply the following generalization guarantee for countable, weighted hypothesis sets $\cF$, see, e.g.,~\cite{bousquet2003introduction}.
\begin{theorem}[Generalization bound for countable, weighted hypothesis sets] \label{thm:gen_finite}
Let $m\in\N$, $\delta\in(0,1)$ and assume that $\cF$ is countable. Further let $p$ be a probability distribution on $\cF$ and assume that $\mathcal{L}(f,Z) \in [0, 1]$ almost surely for every $f\in \cF$. Then with probability $1-\delta$ (with respect to repeated sampling of $\P_Z^m$-distributed training data $\cS$) it holds for every $f\in\cF$ that
\begin{equation*}
  |\cR(f)-\widehat \cR_\cS( f)| \le \sqrt{\frac{\ln(1/p(f))+\ln(2/\delta)}{2m}}.
\end{equation*}
\end{theorem}

While the weighting $p$ needs to be chosen before seeing the training data, one could incorporate prior information on the learning algorithm $\cA$. For finite hypothesis sets without prior information, setting $p(f)=1/|\cF|$ for every $f\in\cF$, Theorem~\ref{thm:gen_finite} implies that, with high probability, it holds that
\begin{equation}
\label{eq:finite_gen}
    \eps^{\mathrm{gen}} \lesssim  \sqrt{\frac{\ln(|\cF|)}{m}}.
\end{equation}
Again, one notices that, in line with the bias-variance trade-off, the generalization bound is increasing with the size of the hypothesis set $|\cF|$. Although in practice the parameters $\theta\in\R^{P(N)}$ of a NN are discretized according to floating-point arithmetic, the corresponding quantities $|\cF_{a}|$ or $|\cF_{a,\sgn}|$ would be huge and we need to find a replacement for the finiteness condition.

We will focus on binary classification tasks and present a main result of VC theory which is to a great extent derived from the work of Vladimir Vapnik and Alexey Chervonenkis~\cite{vapnik1971uniform}. While in~\eqref{eq:finite_gen} we counted the number of functions in $\cF$, we now refine this analysis to the number of functions restricted to a finite subset of $\cX$, given by the \emph{growth function}
\begin{equation*}
    \operatorname{growth}(m,\cF)\coloneqq\max_{(x^{(i)})_{i=1}^m \in \cX^m} |\{f|_{(x^{(i)})_{i=1}^m}\colon f \in \cF\}|.
\end{equation*}
The growth function can be interpreted as the maximal number of classification patterns in $\{-1,1\}^m$ which functions in $\cF$ can realize on $m$ points and thus $\operatorname{growth}(m,\cF) \le 2^m$.
The asymptotic behavior of the growth function is determined by a single intrinsic dimension of our hypothesis set $\cF$, the so-called \emph{VC-dimension}
\begin{equation*}
\operatorname{VCdim}(\cF)\coloneqq\sup \big\{m\in\N \cup \{0\} \colon \operatorname{growth}(m,\cF)=2^m \big\},
\end{equation*} 
which defines the largest number of points such that $\cF$ can realize any classification pattern, see, e.g.,~\cite{anthony1999neural, bousquet2003introduction}.
There exist various results on VC-dimensions of NNs with different activation functions, see, for instance,~\cite{baum1989size, karpinski1997polynomial, bartlett1998almost, sakurai1999tight}. We present the result of~\cite{bartlett1998almost} for piecewise polynomial activation functions $\varrho$. It establishes a bound on the VC-dimension of hypothesis sets of NNs for classification tasks $\cF_{(N,\varrho),\sgn}$ that scales, up to logarithmic factors, linear in the number of parameters $P(N)$ and quadratic in the number of layers $L$.
\begin{theorem}[VC-dimension of neural network hypothesis sets]
\label{thm:nn_vc}
Let $\varrho$ be a piecewise polynomial activation function. Then there exists a constant $c\in(0,\infty)$ such that for every $L\in\N$ and $N\in\N^{L+1}$
it holds that
\begin{equation*}
    \operatorname{VCdim}(\cF_{(N,\varrho),\sgn})\le c \big( P(N) L \log(P(N)) + P(N)L^2\big).
\end{equation*}
\end{theorem}

Given $(x^{(i)})_{i=1}^m\in \cX^m$, there exists a partition of $\R^{P(N)}$ such that $\Phi(x^{(i)},\cdot)$, $i\in[m]$, are polynomials on each region of the partition. The proof of Theorem~\ref{thm:nn_vc} is based on bounding the number of such regions and the number of classification patterns of a set of polynomials. 

A finite VC-dimension ensures the following generalization bound~\cite{talagrand1994sharper,anthony1999neural}:
\begin{theorem}[VC-dimension generalization bound]
\label{thm:vc_gen_bound}
    There exists a constant $c\in(0,\infty)$ with the following property: For every classification task as in Definition~\ref{def:pred_task}, every $\cZ$-valued random variable $Z$, and every $m\in\N$, $\delta\in(0,1)$ it holds with probability $1-\delta$ (with respect to repeated sampling of $\P_Z^m$-distributed training data $\cS$) that
\begin{equation*}
  \sup_{f\in\cF}|\cR(f)-\widehat \cR_\cS( f)| \le c\sqrt{\frac{\operatorname{VCdim}(\cF)+\log(1/\delta))}{m}}.
\end{equation*}
\end{theorem}

In summary, using NN hypothesis sets $\cF_{(N,\varrho),\sgn}$ with a fixed depth and piecewise polynomial activation $\varrho$
for a classification task, with high probability it holds that
\begin{equation}
\label{eq:vc_bound}
    \eps^{\mathrm{gen}} \lesssim  \sqrt{\frac{P(N)\log(P(N))}{m}}.
\end{equation}

In the remainder of this section we will sketch a proof of Theorem~\ref{thm:vc_gen_bound} and, in doing so, present further concepts and complexity measures connected to generalization bounds.
We start by observing that McDiarmid's inequality~\cite{mcdiarmid1989method} ensures that $\eps^{\mathrm{gen}}$ is sharply concentrated around its expectation, i.e., with probability $1-\delta$ it holds that\footnote{For precise conditions to ensure that the expectation of $\eps^{\mathrm{gen}}$ is well-defined, we refer the reader to~\cite{vaart1997weak, dudley2014uniform}.}
\begin{equation}
\label{eq:gen_concentration}
   \big |\eps^{\mathrm{gen}} - \E\big[\eps^{\mathrm{gen}}\big]\big| \lesssim
   \sqrt{\frac{\log(1/\delta)}{m}}.
\end{equation}

To estimate the expectation of the uniform generalization error we employ a \emph{symmetrization argument}~\cite{gine1984some}.
Define $\cG\coloneqq\cL\circ \cF \coloneqq \{ \cL(f,\cdot) \colon f\in\cF\}$, let $\widetilde{\cS}=(\widetilde{Z}^{(i)})_{i=1}^m\sim\P_Z^m$ be a test data set independent of $\cS$,
and note that $\cR(f) = \E[\widehat \cR_{\widetilde{\cS}}(f)]$.
By properties of the conditional expectation and Jensen's inequality it holds that
\begin{equation*}
\begin{split}
	\E\big[
    \eps^{\mathrm{gen}}
    \big] = \E\Big[\sup_{f\in\cF}|\cR(f)-\widehat \cR_\cS( f)|\Big]&=\E\Big[ \sup_{g\in\cG} \frac{1}{m} \big| \sum_{i=1}^m  \E\big[ g(\widetilde{Z}^{(i)})-g(Z^{(i)})| \cS\big]\big|  \Big] \\
	&\le \E\Big[ \sup_{g\in\cG} \frac{1}{m} \big| \sum_{i=1}^m g(\widetilde{Z}^{(i)})-g(Z^{(i)})\big| \Big] \\
	&=\E\Big[ \sup_{g\in\cG} \frac{1}{m} \big| \sum_{i=1}^m  \tau_i \big(g(\widetilde{Z}^{(i)})-g(Z^{(i)})\big)\big| \Big] \\
	&\le 2\E\Big[ \sup_{g\in\cG} \frac{1}{m} \big| \sum_{i=1}^m  \tau_i g(Z^{(i)})\big| \Big],
\end{split}
\end{equation*}
where we used that multiplications with Rademacher variables $(\tau_1,\dots,\tau_m)\sim\cU(\{-1,1\}^m)$ only amount to interchanging $Z^{(i)}$ with $\widetilde{Z}^{(i)}$ which has no effect on the expectation, since $Z^{(i)}$ and $\widetilde{Z}^{(i)}$ have the same distribution.
The quantity
\begin{equation*}
    \fR_m(\cG)\coloneqq
    \E\Big[\sup_{g\in \cG}\big|\frac{1}{m}\sum_{i=1}^m \tau_i g(Z^{(i)})\big|\Big]
\end{equation*}
is called the \emph{Rademacher complexity}\footnote{Due to our decomposition in~\eqref{eq:errorDecompositionClassical}, we want to uniformly bound the absolute value of the difference between the risk and the empirical risk. It is also common to just bound $\sup_{f\in\cF} \cR(f)-\widehat \cR_\cS( f)$ leading to a definition of the Rademacher complexity without the absolute values which can be easier to deal with.} of $\cG$. One can also prove a corresponding lower 
bound~\cite{vaart1997weak}, i.e.,
\begin{equation}
\label{eq:symmetrization}
     \fR_m(\cG)-\frac{1}{\sqrt{m}} \lesssim \E\big[
    \eps^{\mathrm{gen}}
    \big] \lesssim \fR_m(\cG).
\end{equation}
Now we use a \emph{chaining method} to bound the Rademacher complexity of $\cF$ by covering numbers on different scales. Specifically, Dudley's entropy integral~\cite{DUDLEY1967290,ledoux1991probability}
implies that
\begin{equation}
\label{eq:chaining}
   \fR_m(\cG) \lesssim \E\Big[ \int_0^\infty \sqrt{\frac{\log N_\alpha(\cG,d_\cS)}{m}}\, \mathrm{d}\alpha\Big],
\end{equation}
where
\begin{equation*}
    N_\alpha(\cG,d_\cS)\coloneqq\inf\Big\{|G|\colon G\subset \cG, \, \cG\subset \bigcup_{g\in G} B^{d_\cS}_\alpha(g)   \Big\}
\end{equation*}
denotes the covering number with respect to the (random) pseudometric given by
\begin{equation*}
    d_\cS(f,g)=d_{(Z^{(i)})_{i=1}^m}(f,g) \coloneqq \sqrt{ \frac{1}{m} \sum_{i=1}^m \big(f(Z^{(i)})-g(Z^{(i)})\big)^2}.
\end{equation*}
For the $0$-$1$ loss $\cL(f,z)=\mathds{1}_{(-\infty,0)}(yf(x))=(1-f(x)y)/2$, we can get rid of the loss function by the fact that
\begin{equation}
\label{eq:strip_indicator_loss}
    N_\alpha(\cG,d_S) =  N_{2\alpha}(\cF,d_{(X^{(i)})_{i=1}^m}).
\end{equation}
The proof is completed by combining the inequalities in~\eqref{eq:gen_concentration},~\eqref{eq:symmetrization},~\eqref{eq:chaining} and~\eqref{eq:strip_indicator_loss} with a result of David Haussler~\cite{haussler1995sphere} which shows that for $\alpha\in(0,1)$ we have
\begin{equation}
\label{eq:haussler}
    \log(N_\alpha(\cF,d_{(X^{(i)})_{i=1}^m}))\lesssim  \operatorname{VCdim}(\cF)\log(1/\alpha).
\end{equation}

We remark that this resembles a typical behavior of covering numbers. For instance, the logarithm of the covering number $\log(N_\alpha(\cM))$ of a compact $d$-dimensional Riemannian manifold $\cM$ essentially scales like $d\log(1/\alpha)$.
Finally, note that there exists a similar bound to the one in~\eqref{eq:haussler} for bounded regression tasks making use of the so-called \emph{fat-shattering dimension}~\cite[Theorem 1]{mendelson2003entropy}.

\subsection{Do we need a new theory?}\label{subsec:NewTheories}

Despite the already substantial insight that the classical theories provide, a lot of open questions remain. We will outline these questions below. The remainder of this
\ifbook
\else%
article
\fi%
then collects modern approaches to explain the following issues:
\paragraph{Why do large neural networks not overfit?}
    In Subsection~\ref{subsubsec:approximationError}, we have observed that three-layer NNs with commonly used activation functions and only
    $\cO(d+m)$ parameters
    can interpolate any training data $(x^{(i)},y^{(i)}) \in \R^d \times \R$, $i\in [m]$.
    While this specific representation might not be found in practice,~\cite{zhang2016understanding} indeed trained convolutional\footnote{The basic definition of a convolutional NN will be given in Section~\ref{sec:architectures}. In~\cite{zhang2016understanding} more elaborate versions such as an \emph{Inception} architecture~\cite{szegedy2015going} are employed.} NNs with ReLU activation function and about $1.6$ million parameters to achieve zero empirical risk on $m=50000$ training images of the CIFAR10 dataset~\cite{krizhevsky2009learning} with $32\times 32$ pixels per image, i.e., $d=1024$.
    For such large NNs, generalization bounds scaling with the number of parameters $P(N)$ as the VC-dimension bound in~\eqref{eq:vc_bound} are vacuous. However, they observed close to state-of-the-art generalization performance\footnote{In practice one usually cannot measure the risk $\cR(f_\cs)$ and instead evaluates the performance of a trained model $f_\cs$ by $\widehat \cR_{\tilde{\cs}}(f_\cs)$ using test data $\tilde{s}$, i.e., realizations of i.i.d.\@ random variables distributed according to $\P_Z$ and drawn independently of the training data. In this context one often calls $\cR_\cs(f_\cs)$ the \emph{training error} and $\cR_{\tilde{\cs}}(f_\cs)$ the \emph{test error}.}.
    
    Generally speaking, NNs in practice are observed to generalize well despite having more parameters than training samples (usually referred to as \emph{overparametrization}) and
    approximately interpolating the training data (usually referred to as \emph{overfitting}).
    As we cannot perform any better on the training data, there is no trade-off between fit to training data and complexity of the hypothesis set $\cF$ happening, seemingly contradicting the classical bias-variance trade-off of statistical learning theory. This is quite surprising, especially given the following additional empirical observations in this regime, see~\cite{neyshabur2014search, zhang2016understanding, neyshabur2017exploring, belkin2019reconciling, nakkiran2020deep}:
    \begin{enumerate}
    \item \textit{Zero training error on random labels:} Zero empirical risk can also be achieved for random labels using the same architecture and training scheme with only slightly increased training time: This suggests that the considered hypothesis set of NNs $\cF$ can fit arbitrary binary labels, which would imply that $\operatorname{VCdim}(\cF)\approx m$ or $\fR_m(\cF)\approx 1$ rendering our uniform generalization bounds in Theorem~\ref{thm:vc_gen_bound} and in~\eqref{eq:symmetrization} vacuous.
    \item \textit{Lack of explicit regularization:} The test error depends only mildly on explicit regularization like norm-based penalty terms or dropout 
    (see~\cite{geron2017hands} for an explanation of different regularization methods): As such regularization methods are typically used to decrease the complexity of $\cF$, one might ask if there is any \emph{implicit} regularization (see Figure~\ref{fig:tradeoff}), constraining the range of our learning algorithm $\cA$ to some smaller, potentially data-dependent subset, i.e., $\cA(\cs)\in \widetilde{\cF}_\cs\subsetneq \cF$.
    \begin{figure}[t]
        \centering
        \includegraphics[width = \textwidth]{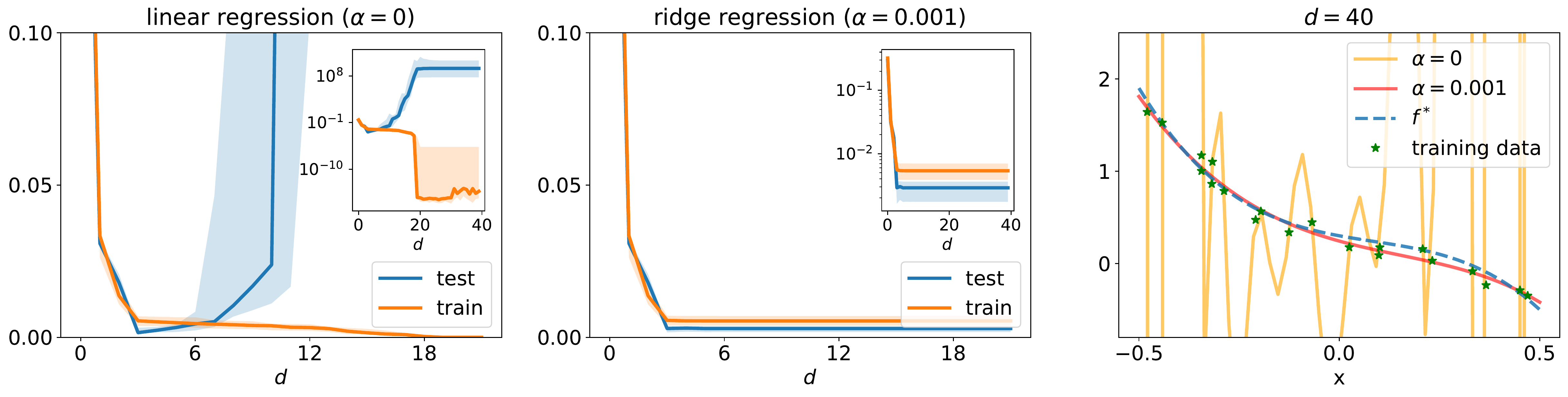}
        \caption{The first plot (and its semi-log inset) shows median and interquartile range of the test and training errors of ten independent linear regressions with $m=20$ samples, polynomial input features $X=(1,Z,\dots,Z^d)$ of degree $d\in[40]$, and labels $Y=f^*(Z)+\nu$, where $Z\sim\mathcal{U}([-0.5,0.5])$, $f^*$ is a polynomial of degree three, and $\nu\sim\cN(0, 0.01)$. This clearly reflects the classical u-shaped bias-variance curve with a sweet-spot at $d=3$ and drastic overfitting beyond the interpolation threshold at $d=20$. However, the second plot shows that we can control the complexity of our hypothesis set
        of linear models by restricting the Euclidean norm of their parameters using ridge regression with a small regularization parameter $\alpha=10^{-3}$, i.e., minimizing the regularized empirical risk $\frac{1}{m} \sum_{i=1}^m (\Phi(X^{(i)},\theta)-Y^{(i)} )^2 + \alpha\|\theta\|_2^2$,
        where $\Phi(\cdot,\theta)= \langle \theta, \cdot \rangle$. 
        Corresponding examples of $\widehat f_\cs $ are depicted in the last plot.}
        \label{fig:tradeoff}
    \end{figure}
    \item \textit{Dependence on the optimization:} The same NN trained to zero empirical risk using different variants of SGD or starting from different initializations can exhibit different test errors: This indicates that the dynamics of gradient descent and properties of the local neighborhood around the model $f_\cs=\cA(\cs)$ might be correlated with generalization performance. 
    \item \textit{Interpolation of noisy training data:} One still observes low test error when training up to approximately zero empirical risk using a regression (or surrogate) loss on noisy training data. This is particularly interesting, as the noise is captured by the model but seems not to hurt generalization performance.
    \item \textit{Further overparametrization improves generalization performance:} Further increasing the NN size can lead to even lower test error: Together with the previous item, this might ask for a different treatment of models complex enough to fit the training data. According to the traditional lore \enquote{The training error tends to decrease whenever we increase the model complexity, that is, whenever we fit the data harder. However with too much fitting, the model adapts itself too closely to the training data, and will not generalize well (i.e., have large test error)},~\cite{hastie2001elements}. While this flawlessly describes the situation for certain machine learning tasks (see Figure~\ref{fig:tradeoff}), it seems not to be directly applicable here.
    \end{enumerate}
    In summary, this suggests that the generalization performance of NNs depends on an interplay of the data distribution $\P_Z$ combined with properties of the learning algorithm $\cA$, such as the optimization procedure and its range. In particular, classical uniform bounds as in Item~\ref{it:uniform_generalization} of our error decomposition might only deliver insufficient explanation, see also~\cite{nagarajan2019uniform}. 
    The mismatch between predictions of classical theory and the practical generalization performance of deep NNs is often referred to as \emph{generalization puzzle}. In Section~\ref{sec:gen} we will present possible explanations for this phenomenon. 
    
    \paragraph{What is the role of depth?}
    
    We have seen in Subsection~\ref{subsubsec:approximationError} that NNs can closely approximate every function if they are sufficiently wide~\cite{Cybenko1989, Funahashi1989183, Hornik1989universalApprox}. There are additional classical results that even provide a trade-off between the width and the approximation accuracy~\cite{chui1994neural, mhaskar1996neural, maiorov1999lower}. In these results, the central concept is the width of a NN. In modern applications, however at least as much focus if not more lies on the depth of the underlying architectures, which can have more than 1000 layers~\cite{he2016deep}. After all, the depth of NNs is responsible for the name of deep learning. 

    This consideration begs the question of whether there is a concrete mathematically quantifiable benefit of deep architectures over shallow NNs. Indeed, we will see effects of depth at many places throughout this 
    \ifbook
\else%
    manuscript.
    \fi%
    However, one of the aspects of deep learning that is most clearly affected by deep architectures is the approximation theoretical aspect. In this framework, we will discuss in Section~\ref{sec:expressivity} multiple approaches that describe the effect of depth.

    \paragraph{Why do neural networks perform well in very high-dimensional environments?}

    We have seen in Subsection~\ref{subsubsec:approximationError} and will see in Section~\ref{sec:expressivity} that from the perspective of approximation theory deep NNs match the performance of the best classical approximation tool in virtually every task. In practice, we observe something that is even more astounding. In fact, NNs seem to perform incredibly well on tasks that no classical, non-specialized approximation method can even remotely handle. The approximation problem that we are talking about here is that of approximation of high-dimensional functions. Indeed, the classical \emph{curse of dimensionality}~\cite{bellman1952theory, novak2009approximation} postulates that essentially every approximation method deteriorates exponentially fast with increasing dimension.

    For example, for the uniform approximation error of 1-Lipschitz continuous functions on a $d$-dimensional unit cube in the uniform norm, we have a lower bound of $\Omega(p^{-1/d})$, for $p \to \infty$, when approximating with a continuous scheme\footnote{One can achieve better rates at the cost of discontinuous (with respect to the function to be approximated) parameter assignment. This can be motivated by the use of space-filling curves. In the context of NNs with piecewise polynomial activation functions, a rate of $p^{-2/d}$ can be achieved by very deep architectures~\cite{yarotsky2018optimal,yarotsky2020phase}.} of $p$ free parameters~\cite{devore1998nonlinear}.
    
    On the other hand, in most applications, the input dimensions are massive. For example, the following datasets are typically used as benchmarks in image classification problems: MNIST~\cite{lecun1998gradient} with $28 \times 28$ pixels per image, CIFAR-10/CIFAR-100~\cite{krizhevsky2009learning} with $32\times 32$ pixels per image and ImageNet~\cite{deng2009imagenet, krizhevsky2012imagenet} which contains high-resolution images
    that are typically down-sampled to $256\times 256$ pixels. Naturally, in real-world applications, the input dimensions may well exceed those of these test problems. However, already for the simplest of the test cases above, the input dimension is $d = 784$. If we use $d = 784$ in the aforementioned lower bound for the approximation of 1-Lipschitz functions, then we require $\cO(\eps^{-784})$ parameters to achieve a uniform error of $\eps \in (0,1)$. Already for moderate $\eps$ this value will quickly exceed the storage capacity of any conceivable machine in this universe.
    Considering the aforementioned curse of dimensionality, it is puzzling to see that NNs perform adequately in this regime. In Section~\ref{sec:CurseOfDimension}, we describe three approaches that offer explanations as to why deep NN-based approximation is not rendered meaningless in the context of high-dimensional input dimensions.

    \paragraph{Why does stochastic gradient descent converge to good local minima despite the non-convexity of the problem?} 
    
    As mentioned in Subsection~\ref{subsubsec:optimization}, a convergence guarantee of stochastic gradient descent to a global minimum is typically only given if the underlying objective function admits some form of convexity. However, the empirical risk of a NN, i.e., $\erisk{\cs}(\Phi(\cdot, \theta))$, is typically not a convex function with respect to the parameters $\theta$. For a simple intuitive reason why this function fails to be convex, it is instructive to consider the following example.
    \begin{example}\label{ex:nonconvexity}
    Consider the NN
    \begin{equation*}
        \Phi(x, \theta) = \theta_1 \ReLU( \theta_3 x + \theta_5) + \theta_2\ReLU(\theta_4 x + \theta_6), \qquad \theta \in \R^6, \quad x \in \R,
    \end{equation*}
    with the ReLU activation function $\ReLU(x)=\max\{0,x\}$.
    It is not hard to see that the two parameter values $\theta = (1, -1, 1, 1, 1, 0)$ and $\bar{\theta} = (-1, 1, 1, 1, 0, 1)$ produce the same realization function\footnote{This corresponds to interchanging the two neurons in the hidden layer. In general it holds that the realization function of a FC NN is invariant under permutations of the neurons in a given hidden layer.}, i.e., $\Phi(\cdot, \theta) =  \Phi(\cdot, \bar{\theta})$. However, since $(\theta + \bar{\theta})/2 = (0, 0, 1, 1, 1/2, 1/2)$, we conclude that $\Phi(\cdot, (\theta + \bar{\theta})/2) = 0$. Clearly, for the data $\cs = ((-1, 0 ), (1,1))$, we now have that 
    \begin{equation*}
        \erisk{\cs}(\Phi(\cdot, \theta)) = \erisk{\cs}(\Phi(\cdot, \bar{\theta}) ) = 0 \quad \text{and} \quad \erisk{\cs}\left(\Phi(\cdot, (\theta + \bar{\theta})/2) \right) = \frac{1}{2},
    \end{equation*} 
    showing the non-convexity of $\erisk{\cs}$. 
\end{example}

\begin{figure}[t]
    \centering
    \includegraphics[width = 0.4\textwidth]{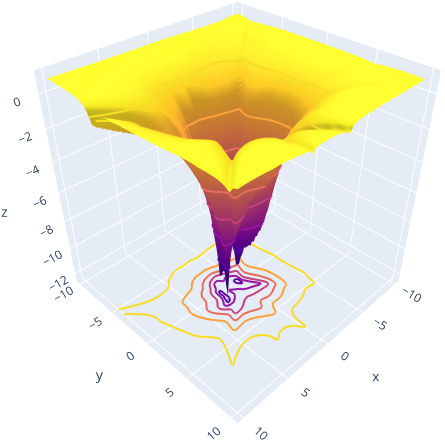}\includegraphics[width = 0.4\textwidth]{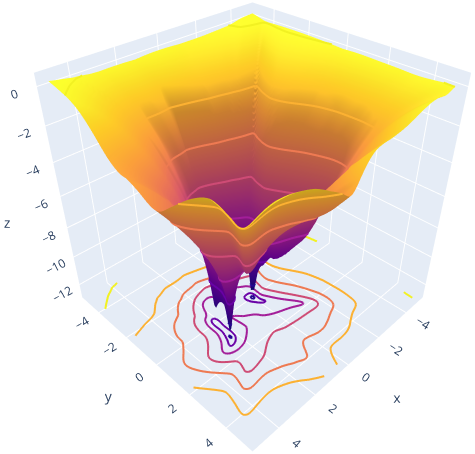}
    
    \includegraphics[width = 0.4\textwidth]{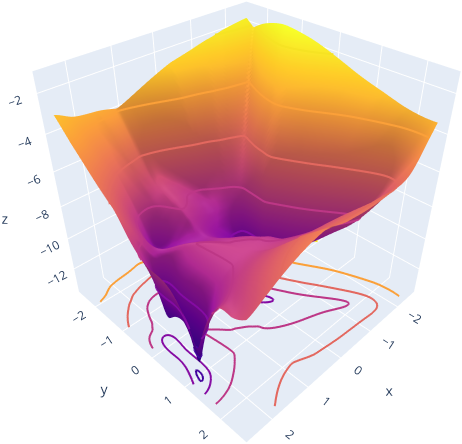}\includegraphics[width = 0.4\textwidth]{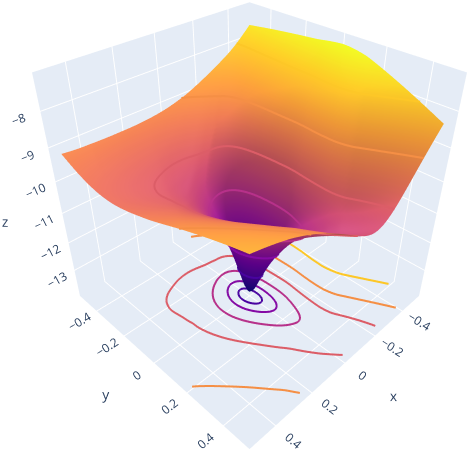}
    \caption{Two-dimensional projection of the loss landscape of a neural network with four layers and ReLU activation function on four different scales. From top-left to bottom-right, we zoom into the global minimum of the landscape.}
    \label{fig:lossLandscapes}
\end{figure}
    
    Given this non-convexity, Algorithm~\ref{alg:1} faces serious challenges. Firstly, there may exist multiple suboptimal local minima. Secondly, the objective may exhibit saddle points, some of which may be of higher order, i.e., the Hessian vanishes. Finally, even if no suboptimal local minima exist, there may be extensive areas of the parameter space where the gradient is very small, so that escaping these regions can take a very long time. 
    
    These issues are not mere theoretical possibilities, but will almost certainly arise. 
    For example,~\cite{NIPS1995_3806734b, safran2018spurious} show the existence of many suboptimal local minima in typical learning tasks. Moreover, for fixed-sized NNs, it has been shown in~\cite{berner2019degenerate, petersen2020topological}, that with respect to $L^p$-norms the set of NNs is generally a very non-convex and non-closed set. Also, the map $\theta \mapsto \Phi_\arch(\cdot, \theta)$ is not a quotient map, i.e., not continuously invertible when accounting for its non-injectivity. 
    In addition, in various situations finding the global optimum of the minimization problem is shown to be NP-hard in general~\cite{blum1989training, judd1990neural, sima2002training}. In Figure~\ref{fig:lossLandscapes} we show the two-dimensional projection of a loss landscape, i.e., the projection of the graph of the function $\theta\mapsto \widehat \cR_\cs(\Phi(\cdot,\theta))$. It is apparent from the visualization that the problem exhibits more than one minimum. We also want to add that in practice one neglects that the loss is only almost everywhere differentiable in case of piecewise smooth activation functions, such as the ReLU, although one could resort to subgradient methods~\cite{kakade2018provably}. 
    
    In view of these considerations, the classical framework presented in Subsection~\ref{subsubsec:optimization} offers no explanation as to why deep learning works in practice. Indeed, in the survey~\cite[Section 1.4]{orr1998neural} the state of the art in 1998 was summarized by the following assessment: \enquote{There is no formula to guarantee that (1) the NN will converge to a good solution, (2) convergence is swift, or (3) convergence even occurs at all.}
    
    Nonetheless, in applications, not only would an explanation of when and why SGD converges be extremely desirable, convergence is also quite often observed even though there is little theoretical explanation for it in the classical set-up. In Section~\ref{sec:optimizationNew}, we collect modern approaches explaining why and when convergence occurs and can be guaranteed.
    
        \paragraph{Which aspects of a neural network architecture affect the performance of deep learning?}
    
        In the introduction to classical approaches to deep learning above, we have seen that in classical results, such as in Theorem~\ref{thm:MhaskarForPresidentOfNeuralNetworkLand}, only the effect of few aspects of the NN architectures are considered. In Theorem~\ref{thm:MhaskarForPresidentOfNeuralNetworkLand} only the impact of the width of the NN was studied. In further approximation theorems below, e.g., in Theorems~\ref{thm:rademacherbound} and~\ref{theo:approx_smooth}, we will additionally have a variable depth of NNs. However, for deeper architectures, there are many additional aspects of the architecture that could potentially affect the performance of the model for the associated learning task. For example, even for a standard FC NN with $L$ layers as in Definition~\ref{def:classical_nns}, there is a lot of flexibility in choosing the number of neurons $(N_1, \dots, N_{L-1}) \in \N^{L-1}$ in the hidden layers. One would expect that certain choices affect the capabilities of the NNs considerably and some choices are preferable over others. Note that, one aspect of the neural network architecture that can have a profound effect on the performance, especially regarding approximation theoretical aspects of the performance, is the choice of the activation function. For example, in \cite{maiorov1999lower, yarotsky2021elementary} activation functions were found that allow uniform approximation of continuous functions to arbitrary accuracy with fixed-size neural networks. In the sequel we will, however, focus on architectural aspects other than the activation function. 
        
        In addition, practitioners have invented an immense variety of NN architectures for specific problems. These include NNs with convolutional blocks~\cite{lecun1998gradient}, with skip connections~\cite{he2016deep}, sparse connections~\cite{zhou2016less, bourely2017sparse}, batch normalization blocks~\cite{ioffe2015batch}, and many more. In addition, for sequential data, recurrent connections are used~\cite{rumelhart1986learning} and these often have forget mechanisms~\cite{hochreiter1997long} or other gates~\cite{cho2014learning} included in their architectures.
        
        The choice of an appropriate NN architecture is essential to the success of many deep learning tasks. This goes so far, that frequently an architecture search is applied to find the most suitable one~\cite{zoph2017neural, pham2018efficient}. In most cases, though, the design and choice of the architecture is based on the intuition of the practitioner.
        
        Naturally, from a theoretical point of view, this situation is not satisfactory. Instead, it would be highly desirable to have a mathematical theory guiding the choice of NN architectures. More concretely, one would wish for mathematical theorems that identify those architectures that work for a specific problem and those that will yield suboptimal results. In Section~\ref{sec:architectures}, we discuss various results that explain theoretically quantifiable effects of certain aspects or building blocks of NN architectures.

    \paragraph{Which features of data are learned by deep architectures?}
    
    It is commonly believed that the neurons of NNs constitute feature extractors in different levels of abstraction that correspond to the layers. This belief is partially grounded in experimental evidence as well as in drawing connections to the human visual cortex, see~\cite[Chapter 9.10]{goodfellow2016deep}. 
        
    Understanding the features that are learned can, in a way, be linked to understanding the reasoning with which a NN-based model ended up with its result. Therefore, analyzing the features that a NN learns constitutes a data-aware approach to understanding deep learning. Naturally, this falls outside of the scope of the classical theory, which is formulated in terms of optimization, generalization, and approximation errors.  

    One central obstacle towards understanding these features theoretically is that, at least for practical problems, the data distribution is unknown. However, one often has partial knowledge. One example is that in image classification it appears reasonable to assume that any classifier is translation and rotation invariant as well as invariant under small deformations. In this context, it is interesting to understand under which conditions trained NNs admit the same invariances.
    
    Biological NNs such as the visual cortex are believed to be evolved in a way that is based on sparse multiscale representations of visual information~\cite{olshausen1996sparse}. Again, a fascinating question is whether NNs trained in practice can be shown to favor such multiscale representations based on sparsity or if the architecture is theoretically linked to sparse representations.
    We will discuss various approaches studying the features learned by neural networks in Section~\ref{sec:features}.
    
    \paragraph{Are neural networks capable of replacing highly specialized numerical algorithms in natural sciences?}
    
    Shortly after their successes in various data-driven tasks in data science and AI applications, NNs have been used also as a numerical ansatz for solving highly complex models from the natural sciences which may be combined with data driven methods. This per se is not very surprising as many such models can be formulated as optimization problems where the common deep learning paradigm can be directly applied. What might be considered surprising is that this approach seems to be applicable to a wide range of problems which have previously been tackled by highly specialized numerical methods. 
    
    Particular successes include the data-driven solution of ill-posed \emph{inverse problems}~\cite{arridge2019solving} which have, for example, led to a fourfold speedup in MRI scantimes~\cite{zbontar2018fastMRI} igniting the research project~\url{fastmri.org}. Deep-learning-based approaches have also been very successful in solving a vast array of different \emph{partial differential equation} (PDE) models, especially in the high-dimensional regime~\cite{weinan2018deep, raissi2019physics,hermann2020deep,pfau2020ab} where most other methods would suffer from the curse of dimensionality. 
    
    Despite these encouraging applications, the foundational mechanisms governing their workings and limitations are still not well understood. In Subsection~\ref{subsec:PDEapprox} and Section~\ref{sec:effectiveness} we discuss some theoretical and practical aspects of deep learning methods applied to the solution of inverse problems and PDEs.

\section{Generalization of large neural networks}
\label{sec:gen}
In the following, we will shed light on the generalization puzzle of NNs as described in Subsection~\ref{subsec:NewTheories}. We focus on four different lines of research which, of course, do not cover the wide range of available results. In fact, we had to omit a discussion of a multitude of important works, some of which we reference in the following paragraph.

First, let us mention extensions of the generalization bounds presented in Subsection~\ref{subsubsec:gen_error} making use of \emph{local} Rademacher complexities~\cite{bartlett2005local} or dropping assumptions on boundedness or rapidly decaying tails~\cite{mendelson2014learning}. Furthermore, there are approaches to generalization which do not focus on the hypothesis set $\cF$, i.e., the range of the learning algorithm $\cA$, but the way $\cA$ chooses its model $f_\cs$. For instance, one can assume that $f_\cs$ does not depend too strongly on each individual sample (\emph{algorithmic stability}~\cite{bousquet2002stability, poggio2004general}), only on a subset of the samples (\emph{compression bounds}~\cite{arora2018stronger}), or satisfies local properties (\emph{algorithmic robustness}~\cite{xu2012robustness}). 
Finally, we refer the reader to~\cite{jiang2020fantastic} and the references mentioned therein for an empirical study of various measures related to generalization.

Note that many results on generalization capabilities of NNs can still only be proven in simplified settings, e.g., for deep linear NNs, i.e., $\varrho(x)=x$, or basic linear models, i.e., one-layer NNs. Thus, we start by emphasizing the connection of deep, nonlinear NNs to linear models (operating on features given by a suitable kernel) in the \emph{infinite width limit}.

\subsection{Kernel regime}
\label{subsec:ntk}
We consider a one-dimensional prediction setting where the loss $\cL(f,(x, y))$ depends on $x\in\cX$ only through $f(x)\in\cY$, i.e., there exists a function $\ell\colon \cY \times \cY \to\R$ such that
\begin{equation*}
    \cL(f,(x,y))=\ell(f(x),y).
\end{equation*}
For instance, in case of the quadratic loss we have that $\ell(\hat{y},y)=(\hat{y}-y)^2$.
Further, let $\Phi$ be a NN with architecture $(N,\varrho)=((d,N_1,\dots,N_{L-1}, 1),\varrho)$ and let $\Theta_0$ be a $\R^{P(N)}$-valued random variable. For simplicity, we evolve the parameters of $\Phi$ according to the continuous version of gradient descent, so-called \emph{gradient flow}, given by
\begin{equation}
\label{eq:gradient_flow}
    \frac{\mathrm{d}\Theta(t)}{\mathrm{d}t} =- \nabla_\theta  \widehat \cR_\cs(\Phi(\cdot, \Theta(t))) =- \frac{1}{m} \sum_{i=1}^m 
    \nabla_\theta \Phi(x^{(i)},\Theta(t)) D_i(t), \quad \Theta(0)=\Theta_0,
\end{equation}
where 
$D_i(t)\coloneqq
\frac{\partial \ell(\hat{y},  y^{(i)})}{\partial \hat{y}}|_{\hat{y} = \Phi(x^{(i)},\Theta(t))}$
is the derivative of the loss with respect to the prediction at input feature $x^{(i)}$ at time $t\in[0,\infty)$.
The chain rule implies the following dynamics of the NN realization
\begin{equation}
\label{eq:pred_flow}
    \frac{\mathrm{d} \Phi({\cdot},\Theta(t))}{\mathrm{d}t} = 
    - \frac{1}{m} \sum_{i=1}^m  
    K_{\Theta(t)}({\cdot}, x^{(i)}) D_i(t)
\end{equation}
and its empirical risk
\begin{equation}
\label{eq:risk_flow}
    \frac{\mathrm{d}\widehat\cR_\cs (\Phi(\cdot,\Theta(t))}{\mathrm{d}t} = - \frac{1}{m^2} \sum_{i=1}^m  \sum_{j=1}^m D_i(t)
    K_{\Theta(t)}(x^{(i)}, x^{(j)})D_j(t),
\end{equation}
where $K_\theta$, $\theta\in\R^{P(N)}$, 
is the so-called \emph{neural tangent kernel} (NTK)
\begin{equation}
\label{eq:ntk}
    K_\theta\colon \R^d\times\R^d\to\R, \quad K_\theta(x_1,x_2) = \big(\nabla_\theta \Phi(x_1,\theta)\big)^T \nabla_\theta \Phi(x_2,\theta).
\end{equation}
Now let $\sigma_w, \sigma_b\in (0,\infty)$ and assume that the initialization $\Theta_0$ consists of independent entries, where entries corresponding to the weight matrix and bias vector in the $\ell$-th layer follow a normal distribution with zero mean and variances $\sigma_w^2/N_\ell$ and $\sigma_b^2$, respectively.
Under weak assumptions on the activation function, the central limit theorem implies that the pre-activations converge to i.i.d.\@ centered Gaussian processes in the infinite width limit $N_1, \dots, N_{L-1}\to \infty$, see~\cite{lee2018deep,matthews2018gaussian}. Similarly, also $K_{\Theta_0}$ converges to a deterministic kernel $K^\infty$ which stays constant in time and only depends on the activation function $\varrho$, the depth $L$, and the initialization parameters $\sigma_w$ and $\sigma_b$~\cite{jacot2018neural,arora2019exact, yang2019scaling, lee2020wide}. Thus, within the infinite width limit, gradient flow on the NN parameters
as in~\eqref{eq:gradient_flow} is equivalent to functional gradient flow in the \emph{reproducing kernel Hilbert space} $(\mathcal{H}_{K^{\infty}},\|\cdot\|_{K^\infty})$ corresponding to $K^{\infty}$, see~\eqref{eq:pred_flow}. 

By~\eqref{eq:risk_flow}, the empirical risk converges to a global minimum as long as the kernel evaluated at the input features, $\bar{K}^\infty \coloneqq(K^\infty(x^{(i)}, x^{(j)}))_{i,j=1}^m\in\R^{m \times m}$, is positive definite (see, e.g.,~\cite{jacot2018neural, du2019gradient} for suitable conditions) and the $\ell(\cdot, y^{(i)})$ are convex and lower bounded. For instance, in case of the quadratic loss
the solution of~\eqref{eq:pred_flow} is then given by 
\begin{equation}
\label{eq:evolution_quadratic}
    \Phi({\cdot},\Theta(t)) = C(t)(y^{(i)})_{i=1}^m + \big(\Phi(\cdot,\Theta_0) - C(t)(\Phi(x^{(i)},\Theta_0))_{i=1}^m\big),
\end{equation}
where $C(t):=\big((K^\infty(\cdot,x^{(i)}))_{i=1}^m\big)^T (\bar{K}^\infty)^{-1} (\mathrm{I}_m-e^{-\frac{2\bar{K}^\infty t}{m}})$. As the initial realization $\Phi(\cdot,\Theta_0)$ constitutes a centered Gaussian process, the second term in~\eqref{eq:evolution_quadratic} follows a normal distribution with zero mean at each input. In the limit $t\to \infty$, its variance vanishes on the input features $x^{(i)}$, $i\in [m]$, and the first term convergences to the minimum kernel-norm interpolator, i.e., to the solution of
\begin{equation*}
    \min_{f\in \mathcal{H}_{K^\infty}} \|f\|_{K^\infty} \quad \text{s.t.} \quad f(x^{(i)})=y^{(i)}.
\end{equation*}
Therefore, within the infinite width limit, the generalization properties of the NN could be described by the generalization properties of the minimizer in the reproducing kernel Hilbert space corresponding to the kernel $K^\infty$~\cite{belkin2018understand,liang2020just,liang2020multiple,ghorbani2021linearized,li2021generalization}.

This so-called \emph{lazy training}, where a NN essentially behaves like a linear model with respect to the nonlinear features $x\mapsto  \nabla_\theta \Phi(x,\theta)$, can already be observed in the non-asymptotic regime, see also Subsection~\ref{subsec:lazy_training}. For sufficiently overparametrized ($P(N)\gg m$) and suitably initialized models, one can show that $K_{\theta(0)}$ is close to $K^{\infty}$ at initialization and $K_{\theta(t)}$ stays close to $K_{\theta(0)}$ throughout training, see~\cite{du2018gradient,arora2019exact,chizat2019lazy,du2019gradient}. The dynamics of the NN under gradient flow in~\eqref{eq:pred_flow} and~\eqref{eq:risk_flow} can thus be approximated by the dynamics of the linearization of $\Phi$ at initialization $\Theta_0$, given by 
\begin{equation}
\label{eq:linearization}
     \Phi^{\mathrm{lin}}(\cdot, \theta)\coloneqq\Phi(\cdot,\Theta_0) + \langle \nabla_\theta \Phi(\cdot,\Theta_0), \theta - \Theta_0\rangle,
\end{equation}
which motivates to study the behavior of linear models in the overparametrized regime.

\subsection{Norm-based bounds and margin theory} 
\label{subsec:norm_margin}
For piecewise linear activation functions, one can improve upon the VC-dimension bounds in Theorem~\ref{thm:nn_vc} and show that, up to logarithmic factors, the VC-dimension is asymptotically bounded both above and below by $P(N)L$, see~\cite{bartlett2019nearly}. The lower bound shows that the generalization bound in Theorem~\ref{thm:vc_gen_bound} can only be non-vacuous if the number of samples $m$ scales at least linearly with the number of NN parameters $P(N)$. However, heavily overparametrized NNs used in practice seem to generalize well outside of this regime. 

One solution is to bound other complexity measures of NNs taking into account various norms on the parameters and avoid the direct dependence on the number of parameters~\cite{bartlett1998sample}.
For instance, we can compute bounds on the Rademacher complexity of NNs with positively homogeneous activation function, where the Frobenius norm of the weight matrices is bounded, see also~\cite{neyshabur2015norm}. Note that, for instance, the ReLU activation is positively homogeneous, i.e., it satisfies that $\ReLU(\lambda x)=\lambda \ReLU(x)$ for all $x\in\R$ and $\lambda \in (0,\infty)$.
\begin{theorem}[Rademacher complexity of neural networks]\label{thm:rademacherbound}
	Let $d\in\N$, assume that $\cX=B_1(0)\subset \R^d$, and let $\varrho$ be a positively homogeneous activation function with Lipschitz constant $1$. We define the set of all biasless NN realizations with depth $L\in\N$, output dimension $1$, and Frobenius norm of the weight matrices bounded by $C\in(0,\infty)$ as
	\begin{equation*}
	    \widetilde{\cF}_{L,C}\coloneqq\big\{\Phi_{(N,\varrho)}(\cdot, \theta)\colon N\in\N^{L+1}, \ N_0=d, \ N_L=1, \ \theta =((W^{(\ell)},0))_{\ell=1}^L \in \R^{P(N)},
	    \ \|W^{(\ell)}\|_F \le C\big\}.
	\end{equation*}
	Then for every $m\in\N$ it holds that
	\begin{equation*}
		\fR_m(\widetilde{\cF}_{L,C})\le \frac{C(2C)^{L-1} }{\sqrt{m}}.
	\end{equation*}
\end{theorem}

The term $2^{L-1}$ depending exponentially on the depth can be reduced to $\sqrt{L}$ or completely omitted by invoking also the spectral norm of the weight matrices~\cite{golowich2018size}.
Further, observe that for $L=1$, i.e., linear classifiers with bounded Euclidean norm, this bound is independent of the input dimension $d$. Together with~\eqref{eq:symmetrization}, this motivates why the regularized linear model in Figure~\ref{fig:tradeoff} did perform well in the overparametrized regime.

The proof of Theorem~\ref{thm:rademacherbound} is based on the contraction property of the Rademacher complexity~\cite{ledoux1991probability} which establishes that
\begin{equation*}
	\fR_m(\varrho\circ\widetilde{\cF}_{\ell,C}) \le 2 \fR_m(\widetilde{\cF}_{\ell,C}), \quad \ell\in\N.
\end{equation*}
We can iterate this together with the fact that for every $\tau\in\{-1,1\}^m$, and $x\in\R^{N_{\ell-1}}$ it holds that
\begin{equation*}
    \sup_{\|W^{(\ell)}\|_F\le C} \big\|\sum_{i=1}^m \tau_i  \varrho(W^{(\ell)}x)\big\|_2 = C  \sup_{\|w\|_2\le 1} \big|\sum_{i=1}^m \tau_i \varrho(\langle w, x\rangle)\big|.
\end{equation*}
In summary, one establishes that
\begin{equation*}
    \fR_m(\widetilde{\cF}_{L,C}) = \frac{C}{m} \E\big[\sup_{f\in \widetilde{\cF}_{L-1,C}} \big\| \sum_{i=1}^{m} \tau_i \varrho(f(X^{(i)}))\big\|_2\big] \le \frac{C(2C)^{L-1}}{m} \E\big[\big\| \sum_{i=1}^{m} \tau_i X^{(i)}\big\|_2\big],
\end{equation*}
which by Jensen's inequality yields the claim.

Recall that for classification problems one typically minimizes a surrogate loss $\cL^{\mathrm{surr}}$, see Remark~\ref{rem:surrogate_loss}. This suggests that there could be a trade-off happening between complexity of the hypothesis class $\cF_{\arch}$ and the corresponding regression fit underneath, i.e., the \emph{margin} $M(f,z)\coloneqq yf(x)$
by which a training example $z=(x,y)$ has been classified correctly by $f\in\cF_{\arch}$, see~\cite{bartlett2017spectrally, neyshabur2018pac, jiang2019predicting}. 
For simplicity, let us focus on the ramp surrogate loss with confidence $\gamma > 0$, i.e., $ \cL^{\mathrm{surr}}_\gamma(f,z) \coloneqq \ell_\gamma(M(f,z))$, where
\begin{equation*}
    \ell_\gamma(t)\coloneqq \mathds{1}_{(-\infty,\gamma]}(t) - \tfrac{t}{\gamma} \mathds{1}_{[0,\gamma]}(t), \quad t\in\R.
\end{equation*}
Note that the ramp function $\ell_\gamma$ is $1/\gamma$-Lipschitz continuous. Using McDiarmid's inequality and a symmetrization argument similar to the proof of Theorem~\ref{thm:vc_gen_bound}, combined with the contraction property of the Rademacher complexity, yields the following bound on the probability of misclassification:
With probability $1-\delta$ for every $f\in\cF_{\arch}$ it holds that
\begin{equation*}
\begin{split}
    \P[\sgn(f(X)) \neq Y] \le \E\big[\cL^{\mathrm{surr}}_\gamma(f,Z)\big] &\lesssim \frac{1}{m} \sum_{i=1}^m  \cL^{\mathrm{surr}}_\gamma(f,Z^{(i)}) + \fR_m(\cL^{\mathrm{surr}}_\gamma \circ \cF_{\arch}) + \sqrt{\frac{\ln(1/\delta)}{m}} \\&\lesssim \frac{1}{m} \sum_{i=1}^m \mathds{1}_{(-\infty,\gamma)}(Y^{(i)}f(X^{(i)})) + \frac{\fR_m(M\circ\cF_\arch)}{\gamma} +\sqrt{\frac{\ln(1/\delta)}{m}}
    \\& = \frac{1}{m} \sum_{i=1}^m \mathds{1}_{(-\infty,\gamma)}(Y^{(i)}f(X^{(i)})) + \frac{\fR_m(\cF_{\arch})}{\gamma} +\sqrt{\frac{\ln(1/\delta)}{m}}.
\end{split}
\end{equation*}
This shows the trade-off between the complexity of $\cF_\arch$ measured by $\fR_m(\cF_\arch)$ and the fraction of training data that has been classified correctly with a margin of at least $\gamma$. In particular this suggests, that (even if we classify the training data correctly with respect to the $0$-$1$ loss) it might be beneficial to further increase the complexity of $\cF_\arch$ to simultaneously increase the margins by which the training data has been classified correctly and thus obtain a better generalization bound.

\subsection{Optimization and implicit regularization} 
The optimization algorithm, which is usually a variant of SGD, seems to play an important role for the generalization performance. Potential indicators for good generalization performance are high speed of convergence~\cite{hardt2016train} or flatness of the local minimum to which SGD converged, which can be characterized by the magnitude of the eigenvalues of the Hessian (or approximately as the robustness of the minimizer to adversarial perturbations on the parameter space), see~\cite{keskar2017large}.
In~\cite{dziugaite2017computing, neyshabur2017exploring} generalization bounds depending on a concept of flatness are established by employing a PAC-Bayesian framework, which can be viewed as a generalization of Theorem~\ref{thm:gen_finite}, see~\cite{mcallester1999pac}. Further, one can also unite flatness and norm-based bounds by the \emph{Fisher--Rao metric} of information geometry~\cite{liang2019fisher}.

Let us motivate the link between generalization and flatness in the case of simple linear models:
We assume that our model takes the form $\langle \theta, \cdot \rangle$, $\theta\in \R^d$, and we will use the abbreviations 
\begin{equation*}
    r(\theta) \coloneqq\widehat \cR_s(\langle \theta, \cdot \rangle) \quad \text{and} \quad  \gamma(\theta)\coloneqq \min_{i\in[m]} M(\langle \theta, \cdot \rangle, z^{(i)})=\min_{i\in[m]} y^{(i)}\langle \theta, x^{(i)} \rangle
\end{equation*}
throughout this subsection to denote the empirical risk and the margin for given training data $\cs=((x^{(i)},y^{(i)}))_{i=1}^m$. 
We assume that we are solving a classification task with the $0$-$1$ loss and that our training data is linearly separable. This means that there exists a minimizer $\hat \theta\in\R^d$ such that $r(\hat \theta) = 0$.
We observe that $\delta$-robustness in the sense that
\begin{equation*}
    \max_{\theta \in B_\delta(0)} r( \hat \theta + \theta) = r(\hat \theta) = 0
\end{equation*}
implies that
\begin{equation*}
    0 < \min_{i\in [m]} y^{(i)}\langle \hat{\theta}-\delta y^{(i)}\tfrac{x^{(i)}}{\|x^{(i)}\|_2}, x^{(i)} \rangle \le \gamma(\hat \theta) -\delta \min_{i\in [m]} \|x^{(i)}\|_2 ,
\end{equation*}
see also~\cite{poggio2017theory}. This lower bound on the margin $\gamma(\hat \theta)$ then ensures generalization guarantees as described in Subsection~\ref{subsec:norm_margin}. 

Even without explicit\footnote{Note that also different architectures can exhibit vastly different inductive biases~\cite{zhang2020identity} and also within the architecture different parameters have different importance, see~\cite{frankle2018lottery, zhang2019all} and Proposition~\ref{prop:CaratheordoryPruning}.} control on the complexity of $\cF_\arch$, there do exist results showing that SGD acts as implicit regularization~\cite{neyshabur2014search}. This is motivated by linear models where SGD converges to the minimal Euclidean norm solution for the quadratic loss and in the direction of the hard margin support vector machine solution for the logistic loss on linearly separable data~\cite{soudry2018implicit}. Note that convergence to minimum norm or maximum margin solutions in particular decreases the complexity of our hypothesis set and thus improves generalization bounds, see Subsection~\ref{subsec:norm_margin}. 

While we have seen this behavior of gradient descent for linear regression already in the more general context of kernel regression in Subsection~\ref{subsec:ntk}, we want to motivate the corresponding result for classification tasks in the following. We focus on the exponential surrogate loss $\cL^{\mathrm{surr}}(f, z)=\ell(M(f,z))=e^{-yf(x)}$ with $\ell(z)=e^{-z}$, but similar observations can be made for the logistic loss defined in Remark~\ref{rem:surrogate_loss}.
We assume that the training data is linearly separable, which guarantees the existence of $\hat \theta \neq 0$ with $\gamma(\hat \theta)> 0$. Then for every linear model $\langle \theta, \cdot \rangle$, $\theta \in \R^d$, it holds that
\begin{equation*}
    \big\langle \hat{\theta}, \nabla_\theta r(\theta) \rangle = \frac{1}{m} \sum_{i=1}^m \underbrace{\ell'( y^{(i)}\langle \theta, x^{(i)}\rangle)}_{<0} \underbrace{y^{(i)}\langle \hat{\theta}, x^{(i)} \rangle}_{>0}.
\end{equation*}
A critical point $\nabla_\theta r(\theta) = 0$ can therefore be approached if and only if for every $i\in[m]$ we have
\begin{equation*}
    \ell'(y^{(i)}\langle \theta, x^{(i)}\rangle)= {-e^{-y^{(i)} \langle \theta, x^{(i)}\rangle}}\to 0,
\end{equation*}
which is equivalent to $\|\theta\|_2\to\infty$ and 
$\gamma(\theta)>0$.
Let us now define 
\begin{equation*}
    r_\beta(\theta) \coloneqq \frac{\ell^{-1}(r(\beta\theta))}{\beta}, \quad \theta\in\R^d, \ \beta \in (0,\infty),
\end{equation*}
and observe that 
\begin{equation}
\label{eq:smooth_margin}
    r_\beta(\theta)=-\frac{\log(r(\beta \theta))}{\beta} \to \gamma(\theta),
    \quad \beta \to \infty.
\end{equation}
Due to this property, $r_\beta$ is often referred to as the \emph{smoothed margin}~\cite{lyu2019gradient, ji2019refined}.
We evolve $\theta$ according to gradient flow with respect to the smoothed margin $r_1$, i.e.,
\begin{equation*}
    \frac{\mathrm{d}\theta(t)}{\mathrm{d}t} = \nabla_\theta r_1(\theta(t))=-\frac{1}{r(\theta(t))} \nabla_\theta r(\theta(t)),
\end{equation*}
which produces the same trajectory as gradient flow with respect to the empirical risk $r$ under a rescaling of the time $t$. Looking at the evolution of the normalized parameters $\tilde{\theta}(t)=\theta(t)/\|\theta(t)\|_2$, the chain rule establishes that
\begin{equation*}
    \frac{\mathrm{d}\tilde{\theta}(t)}{\mathrm{d}t} =
    P_{\tilde{\theta}(t)}\frac{\nabla_{\theta} r_{\beta(t)}(\tilde{\theta}(t))}{\beta(t)} \quad \text{with} \quad  \beta(t)\coloneqq\|\theta(t)\|_2 \quad \text{and} \quad P_{\theta} \coloneqq \mathrm{I}_d - \theta \theta^T, \quad \theta \in \R^d.
\end{equation*}
This shows that the normalized parameters perform projected gradient ascent with respect to the function $r_{\beta(t)}$, which converges to the margin due to~\eqref{eq:smooth_margin} and the fact that $\beta(t)=\|\theta(t)\|_2\to\infty$ when approaching a critical point. 
This motivates that during gradient flow the normalized parameters implicitly maximize the margin. See~\cite{gunasekar2018characterizing, gunasekar2018implicit,  lyu2019gradient, nacson2019convergence, chizat2020implicit, ji2020directional} for a precise analysis and various extensions, e.g., to homogeneous or two-layer NNs and other optimization geometries.

To illustrate one research direction, we present an exemplary result in the following. Let $\Phi=\Phi_{(N,\varrho)}$ be a biasless NN with parameters $\theta=((W^{(\ell)},0))_{\ell=0}^L$ and output dimension $N_L=1$. For given input features $x\in\R^{N_0}$, the gradient $\nabla_{W^{(\ell)}}\Phi=\nabla_{W^{(\ell)}}\Phi(x,\theta)\in \R^{N_{\ell-1}\times N_{\ell}}$ with respect to the weight matrix in the $\ell$-th layer satisfies that
\begin{equation*}
    \nabla_{W^{(\ell)}}\Phi =\varrho(\Phi^{(\ell-1)}) \frac{\partial \Phi}{\partial \Phi^{(\ell+1)}} \frac{\partial \Phi^{(\ell+1)}}{\partial \Phi^{(\ell)}} =  \varrho(\Phi^{(\ell-1)}) \frac{\partial \Phi}{\partial \Phi^{(\ell+1)}} W^{(\ell+1)} \operatorname{diag}\big(\varrho'(\Phi^{(\ell)})\big),
\end{equation*}
where the pre-activations $(\Phi^{(\ell)})_{\ell=1}^L$ are given as in~\eqref{eq:def_nn_classical}.
Evolving the parameters according to gradient flow as in~\eqref{eq:gradient_flow} and using an activation function $\varrho$ with $\varrho(x)=\varrho'(x)x$, such as the ReLU, this implies that
\begin{equation}
\label{eq:conservation}
    \operatorname{diag}\big(\varrho'(\Phi^{(\ell)})\big) W^{(\ell)}(t)\Big(\frac{{\mathrm{d}W^{(\ell)}(t)}}{\mathrm{d}t}\Big)^T
    = \Big(\frac{{\mathrm{d}W^{(\ell+1)}(t)}}{\mathrm{d}t}\Big)^T W^{(\ell+1)}(t)\operatorname{diag}\big(\varrho'(\Phi^{(\ell)})\big).
\end{equation}
Note that this ensures the conservation of balancedness between weight matrices of adjacent layers, i.e., 
\begin{equation*}
    \frac{\mathrm{d}}{\mathrm{d}t}\big(\|W^{(\ell+1)}(t)\|^2_F-\|W^{(\ell)}(t)\|^2_F\big)=0,
\end{equation*}
see~\cite{du2018algorithmic}. Furthermore, for deep linear NNs, i.e., $\varrho(x)=x$, the property in~\eqref{eq:conservation} implies conservation of alignment of left and right singular spaces of $W^{(\ell)}$ and $W^{(\ell+1)}$.
This can then be used to show implicit preconditioning and convergence of gradient descent~\cite{arora2018optimization, arora2019convergence} and that, under additional assumptions, gradient descent converges to a linear predictor that is aligned with the maximum margin solution~\cite{ji2019gradient}.
\subsection{Limits of classical theory and double descent} 
\begin{figure}[t]
\centering
\includegraphics[width=0.8\linewidth]{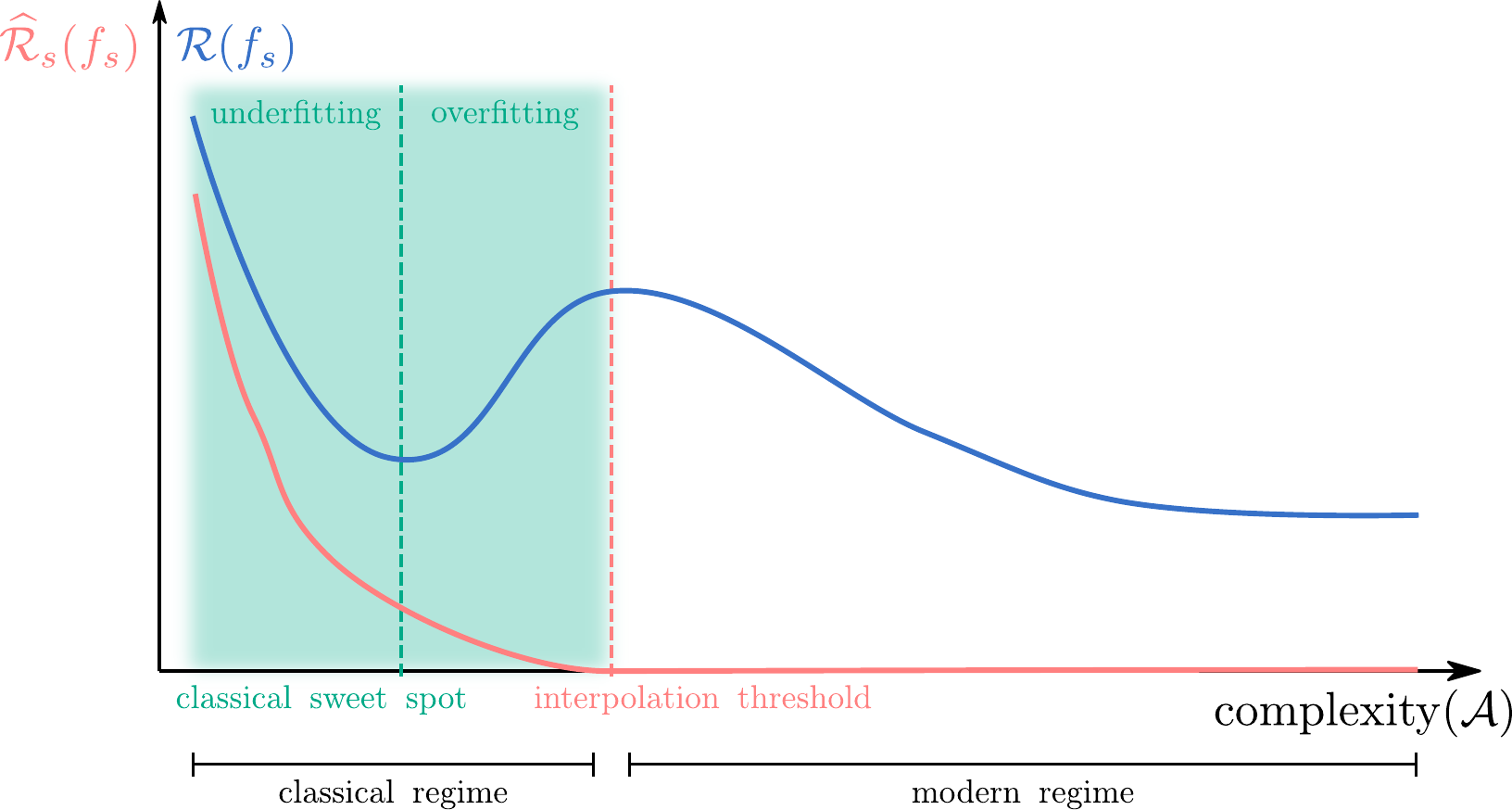}
\caption{This illustration shows the classical, underparametrized regime in green, where the u-shaped curve depicts the bias-variance trade-off as explained in Section~\ref{subsec:foundations_learning}. Starting with complexity of our algorithm $\cA$ larger than the interpolation threshold we can achieve zero empirical risk $\widehat \cR_\cs(f_\cs)$ (training error), where $f_\cs=\mathcal{A}(\cs)$. Within this modern interpolation regime, the risk $\cR(f_\cs)$ (test error) might be even lower than at the classical sweet spot. Whereas $\operatorname{complexity}(\cA)$ traditionally refers to the complexity of the hypothesis set $\cF$, there is evidence that also the optimization scheme and the data is influencing the complexity leading to definitions like $\operatorname{complexity}(\cA)\coloneqq\max\big\{ m\in\N \colon 
\E\big[\widehat \cR_\cS(\cA(\cS)) \big] \le \eps \ \text{with} \ \cS\sim \P_Z^m\big\}$, for suitable $\eps > 0$~\cite{nakkiran2020deep}. 
This illustration is based on~\cite{belkin2019reconciling}.}
\label{fig:double_descent}
\end{figure}

There is ample evidence that classical tools from statistical learning theory alone, such as Rademacher averages, uniform convergence, or algorithmic stability may be unable to explain the full generalization capabilities of NNs~\cite{zhang2016understanding, nagarajan2019uniform}. It is especially hard to reconcile the classical bias-variance trade-off with the observation of good generalization performance when achieving zero empirical risk on noisy data using a regression loss. On top of that, this behavior of overparametrized models in the interpolation regime turns out not to be unique to NNs.
Empirically, one observes for various methods (decision trees, random features, linear models) that the test error decreases even below the sweet-spot in the u-shaped bias-variance curve when further increasing the number of parameters~\cite{belkin2019reconciling, geiger2020scaling, nakkiran2020deep}. This is often referred to as the \emph{double descent curve} or \emph{benign overfitting}, see Figure~\ref{fig:double_descent}.
For special cases, e.g., linear regression or random feature regression, such behavior can even be proven, see~\cite{hastie2019surprises, mei2019generalization, bartlett2020benign, belkin2020two, muthukumar2020harmless}. 

In the following we analyze this phenomenon in the context of linear regression. Specifically, we focus on a prediction task with quadratic loss, input features given by a centered $\R^d$-valued random variable $X$, and labels given by $Y=\langle \theta^*, X \rangle + \nu$, where $\theta^*\in\R^d$ and $\nu$ is a centered random variable independent of $X$.
For training data $\cS=((X^{(i)},Y^{(i)}))_{i=1}^m$, we consider the empirical risk minimizer $\widehat f_\cS=\langle \hat \theta,  \cdot\rangle$ with minimum Euclidean norm of its parameters $\hat \theta$ or, equivalently, the limit of gradient flow with zero initialization. Using~\eqref{eq:regression_risk} and a bias-variance decomposition we can write
\begin{equation*}
\begin{split}
    \E[\cR(\widehat f_\cS)|(X^{(i)})_{i=1}^m] -\cR^* &= \E[ \|\widehat f_\cS - f^*\|_{L^2(\P_X)} | (X^{(i)})_{i=1}^m] \\ &=  (\theta^*)^T P \E[XX^T] P \theta^* + \E[\nu^2] \trace \big(\Sigma^+ \E[XX^T]\big),
\end{split}
\end{equation*}
where $\Sigma\coloneqq\sum_{i=1}^m X^{(i)} (X^{(i)})^T$, $\Sigma^+$ denotes the Moore--Penrose inverse of $\Sigma$, and $P\coloneqq \mathrm{I}_d-\Sigma^+\Sigma$ is the orthogonal projector onto the kernel of $\Sigma$.
For simplicity, we focus on the variance $\trace \big(\Sigma^+ \E[XX^T]\big)$, which can be viewed as setting $\theta^*=0$ and $\E[\nu^2]=1$. Assuming that $X$ has i.i.d.\@ entries with unit variance and bounded fifth moment, the distribution of the eigenvalues of $\frac{1}{m}\Sigma^+$ in the limit $d,m \to \infty$ with $\frac{d}{m}\to \kappa \in (0,\infty)$ can be described via the Marchenko--Pastur law. Therefore, the asymptotic variance can be computed explicitly as
\begin{equation*}
  \trace \big(\Sigma^+ \E[XX^T]\big) 
  \to \frac{1-\max\{1-\kappa,0\}}{|1-\kappa|} \quad \text{for} \quad 
  d,m\to \infty \quad \text{with} \quad \frac{d}{m}\to \kappa, 
\end{equation*}
almost surely, see~\cite{hastie2019surprises}.
This shows that despite interpolating the data we can decrease the risk in the overparametrized regime $\kappa>1$. In the limit $d,m \to \infty$, such benign overfitting can also be shown for more general settings (including lazy training of NNs), some of which even achieve their optimal risk in the overparametrized regime~\cite{mei2019generalization, montanari2020interpolation, lin2021causes}.

For normally distributed input features $X$ such that $\E[XX^T]$ has rank larger than $m$, one can also compute the behavior of the variance in the non-asymptomatic regime~\cite{bartlett2020benign}.
Define
\begin{equation}
    k^*:=\min\{ k\ge 0 \colon   \frac{\sum_{i>k} \lambda_i}{\lambda_{k+1}} \ge cm \},
\end{equation}
where $\lambda_1 \ge \lambda_2 \ge  \dots \ge \lambda_d \ge 0$ are the eigenvalues of $\E[XX^T]$ in decreasing order and $c\in(0,\infty)$ is a universal constant. Assuming that $k^*/m$ is sufficiently small, with high probability it holds that
\begin{equation*}
  \trace \big(\Sigma^+ \E[XX^T]\big) \approx \frac{k^*}{m} + \frac{m \sum_{i>k^*} \lambda_i^2}{(\sum_{i>k^*} \lambda_i)^2}.
\end{equation*}
\begin{wrapfigure}{r}{0.34\textwidth}
\includegraphics[width=\linewidth]{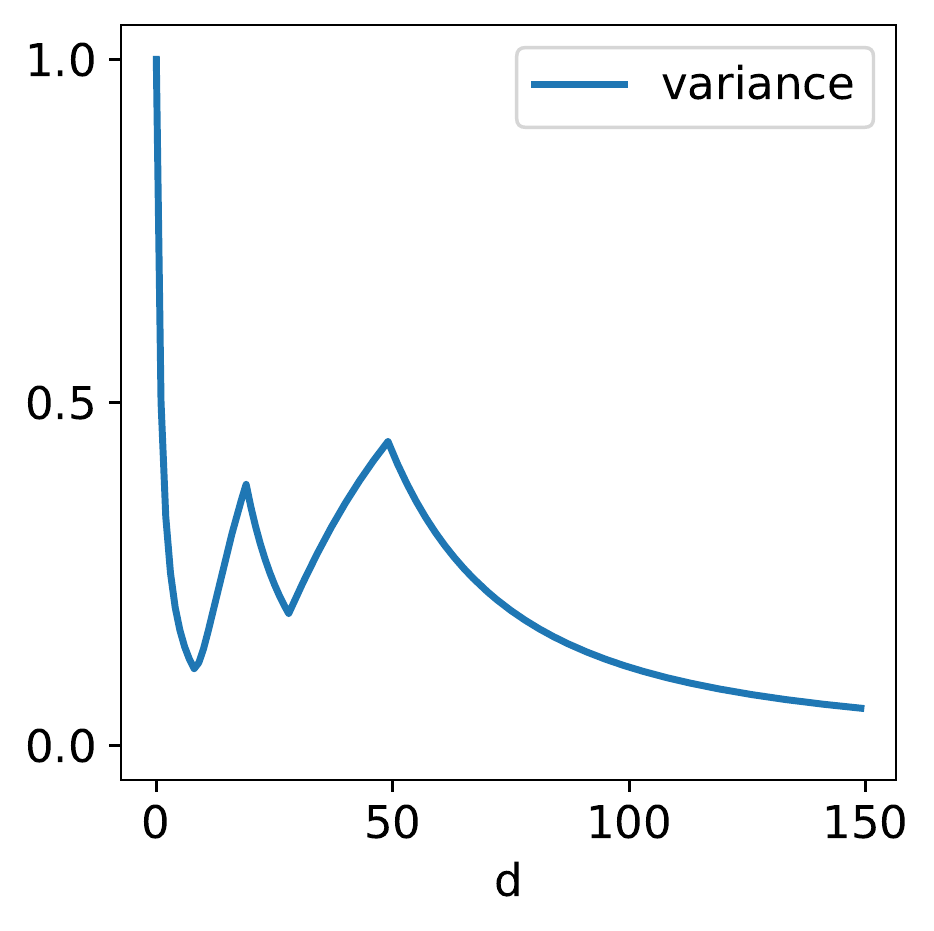}
\caption{The expected variance of linear regression in~\eqref{eq:risk_toy_example} with $d \in[150]$ and $X_i\sim U(\{-1,1\})$, $i\in [150]$, where $X_i=X_1$ for $i\in \{10,\dots,20\}\cup \{30,\dots,50\}$ and all other coordinates are independent.}
\vspace{-1em}
\label{fig:arbitrary_descent}
\end{wrapfigure}
This precisely characterizes the regimes for benign overfitting in terms of the eigenvalues of the covariance matrix $\E[XX^T]$.
Furthermore, it shows that adding new input feature coordinates and thus increasing the number of parameters $d$ can lead to either an increase or decrease of the risk.

To motivate this phenomenon, which is considered in much more depth in~\cite{chen2020multiple}, let us focus on a single sample $m=1$ and 
features $X$ that take values in $\cX=\{-1,1 \}^d$. Then it holds that $\Sigma^+ = \frac{X^{(1)}(X^{(1)})^T}{\|X^{(1)}\|^4} = \frac{X^{(1)}(X^{(1)})^T}{d^2}$ and thus
\begin{equation}
\label{eq:risk_toy_example}
    \E\big[\trace\big(\Sigma^+\E[XX^T]\big)\big] =
    \frac{1}{d^2} \big\|\E\big[ X X^T \big]\big\|_F^2.
\end{equation}
In particular, this shows that incrementing the input feature dimensions $d \mapsto d+1$ one can increase or decrease the risk depending on the correlation of the coordinate $X_{d+1}$ with respect to the previous coordinates $(X_i)_{i=1}^d$, see also Figure~\ref{fig:arbitrary_descent}.

Generally speaking, overparametrization and perfectly fitting noisy data does not exclude good generalization performance, see also~\cite{belkin2019does}. However, the risk crucially depends on the data distribution and the chosen algorithm.

\section{The role of depth in the expressivity of neural networks}\label{sec:expressivity}

The approximation theoretical aspect of a NN architecture, responsible for the approximation component $\eps^{\mathrm{approx}}\coloneqq\cR(f^*_\cF) - \cR^*$ of the error $\cR(f_\cS) - \cR^*$ in~\eqref{eq:errorDecompositionClassical}, is probably one of the most well-studied parts of the deep learning pipe-line. The achievable approximation error of an architecture most directly describes the power of the architecture. 

As mentioned in Subsection~\ref{subsec:NewTheories}, many classical approaches only study the approximation theory of NNs with few layers, whereas modern architectures are typically very deep. A first observation into the effect of depth is that it can often compensate for insufficient width. For example, in the context of the universal approximation theorem, it was shown that very narrow NNs are still universal if instead of increasing the width, the number of layers can be chosen arbitrarily~\cite{hanin2017approximating, hanin2019universal, kidger2020universal}. However, if the width of a NN falls below a critical number, then the universality will not hold any longer.

Below, we discuss three additional observations that shed light on the effect of depth on the approximation capacities or alternative notions of expressivity of NNs. 

\subsection{Approximation of radial functions}\label{subsec:impactOfDepth}

One technique to study the impact of depth relies on the construction of specific functions which can be well approximated by NNs of a certain depth, but require significantly more parameters when approximated to the same accuracy by NNs of smaller depth. In the following we present one example for this type of approach, which can be found in~\cite{eldan2016power}. 
\begin{theorem}[Power of depth]
\label{thm:power_of_depth}
    Let $\varrho\in \{\ReLU, \varrho_\sigma, \mathds{1}_{(0, \infty)}\}$ be the ReLU, the logistic, or the Heaviside function. Then there exist constants $c,C\in(0,\infty)$ with the following property: For every $d\in\N$ with $d\ge C$ there exist a probability measure $\mu$ on $\R^d$, a three-layer NN architecture $\arch=(N,\varrho)=((d,N_1,N_2,1),\varrho)$ with $\|N\|_{\infty}\le Cd^{5}$, and corresponding parameters $\theta^*\in \R^{P(N)}$ with $\|\theta^*\|_{\infty}\le Cd^C$ and $\|\Phi_\arch(\cdot,\theta^*)\|_{L^\infty(\R^d)}\le 2$ such that for every $n\le ce^{cd}$ it holds that
    \begin{equation*}
        \inf_{\theta \in \R^{P((d,n,1))}} \|\Phi_{((d,n,1),\varrho)}(\cdot,\theta)- \Phi_{\arch}(\cdot,\theta^*)\|_{L^2(\mu)} \ge c.
    \end{equation*}
\end{theorem}

In fact, the activation function in Theorem~\ref{thm:power_of_depth} is only required to satisfy mild conditions and the result holds, for instance, also for more general sigmoidal functions.
The proof of Theorem~\ref{thm:power_of_depth} is based on the construction of a suitable radial function $g\colon \R^d \to \R$, i.e., $g(x)=\tilde{g}(\|x\|^2_2)$ for some $\tilde{g}\colon[0,\infty)\to\R$, which can be efficiently approximated by three-layer NNs but approximation by only a two-layer NN requires exponentially large complexity, i.e., the width being exponential in $d$.

The first observation of~\cite{eldan2016power} is that $g$ can typically be well approximated on a bounded domain by a three-layer NN, if $\tilde{g}$ is Lipschitz continuous. Indeed, for the ReLU activation function it is not difficult to show that, emulating a linear interpolation, one can approximate a univariate $C$-Lipschitz function uniformly on $[0,1]$ up to precision $\eps$ by a two-layer architecture of width $\cO(C/\eps)$. The same holds for smooth, non-polynomial activation functions due to Theorem~\ref{thm:MhaskarForPresidentOfNeuralNetworkLand}. This implies that the squared Euclidean norm, as a sum of $d$ univariate functions, i.e., $[0,1]^d \ni x 
\mapsto \sum_{i=1}^d x_i^2$,
can be approximated up to precision $\eps$ by a two-layer architecture of width $\cO(d^2/\eps)$. Moreover, this shows that the third layer can efficiently approximate $\tilde{g}$, establishing approximation of $g$ on a bounded domain up to precision $\eps$ using a three-layer architecture with number of parameters polynomial in $d/\eps$.

The second step in~\cite{eldan2016power} is to choose $g$ in such a way that the realization of any two-layer neural network $\Phi=\Phi_{((d,n,1),\varrho)}(\cdot,\theta)$ with width $n$ not being exponential in $d$ is on average (with respect to the probability measure $\mu$) a constant distance away from $g$. Their argument is heavily based on ideas from Fourier analysis and will be outlined below. In this context, let us recall that we denote by $\hat{f}$ the Fourier transform of a suitable function or, more generally, tempered distribution $f$.

Assuming that the square-root $\varphi$ of the density function associated with the probability measure $\mu$ as well as $\Phi$ and $g$ are well-behaved, the Plancherel theorem yields that
\begin{equation}
\label{eq:mu_approx}
    \|\Phi-g\|_{L^2(\mu)}^2 = \left\|\Phi\varphi - g\varphi\right\|_{L^2(\R^d)}^2 = \big\|\widehat{\Phi\varphi} - \widehat{g\varphi}\big\|_{L^2(\R^d)}^2.
\end{equation}
Next, the specific structure of two-layer NNs is used, which implies that for every $j\in[n]$ there exists $w_j\in\R^d$ with $\|w_j\|_2=1$ and $\varrho_j\colon\R\to\R$ (subsuming the activation function $\varrho$, the norm of $w_j$, and the remaining parameters corresponding to the $j$-th neuron in the hidden layer) such that $\Phi$ is of the form
\begin{wrapfigure}{r}{0.34\textwidth}
\includegraphics[width=\linewidth]{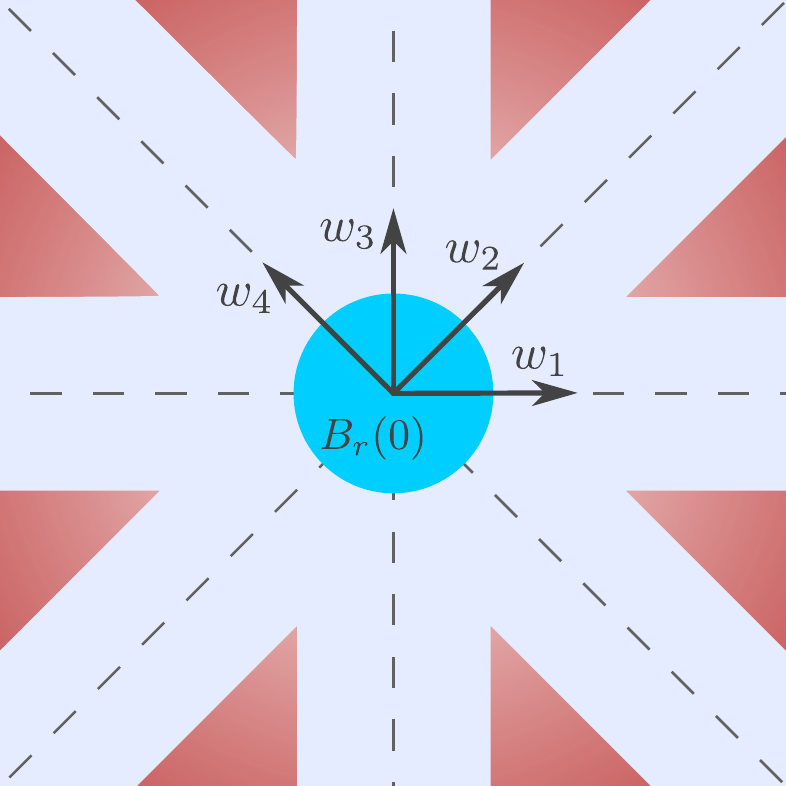}
\caption{This illustration shows the largest possible support (blue) of $\widehat{\Phi\varphi}$, where $\hat{\varphi}=\mathds{1}_{B_r(0)}$ and $\Phi$ is a shallow neural network with architecture $N=(2,4,1)$ and weight matrix $W^{(1)}=[ w_1\dots w_4]^T$ in the first layer. Any radial function with enough of its $L^2$-mass located at high frequencies (indicated by the red area) cannot be well approximated by $\Phi\varphi$.}
\label{fig:power_of_depth}
\end{wrapfigure}
\begin{equation}
\Phi = \sum_{j=1}^n \varrho_j (\langle w_j, \cdot \rangle) = \sum_{j=1}^n ( \varrho_j \otimes \mathds{1}_{\R^{d-1}} ) \circ R_{w_j}.
\end{equation}
The second equality follows by viewing the action of the $j$-th neuron as a tensor product of $ \varrho_j$ and the indicator function $\mathds{1}_{\R^{d-1}}(x)=1$, $x\in\R^{d-1}$, composed with a $d$-dimensional rotation $R_{w_j}\in SO(d)$ which maps $w_j$ to the first standard basis vector $e^{(1)}\in\R^d$.
Noting that the Fourier transform respects linearity, rotations, and tensor products, we can compute
\begin{equation*}
    \hat{\Phi} = \sum_{j=1}^n ( \hat{\varrho}_j \otimes \delta_{\R^{d-1}} ) \circ R_{w_j},
\end{equation*}
where $\delta_{\R^{d-1}}$ denotes the Dirac distribution on $\R^{d-1}$. 
In particular, the support of $\hat{\Phi}$ has a particular star-like shape, namely
$\bigcup_{j=1}^n \spann \{w_j\}$, which are in fact lines passing through the origin.

Now we choose $\varphi$ to be the inverse Fourier transform 
of the indicator function of a ball $B_r(0) \subset \R^d$ with $\operatorname{vol}(B_r(0))=1$,
ensuring that $\varphi^2$ is a valid probability density for $\mu$ as
\begin{equation*}
\begin{split}
  \mu(\R^d)=\|\varphi^2\|_{L^1(\R^d)}=\|\varphi\|^2_{L^2(\R^d)}=\|\hat\varphi\|^2_{L^2(\R^d)}=\|\mathds{1}_{B_r(0)}\|^2_{L^2(\R^d)}=1.
 \end{split}
\end{equation*}
Using the convolution theorem, this choice of $\varphi$ yields that
\begin{equation*}
    \supp(\widehat{\Phi\varphi}) = \supp(\hat{\Phi} * \hat{\varphi}) 
    \subset \bigcup_{j=1}^n \left(\spann \{w_j\} + B_r(0)\right).
\end{equation*}
Thus the lines passing through the origin are enlarged to tubes.
It is this particular shape which allows the construction of some $g$ so that 
$\|\widehat{\Phi\varphi} - \widehat{g\varphi}\|^2_{L^2(\R^d)}$ can be suitably lower bounded, see also Figure~\ref{fig:power_of_depth}.
Intriguingly, the peculiar behavior of high-dimensional 
sets now comes into play. Due to the well known concentration of measure principle, the variable $n$ needs to be exponentially large for the set $\bigcup_{j=1}^n \left(\spann \{w_j\} + B_r(0)\right)$
to be not sparse.
If it is smaller, one can construct a function $g$ so that the main energy content of $\widehat{g\varphi}$ has a certain 
distance from the origin, yielding a lower bound for $\|\widehat{\Phi\varphi} - \widehat{g\varphi}\|^2$ and hence $\|\Phi-g\|_{L^2(\mu)}^2$, see~\eqref{eq:mu_approx}.
One key technical problem is the fact that such a behavior for $\hat{g}$ does not immediately imply a similar behavior
of $\widehat{g\varphi}$, requiring a quite delicate construction of $g$.

\subsection{Deep ReLU networks}\label{subsec:ReLUNNs}
\begin{wrapfigure}{r}{0.34\textwidth}
    \includegraphics[width = \linewidth]{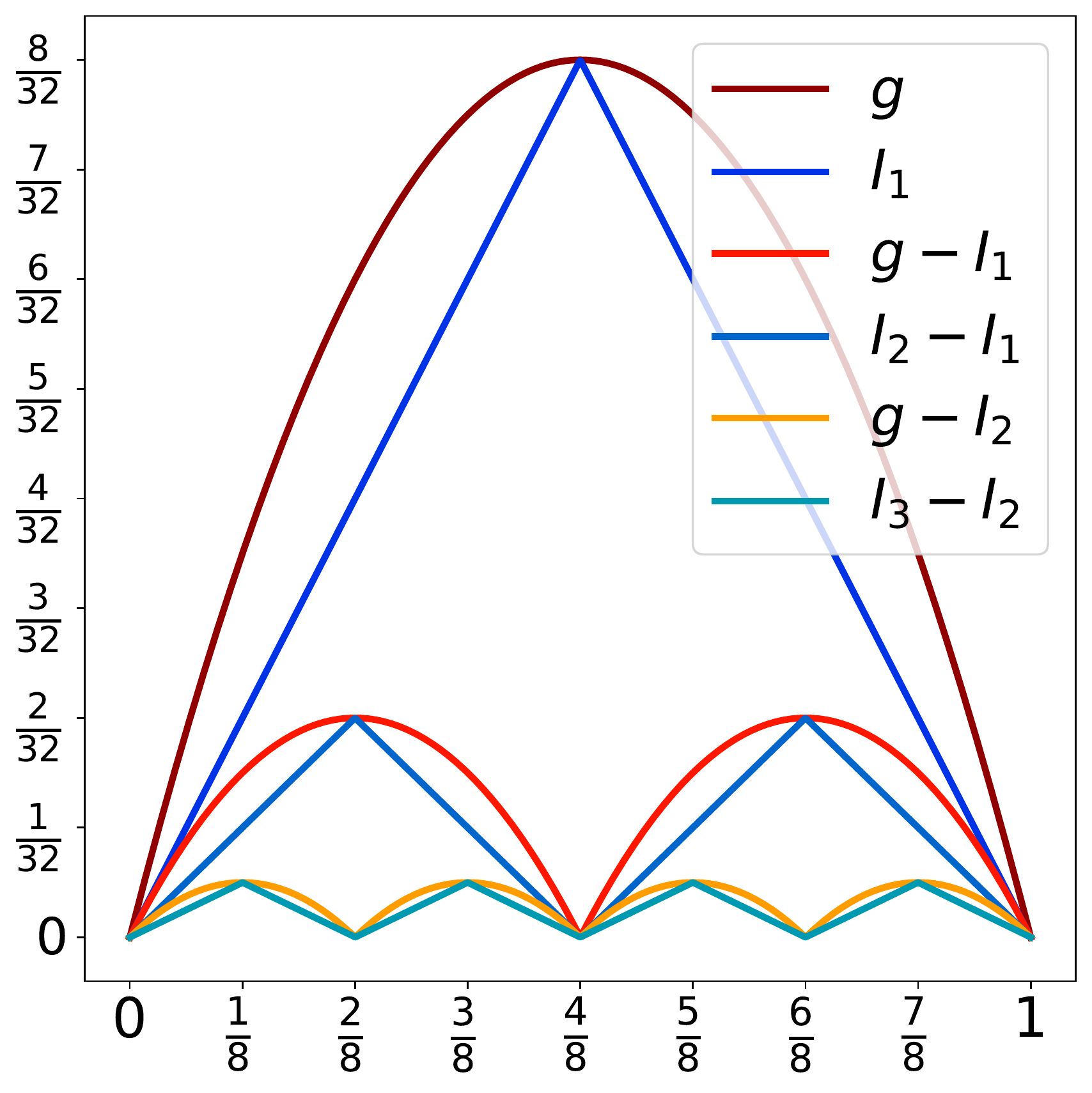}
    \caption{Interpolation $I_n$ of $[0,1]\ni x \mapsto g(x)\coloneqq x-x^2$ on $2^{n}+1$ equidistant points, which can be represented as a sum $I_n= \sum_{k=1}^n I_k - I_{k-1}=\sum_{k=1}^n \tfrac{h_k}{2^{2k}}$ of $n$ sawtooth functions. Each sawtooth function $h_k=h_{k-1}\circ h$ in turn can be written as a $k$-fold composition of a hat function $h$. This illustration is based on~\cite{grohs2019deep}.}
    \vspace{-4em}
    \label{fig:squaring}
\end{wrapfigure}
Maybe for no activation function is the effect of depth clearer than for the ReLU activation function $\ReLU(x)=\max\{0,x\}$. We refer to corresponding NN architectures $(N,\ReLU)$ as \emph{ReLU (neural) networks} (ReLU NNs). 
A two-layer ReLU NN with one-dimensional input and output is a function of the form
\begin{equation*}
    \Phi(x) = \sum_{i=1}^n w_i^{(2)} \ReLU(w_i^{(1)} x + b^{(1)}_i) + b^{(2)}, \quad x \in \R,
\end{equation*}
where $w_i^{(1)}, w_i^{(2)}, b_i^{(1)}, b^{(2)} \in \R$ for $i \in [n]$. 
It is not hard to see that $\Phi$ is a continuous piecewise affine linear function. Moreover, $\Phi$ has at most $n+1$ affine linear pieces. On the other hand, notice that the \emph{hat function}
\begin{equation}
\begin{split}
\label{eq:hat_function}
    h\colon [0,1] &\to [0,1], \\
   x&\mapsto 2\ReLU(x)-4\ReLU(x-\tfrac{1}{2}) =
   \begin{cases}
   2x, &\text{if } 0\le x < \tfrac{1}{2}, \\
   2(1-x), &\text{if } \tfrac{1}{2}\le x \le 1,
   \end{cases}
\end{split}
\end{equation}
is a NN with two layers and two neurons. Telgarsky observed that the $n$-fold convolution $h_n(x)\coloneqq h\circ\dots \circ h$
produces a sawtooth function with $2^n$ spikes~\cite{telgarsky2015representation}. In particular, $h_n$ admits $2^n$ affine linear pieces with only $2n$ many neurons. In this case, we see that deep ReLU NNs are in some sense exponentially more efficient in generating affine linear pieces.

Moreover, it was noted in~\cite{yarotsky2017error} that the difference of interpolations of $[0,1]\ni x\mapsto x-x^2$ at $2^{n}+1$ and $2^{n-1}+1$ equidistant points equals the scaled sawtooth function $\tfrac{h_{n}}{2^{2n}}$, see Figure~\ref{fig:squaring}.
This allows to efficiently implement approximative squaring and, by polarization, also approximative multiplication using ReLU NNs. Composing these simple functions one can approximate localized Taylor polynomials and thus smooth functions, see~\cite{yarotsky2017error}. We state below a generalization~\cite{guhring2020error} of the result of~\cite{yarotsky2017error} which includes more general norms, but for $p = \infty$ and $s = 0$ coincides with the original result of Dmitry Yarotsky.
\begin{theorem}[Approximation of Sobolev-regular functions]\label{theo:approx_smooth}
    Let $d,k\in\N$ with $k\ge2$, let $p\in [1,\infty]$, $s\in[0,1]$, $B\in (0,\infty)$, and let $\varrho$ be a piecewise linear activation function with at least one break-point. Then there exists a constant $c\in(0,\infty)$
    with the following property: For every $\eps\in(0,1/2)$ there exists a NN architecture $\arch=(N,\varrho)$ with 
    \begin{equation*}
        P(N)\le c\eps^{-d/(k-s)} \log(1/\eps)
    \end{equation*}
    such that for every function $g\in W^{k,p}((0,1)^d)$ with $\|g\|_{W^{k,p}((0,1)^d)}\le B$ it holds that
    \begin{equation*}
       \inf_{\theta\in\R^{P(N)}} \|\Phi_\arch(\theta,\cdot) - g\|_{W^{s,p}((0,1)^d)} \le \eps.
    \end{equation*}
\end{theorem}

The ability of deep ReLU neural networks to emulate multiplication has also been employed to reapproximate wide ranges of high-order finite element spaces. In~\cite{opschoorFEM2020} and~\cite{marcati2020exponential} it was shown that deep ReLU neural networks are capable of achieving the approximation rates of $hp$-finite element methods. Concretely, this means that for piecewise analytic functions, which appear, for example, as solutions of elliptic boundary and eigenvalue problems with analytic data, exponential approximation rates can be achieved. In other words, the number of parameters of neural networks to approximate such a function in the $W^{1,2}$-norm up to an error of $\eps$ is logarithmic in $\eps$.

Theorem~\ref{theo:approx_smooth} requires the depth of the NN to grow. In fact, it can be shown that the same approximation rate cannot be achieved with shallow NNs. Indeed, there exists a certain optimal number of layers and, if the architecture has fewer layers than optimal, then the NNs need to have significantly more parameters, to achieve the same approximation fidelity. This has been observed in many different settings in~\cite{liang2017deep, safran2017depth, yarotsky2017error, petersen2018optimal, grohs2019deep}. We state here the result of~\cite{yarotsky2017error}:
\begin{theorem}[Depth-width approximation trade-off]\label{thm:lowerboundpieces}
    Let $d,L\in\N$ with $L \geq 2$ and let $g\in C^2([0,1]^d)$ be a function which is not affine linear. Then there exists a constant $c\in(0,\infty)$ 
    with the following property: 
    For every $\eps\in(0,1)$ and every ReLU NN architecture $\arch=(N,\ReLU)=((d,N_1,\dots,N_{L-1},1),\ReLU)$ with $L$ layers and
    $
        \|N\|_1\le c\eps^{-1/(2(L-1))}
    $
    neurons it holds that
    \begin{equation*}
        \inf_{\theta\in\R^{P(N)}}\|\Phi_\arch(\cdot, \theta) -g\|_{L^\infty([0,1]^d)} \ge \eps.
    \end{equation*}
\end{theorem}

\begin{wrapfigure}{r}{0.5\textwidth}
\includegraphics[width=0.9\linewidth]{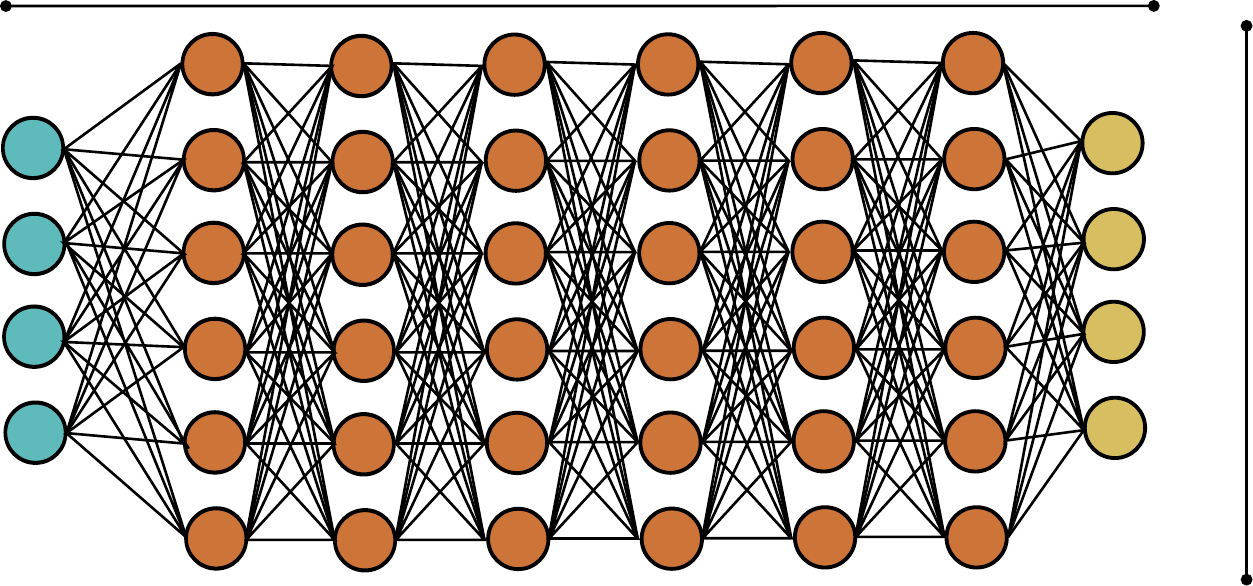}
\put(-124, 103){depth}
\put(5, 63){\rotatebox{270}{width}}
\caption{Standard feed-forward neural network. For certain approximation results, depth and width need to be in a fixed relationship to achieve optimal results.}
\label{fig:DepthWidthTradeOff}
\end{wrapfigure}

This results is based on the observation that ReLU NNs are piecewise affine linear. The number of pieces they admit is linked to their capacity of approximating functions that have non-vanishing curvature. Using a construction similar to the example at the beginning of this subsection, it can be shown that the number of pieces that can be generated using an architecture $((1,N_1,\dots,N_{L-1},1),\ReLU)$ scales roughly like $\prod_{\ell=1}^{L-1} N_\ell$.

In the framework of the aforementioned results, we can speak of a depth-width trade-off, see also Figure~\ref{fig:DepthWidthTradeOff}. A fine-grained estimate of achievable rates for freely varying depths has also been established in~\cite{shen2020deep}.

\subsection{Alternative notions of expressivity}

Conceptual approaches to study the approximation power of deep NNs besides the classical approximation framework usually aim to relate structural properties of the NN to the \enquote{richness} of the set of possibly expressed functions. One early result in this direction is~\cite{montufar2014number} which describes bounds on the number of \emph{affine linear regions} of a ReLU NN $\Phi_{(N,\ReLU)}(\cdot,\theta)$. In a simplified setting, we have seen estimates on the number of affine linear pieces already at the beginning of Subsection~\ref{subsec:ReLUNNs}. Affine linear regions can be defined as the connected components of $\R^{N_0} \setminus H$, where $H$ is the set of non-differentiability of the realization\footnote{One can also study the potentially larger set of \emph{activation regions} given by the connected components of $\R^{N_0}\setminus \big( \cup_{\ell=1}^{L-1} \cup_{i=1}^{N_{\ell}} H_{i,\ell} \big)$, where 
\begin{equation*}
    H_{i,\ell}\coloneqq\{ x\in \R^{N_0} \colon \Phi_i^{(\ell)}(x,\theta)=0\},
\end{equation*}
with $\Phi_i^{(\ell)}$ as in~\eqref{eq:def_nn_classical}, is the set of non-differentiability of the activation of the $i$-th neuron in the $\ell$-th layer. In contrast to the linear regions, the activation regions are necessarily convex~\cite{raghu2017expressivepower,hanin2019deep}.} $\Phi_{(N,\ReLU)}(\cdot,\theta)$. A refined analysis on the number of such regions was, for example, conducted by~\cite{vondeGeer2019}. It is found that deep ReLU neural networks can exhibit significantly more regions than their shallow counterparts. 

The reason for this effectiveness of depth is described by the following analogy: 
Through the ReLU each neuron $\R^d\ni x\mapsto \ReLU(\langle x, w\rangle +b)$, $w\in\R^d$, $b\in\R$, splits the space into two affine linear regions separated by the hyperplane 
\begin{equation}
    \{x\in\R^d \colon \langle x, w\rangle +b=0 \}.
\end{equation}
A shallow ReLU NN $\Phi_{((d,n, 1),\ReLU)}(\cdot,\theta)$ with $n$ neurons in the hidden layer therefore produces a number of regions defined through $n$ hyperplanes. Using classical bounds on the number of regions defined through hyperplane arrangements~\cite{zaslavsky1975facing}, one can bound the number of affine linear regions by $\sum_{j=0}^{d} \binom{n}{j}$. Deepening neural networks then corresponds to a certain folding of the input space. Through this interpretation it can be seen that composing NNs can lead to a multiplication of the number of regions of the individual NNs resulting in an exponential efficiency of deep neural networks in generating affine linear regions\footnote{However, to exploit this efficiency with respect to the depth, one requires highly oscillating pre-activations which in turn can only be achieved with a delicate selection of parameters. In fact, it can be shown that through random initialization the expected number of activation regions per unit cube depends mainly on the number of neurons in the NN, rather than its depth~\cite{hanin2019deep}.}.

This approach was further developed in~\cite{raghu2017expressivepower} to a framework to study expressivity that to some extent allows to include the training phase.
One central object studied in~\cite{raghu2017expressivepower} are so-called \emph{trajectory lengths}. In this context, one analyzes how the length of a non-constant curve in the input space changes in expectation through the layers of a NN. The authors find an exponential dependence of the expected curve length on the depth. Let us motivate this in the special case of a ReLU NN with architecture $a=((N_0,n,\dots,n,N_{L}),\ReLU)$ and depth $L\in\N$. 

\begin{figure}[t]
    \centering
    \includegraphics[width=\textwidth]{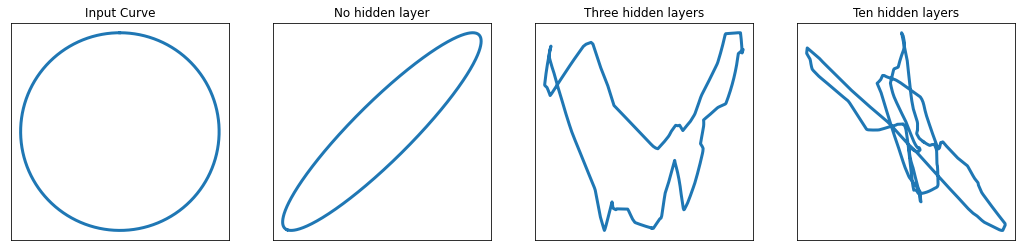}
    \caption{Shape of the trajectory $t \mapsto \Phi_{((2,n,\dots,n,2),\ReLU)}(\gamma(t),\theta)$ of the output of a randomly initialized network with $0, 3, 10$ hidden layers. The input curve $\gamma$ is the circle given in the leftmost image. The hidden layers have $n= 20$ neurons and the variance of the initialization is taken as $4/n$.}
    \label{fig:trajectoryShape}
\end{figure}

Given a non-constant continuous curve $\gamma\colon [0,1]\to \R^{N_0}$ in the input space, the length of the trajectory in the $\ell$-th layer of the NN $\Phi_{a}(\cdot,\theta)$ is then given by
\begin{equation*}
    \operatorname{Length}(\bar{\Phi}^{(\ell)}(\gamma(\cdot),\theta)), \quad \ell \in [L-1], 
\end{equation*}
where $\bar{\Phi}^{(\ell)}(\cdot,\theta)$ is the activation in the $\ell$-th layer, see~\eqref{eq:def_nn_classical}. Here the length of the curve is well-defined since $\bar{\Phi}^{(\ell)}(\cdot,\theta))$ is continuous and therefore $\bar{\Phi}^{(\ell)}(\gamma(\cdot),\theta)$ is continuous.
Now, let the parameters $\Theta_{1}$ of the NN $\Phi_{a}$ be initialized independently so that
the entries corresponding to the weight matrices and bias vectors follow a normal distribution with zero mean and variances $1/n$ and $1$, respectively. It is not hard to see, e.g., by Proposition~\ref{prop:interpolation}, that the probability that $\bar{\Phi}^{(\ell)}(\cdot,\Theta_{1})$ will map $\gamma$ to a non-constant curve is positive and hence, for fixed $\ell\in [L-1]$,
\begin{equation*}
    \E\big[\operatorname{Length}(\bar{\Phi}^{(\ell)}(\gamma(\cdot),\Theta_{1}))\big] = c > 0.
\end{equation*}
Let $\sigma \in(0,\infty)$ and consider a second initialization $\Theta_{\sigma}$, where we change the variances of the entries corresponding to the weight matrices and bias vectors to $\sigma^2/n$ and $\sigma^2$, respectively. Recall that the ReLU is positively homogeneous, i.e., we have that $\ReLU(\lambda x ) = \lambda \ReLU( x )$ for all $\lambda\in (0,\infty)$. Then it is clear that 
\begin{equation*}
\bar{\Phi}^{(\ell)}(\cdot,\Theta_{\sigma}) \sim \sigma^{\ell} \bar{\Phi}^{(\ell)}(\cdot,\Theta_{1}),
\end{equation*}
i.e., the activations corresponding to the two initialization strategies are identically distributed up to the factor $\sigma^{\ell}$. Therefore, we immediately conclude that 
\begin{equation*}
    \E\big[\operatorname{Length}(\bar{\Phi}^{(\ell)}(\gamma(\cdot),\Theta_{\sigma}))\big] = \sigma^{\ell} c.
\end{equation*}
This shows that the expected trajectory length depends exponentially on the depth of the NN, which is in line with the behavior of other notions of expressivity~\cite{poole2016exponential}.
In~\cite{raghu2017expressivepower} this result is also extended to the $\operatorname{tanh}$ activation function and the constant $c$ is more carefully resolved. Empirically one also finds that the shapes of the trajectories become more complex in addition to becoming longer on average, see Figure~\ref{fig:trajectoryShape}.

\section{Deep neural networks overcome the curse of dimensionality}\label{sec:CurseOfDimension}
\begin{wrapfigure}{r}{0.34\textwidth}
\includegraphics[width =\linewidth]{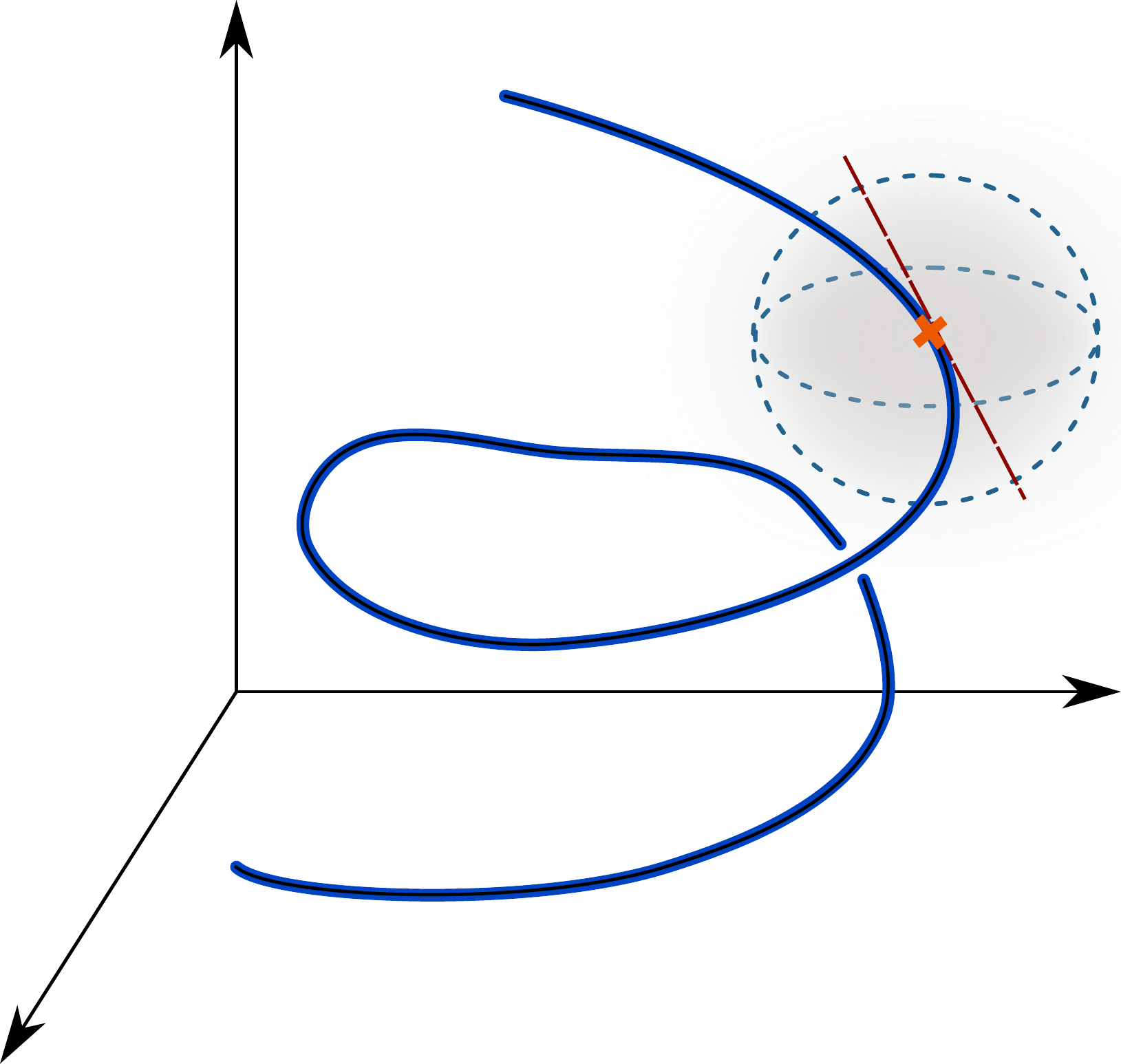}
\put(-38, 25){\color{blue} $\mathbf{\mathcal{M}}$}
\caption{Illustration of a one-dimensional manifold $\mathcal{M}$ embedded in $\R^3$.
For every point $x \in \mathcal{M}$ there exists a neighborhood in which the manifold can be linearly projected onto its tangent space at $x$ such that the corresponding inverse function is differentiable.}
\label{fig:manifold}
\vspace{-1em}
\end{wrapfigure}

In Subsection~\ref{subsec:NewTheories}, one of the main puzzles of deep learning that we identified was the surprising performance of deep architectures on problems where the input dimensions are very high.
This performance cannot be explained in the framework of classical approximation theory, since such results always suffer from the curse of dimensionality~\cite{bellman1952theory, devore1998nonlinear, novak2009approximation}.

In this section, we present three approaches that offer explanations of this phenomenon. As before, we had to omit certain ideas which have been very influential in the literature to keep the length of this section
under control. In particular, an important line of reasoning is that functions to be approximated often have compositional structures which NNs may approximate very well as reviewed in~\cite{poggio2017and}. Note that also a suitable feature descriptor, factoring out invariances, might lead to a significantly reduced effective dimension, see Subsection~\ref{subsec:wavelets}.

\subsection{Manifold assumption}
A first remedy to the high-dimensional curse of dimensionality is what we call the \emph{manifold assumption}. Here it is assumed that we are trying to approximate a function 
\begin{equation*}
    g \colon \R^d \supset \cX \to \R,
\end{equation*}
where $d$ is very large. However, we are not seeking to optimize with respect to the uniform norm or a regular $L^p$ space, but instead consider a measure $\mu$ which is supported on a $d'$-dimensional manifold $\mathcal{M} \subset \cX$. Then the error is measured in the $L^p(\mu)$-norm. Here we consider the case where $d'\ll d$. This setting is appropriate if the data $z=(x,y)$ of a prediction task is generated from a measure supported on $\mathcal{M} \times \R$.

This set-up or generalizations thereof have been fundamental in~\cite{chui2018deep, shaham2018provable, chen2019efficient, schmidt2019deep, cloninger2020relu, nakada2020adaptive}. Let us describe an exemplary approach, where we consider locally $C^k$-regular functions and NNs with ReLU activation functions below:

\begin{enumerate}
    \item \label{it:manifold_step1} \textit{Describe the regularity of $g$ on the manifold:} Naturally, we need to quantify the regularity of the function $g$ restricted to $\mathcal{M}$ in an adequate way. The typical approach would be to make a definition via local coordinate charts. If we assume that $\mathcal{M}$ is an embedded submanifold of $\cX$, then locally, i.e., in a neighborhood of a point $x \in \mathcal{M}$, the orthogonal projection of $\mathcal{M}$ onto the $d'$-dimensional tangent space $T_x\mathcal{M}$ is a diffeomorphism. The situation is depicted in Figure~\ref{fig:manifold}. Assuming $\mathcal{M}$ to be compact, we can choose a finite set of open balls $(U_i)_{i=1}^p$ that cover $\mathcal{M}$ and on which the local projections $\gamma_i$ onto the respective tangent spaces as described above exists and are diffeomorphisms. Now we can define the regularity of $g$ via classical regularity. In this example, we say that $g \in C^k(\mathcal{M})$ if $g \circ \gamma_i^{-1} \in C^{k}( \gamma_i(\mathcal{M} \cap U_i))$ for all $i \in [p]$.
    
    \item \textit{Construct localization and charts via neural networks:} According to the construction of local coordinate charts in Step~\ref{it:manifold_step1}, we can write $g$ as follows:
\begin{equation}
\label{eq:representationOfg} 
         g(x) = \sum_{i = 1}^p \phi_i(x) \left( g \circ \gamma_i^{-1}( \gamma_i(x) )\right)
         \eqqcolon \sum_{i = 1}^p  \tilde{g}_i(\gamma_i(x), \phi_i(x)), \quad x \in \mathcal{M},
\end{equation}
where $\phi_i$ is a partition of unity such that $\supp(\phi_i) \subset U_i$. Note that $\gamma_i$ is a linear map, hence representable by a one-layer NN. Since multiplication is a smooth operation, we have that if $g \in C^k(\cM)$ then $\tilde{g}_i \in C^k(  \gamma_i(\mathcal{M} \cap U_i) \times [0,1])$.
    
The partition of unity $\phi_i$ needs to be emulated by NNs. For example, if the activation function is the ReLU, then such a partition can be efficiently constructed. Indeed, in~\cite{he2020relu} it was shown that such NNs can represent linear finite elements exactly with fixed-size NNs and hence a partition of unity subordinate to any given covering of $\mathcal{M}$ can be constructed.
    
    \item \textit{Use a classical approximation result on the localized functions:} By some form of Whitney's extension theorem~\cite{whitney1934analytic}, we can extend each $\tilde{g}_i$ to a function $\bar{g}_i \in C^{k}(\cX \times[0,1])$ which by classical results can be approximated up to an error of $\eps > 0$ by NNs of size $\cO(\eps^{-(d'+1)/k})$ for $\eps \to 0$, see~\cite{mhaskar1996neural, yarotsky2017error, shaham2018provable}.    
    
    \item \textit{Use the compositionality of neural networks to build the final network:} We have seen that every component in the representation~\eqref{eq:representationOfg}, i.e., $\tilde{g}_i$, $\gamma_i$, and $\phi_i$ can be efficiently represented by NNs. In addition, composition and summation are operations which can directly be implemented by NNs through increasing their depth and widening their layers. Hence~\eqref{eq:representationOfg} is efficiently---i.e., with a rate depending only on $d'$ instead of the potentially much larger $d$---approximated by a NN.
\end{enumerate}

Overall, we see that NNs are capable of learning local coordinate transformations and therefore reduce the complexity of a high-dimensional problem to the underlying low-dimensional problem given by the data distribution.

\subsection{Random sampling}\label{subsec:RandSamp}

Already in 1992, Andrew Barron showed that under certain seemingly very natural assumptions on the function to approximate, a dimension-independent approximation rate by NNs can be achieved~\cite{barron1992neural, barron1993universal}. Specifically, the assumption is formulated as a condition on the Fourier transform of a function and the result is as follows. 
\begin{theorem}[Approximation of Barron-regular functions]
\label{thm:barron}
Let $\varrho\colon\R\to\R$ be the ReLU or a sigmoidal function. Then there exists a constant $c\in(0,\infty)$ with the following property: For every $d,n\in\N$, every probability measure $\mu$ supported on $B_1(0)\subset \R^d$, and every $g\in L^1(\R^d)$ with $C_g\coloneqq\int_{\R^d} \, \|\xi\|_2|\hat{g}(\xi)| \, \mathrm{d}\xi < \infty $ it holds that
\begin{equation*}
    \inf_{\theta \in \R^{P((d,n,1))}}\| \Phi_{((d,n,1),\varrho)}(\cdot,\theta) -g \|_{L^2(\mu)} \le  \frac{c}{\sqrt{n}}C_g,
\end{equation*}
\end{theorem}

Note that the $L^2$-approximation error can be replaced by an $L^\infty$-estimate over the unit ball at the expense of a factor of the order of $\sqrt{d}$ on the right-hand side.

The key idea behind Theorem~\ref{thm:barron} is the following application of the law of large numbers: First, we observe that, per assumption, $g$ can be represented via the inverse Fourier transform, as 
\begin{equation}
\begin{split}
    g -g(0)&= \int_{\R^d} \hat{g}(\xi) (e^{2\pi i \langle \cdot, \xi \rangle} - 1)\, \mathrm{d}\xi \\
    &= C_g \int_{\R^d} \frac{1}{\|\xi\|_2} (e^{2\pi i \langle \cdot, \xi \rangle}-1) \frac{1}{C_g}\|\xi\|_2\hat{g}(\xi)  \, \mathrm{d}\xi\\
    & = C_g \int_{\R^d} \frac{1}{\|\xi\|_2} (e^{2\pi i \langle \cdot, \xi \rangle}-1) \, \mathrm{d}\mu_g(\xi),
    \label{eq:thisShowsThatComplExponentialsAreAsGoodAsNNs}
\end{split}
\end{equation}
where $\mu_g$ is a probability measure. 
Then it is further shown in~\cite{barron1992neural} that there exist $(\R^d \times \R)$-valued random variables $(\Xi, \widetilde{\Xi})$ such that~\eqref{eq:thisShowsThatComplExponentialsAreAsGoodAsNNs} can be written as 
\begin{equation}
\label{eq:ThisIsTheStartingPointForWeinan}
g(x) -g(0) = C_g\int_{\R^d} \frac{1}{\|\xi\|_2} (e^{2\pi i \langle x, \xi \rangle}-1) \, \mathrm{d}\mu_g(\xi) =C_g \E\big[\Gamma(\Xi,\widetilde{\Xi})(x)\big],  \quad x\in\R^d,
\end{equation}
where for every $\xi\in \R^d$, $\tilde{\xi}\in\R$ the function $\Gamma(\xi, \tilde{\xi})\colon \R^d\to\R$ is given by
\begin{equation*}
    \Gamma(\xi, \tilde{\xi}) \coloneqq s(\xi, \tilde{\xi})(\mathds{1}_{(0,\infty)}( -\langle \xi/\|\xi\|_2, \cdot \rangle - \tilde{\xi}) - \mathds{1}_{(0,\infty)}(\langle \xi/\|\xi\|_2, \cdot \rangle - \tilde{\xi})) \quad \text{with} \quad s(\xi, \tilde{\xi})\in\{-1,1\}.
\end{equation*}
Now, let $((\Xi^{(i)}, \widetilde{\Xi}^{(i)}))_{i\in \N}$ be i.i.d.\@ random variables with $(\Xi^{(1)}, \widetilde{\Xi}^{(1)})\sim(\Xi, \widetilde{\Xi}) $. Then, Bienaymé's identity and Fubini's theorem establish that
\begin{equation}
\label{eq:l2_bienayme}
\begin{split}
        \E\left[\Big\|g - g(0) - \frac{C_g}{n} \sum_{i=1}^n \Gamma(\Xi^{(i)}, \widetilde{\Xi}^{(i)})\Big\|_{L^2(\mu)}^2 \right] 
        &=\int_{B_1(0)} \V\left[\frac{C_g}{n} \sum_{i=1}^n \Gamma(\Xi^{(i)}, \widetilde{\Xi}^{(i)})(x) \right] \, \mathrm{d}\mu(x) 
        \\
        &=  \frac{ C_g^2 \int_{B_1(0)}\V\big[\Gamma(\Xi, \widetilde{\Xi})(x)\big] \, \mathrm{d}\mu(x)}{n}
         \le \frac{ (2\pi C_g)^2 }{n},
\end{split}
\end{equation}
where the last inequality follows from combining~\eqref{eq:ThisIsTheStartingPointForWeinan} with the fact that $| e^{2\pi i \langle x, \xi \rangle}-1|/\|\xi\|_2 \leq 2\pi$, $x\in B_1(0)$.
 
This implies that there exists a realization $((\xi^{(i)}, \tilde{\xi}^{(i)}))_{i\in \N}$ of the random variables $((\Xi^{(i)}, \widetilde{\Xi}^{(i)}))_{i\in \N}$ that achieves $L^2$-approximation error of $n^{-1/2}$. Therefore, it remains to show that NNs can well approximate the functions $((\Gamma(\xi^{(i)}, \tilde{\xi}^{(i)}))_{i\in\N}$. Now it is not hard to see that the function $\mathds{1}_{(0,\infty)}$ and hence functions of the form $\Gamma(\xi, \tilde{\xi})$, $\xi\in \R^d$, $\tilde{\xi}\in\R$, can be arbitrarily well approximated with a fixed-size, two-layer NN with a sigmoidal or ReLU activation function.
Thus, we obtain an approximation rate of $n^{-1/2}$ when approximating functions with one finite Fourier moment by two-layer NNs with $n$ hidden neurons.

It was pointed out already in the dissertation of Emmanuel Cand\`es~\cite{candes1998ridgelets} that the approximation rate of NNs for Barron-regular functions is also achievable by $n$-term approximation with complex exponentials, as is apparent by considering~\eqref{eq:thisShowsThatComplExponentialsAreAsGoodAsNNs}. However, for deeper NNs, the results also extend to high-dimensional non-smooth functions, where Fourier-based methods are certain to suffer from the curse of dimensionality~\cite{caragea2020neural}. 

In addition, the random sampling idea above was extended in~\cite{ma2019priori, ma2020towards, wojtowytsch2020priori, wojtowytsch2020representation} to facilitate dimension-independent approximation of vastly more general function spaces. Basically, the idea is to use~\eqref{eq:ThisIsTheStartingPointForWeinan} as an inspiration and define the \emph{generalized Barron space} as all functions that may be represented as 
\begin{equation*}
    \E\big[\mathds{1}_{(0,\infty)}(\langle \Xi, \cdot \rangle - \widetilde{\Xi}) \big] 
\end{equation*}
for any random variable $(\Xi, \widetilde{\Xi})$. In this context, deep and compositional versions of Barron spaces were introduced and studied in~\cite{barron2018approximation, ma2019barron, wojtowytsch2020banach}, which considerably extend the original theory.

\subsection{PDE assumption}\label{subsec:PDEapprox}
Another structural assumption that leads to the absence of the curse of dimensionality in some cases is that the function we are trying to approximate is given as the solution to a partial differential equation. It is by no means clear that this assumption leads to approximation without the curse of dimensionality, since most standard methods, such as finite elements, sparse grids, or spectral methods typically suffer from the curse of dimensionality. 

This is not merely an abstract theoretical problem: Very recently, in~\cite{al2020interactions} it was shown that two different gold standard methods for solving the multi-electron
Schrödinger equation produce completely different interaction energy predictions when applied to large delocalized molecules. Classical numerical representations are simply not expressive enough to accurately represent complicated high-dimensional structures such as wave functions with long-range interactions. 

Interestingly, there exists an emerging body of work that shows that NNs do not suffer from these shortcomings and enjoy superior expressivity properties as compared to standard numerical representations. Such results include, for example,~\cite{grohs2018proof, gonon2020deep, hutzenthaler2020proof} for (linear and semilinear) parabolic evolution equations,~\cite{grohs2022deep} for stationary elliptic PDEs,~\cite{grohs2021deep} for nonlinear Hamilton--Jacobi--Bellman equations, or~\cite{kutyniok2019theoretical} for parametric PDEs. In all these cases, the absence of the curse of dimensionality in terms of the theoretical approximation power of NNs could be rigorously established.

One way to prove such results is via stochastic representations of the PDE solutions, as well as associated sampling methods. We illustrate the idea for the simple case of linear Kolmogorov PDEs, that is
the problem of representing the function $g\colon\mathbb{R}^d\times [0,\infty) \to \mathbb{R}$ satisfying\footnote{The natural solution concept to this type of PDEs is the viscosity solution concept, a thorough study of which can be found in~\cite{hairer2015loss}.}
\begin{equation} \label{eq:kol_pde}
\frac{\partial g }{\partial t} (x,t) = \frac{1}{2} \trace \big(\sigma(x,t)  [\sigma(x,t)  ]^{*}\nabla_x^2 g(x,t) \big) + \langle \mu(x,t) , \nabla_x g(x,t) \rangle, \quad
g (x,0) = \varphi(x),
\end{equation}
where the functions
\begin{equation*}
\varphi  \colon \mathbb{R}^d \rightarrow \mathbb{R} \quad \text{(initial condition)} \quad \text{and} \quad
\sigma \colon \mathbb{R}^d \rightarrow \mathbb{R}^{d \times d}, \quad 
\mu \colon \mathbb{R}^d \rightarrow \mathbb{R}^{d} \quad \text{(coefficient functions)}
\end{equation*}
are continuous and satisfy suitable growth conditions. 
A stochastic representation of $g$ is given via the Ito processes $(\mathcal{S}_{x,t})_{t\ge 0}$ satisfying
\begin{equation}\label{eq:sode} 
d\mathcal{S}_{x,t} = \mu   (\mathcal{S}_{x,t}) dt + \sigma  (\mathcal{S}_{x,t}) dB_t, \quad \mathcal{S}_{x,0} = x,
\end{equation}
 where $(B_t)_{t \ge 0}$ is a $d$-dimensional Brownian motion. Then $g$ is described via the Feynman--Kac formula which states that 
 \begin{equation}\label{eq:feynmankac}
     g(x,t) = \E[\varphi(\mathcal{S}_{x,t})], \quad x\in\R^d, \ t \in [0,\infty).
 \end{equation}
 Roughly speaking, a NN approximation result can be proven by first approximating, via the law of large numbers, 
 \begin{equation}\label{eq:feynmankacMC}
    g(x,t) = \E[\varphi(\mathcal{S}_{x,t})] \approx \frac{1}{n}\sum_{i=1}^n\varphi (\mathcal{S}_{x,t}^{(i)}),     
 \end{equation}
 where $(\mathcal{S}_{x,t}^{(i)})_{i=1}^n$ are i.i.d.\@ random variables with $\mathcal{S}_{x,t}^{(1)} \sim \mathcal{S}_{x,t}$. Care has to be taken to establish such an approximation \emph{uniformly in the computational domain}, for example, for every $(x,t)$ in the unit cube $[0,1]^{d}\times [0,1]$, see~\eqref{eq:l2_bienayme} for a similar estimate and~\cite{grohs2018proof, gonon2020deep} for two general approaches to ensure this property.
Aside from this issue,~\eqref{eq:feynmankacMC} represents a standard Monte Carlo estimator which can be shown to be free of the curse of dimensionality. 

As a next step, one needs to establish that realizations of the processes $(x,t)\mapsto \mathcal{S}_{x,t}$ can be efficiently approximated by NNs. This can be achieved by emulating a suitable time-stepping scheme for the SDE~\eqref{eq:sode} by NNs which, roughly speaking, can be done without incurring the curse of dimensionality whenever the coefficient functions $\mu, \sigma$ can be approximated by NNs without incurring the curse of dimensionality and some growth conditions hold true. In a last step one assumes that the initial condition $\varphi$ can be approximated by NNs without incurring the curse of dimensionality which, by the compositionality of NNs and the previous step, directly implies that realizations of the processes $(x,t)\mapsto \varphi(\mathcal{S}_{x,t})$ can be approximated by NNs without incurring the curse of dimensionality. By~\eqref{eq:feynmankacMC} this implies a corresponding approximation result for the solution of the Kolmogorov PDE $g$ in~\eqref{eq:kol_pde}. 

Informally, we have discovered a regularity result for linear Kolmogorov equations, namely that (modulo some technical conditions on $\mu, \sigma$),
 \emph{the solution $g$ of~\eqref{eq:kol_pde} can be approximated by NNs without incurring the curse of dimensionality whenever the same holds true for the initial condition $\varphi$, as well as the coefficient functions $\mu, \sigma$}.
 In other words, \emph{the property of being approximable by NNs without curse of dimensionality is preserved under the flow induced by the PDE~\eqref{eq:kol_pde}}.
 Some comments are in order:
 \paragraph{Assumption on the initial condition:}
 One may wonder if the assumption that the initial condition $\varphi$ can be approximated by NNs without incurring the curse of dimensionality is justified. This is at least the case in many applications in computational finance where the function $\varphi$ typically represents an option pricing formula and~\eqref{eq:kol_pde} represents the famous Black--Scholes model.
 It turns out that nearly all common option pricing formulas are constructed from iterative applications of linear maps and maximum/minimum functions---in other words, in many applications in computational finance, the initial condition $\varphi$ can be \emph{exactly} represented by a small ReLU NN.
\paragraph{Generalization and optimization error:}
\begin{wrapfigure}{r}{0.5\textwidth}
\vspace{-0.5em}
\centering
\includegraphics[width = \linewidth]{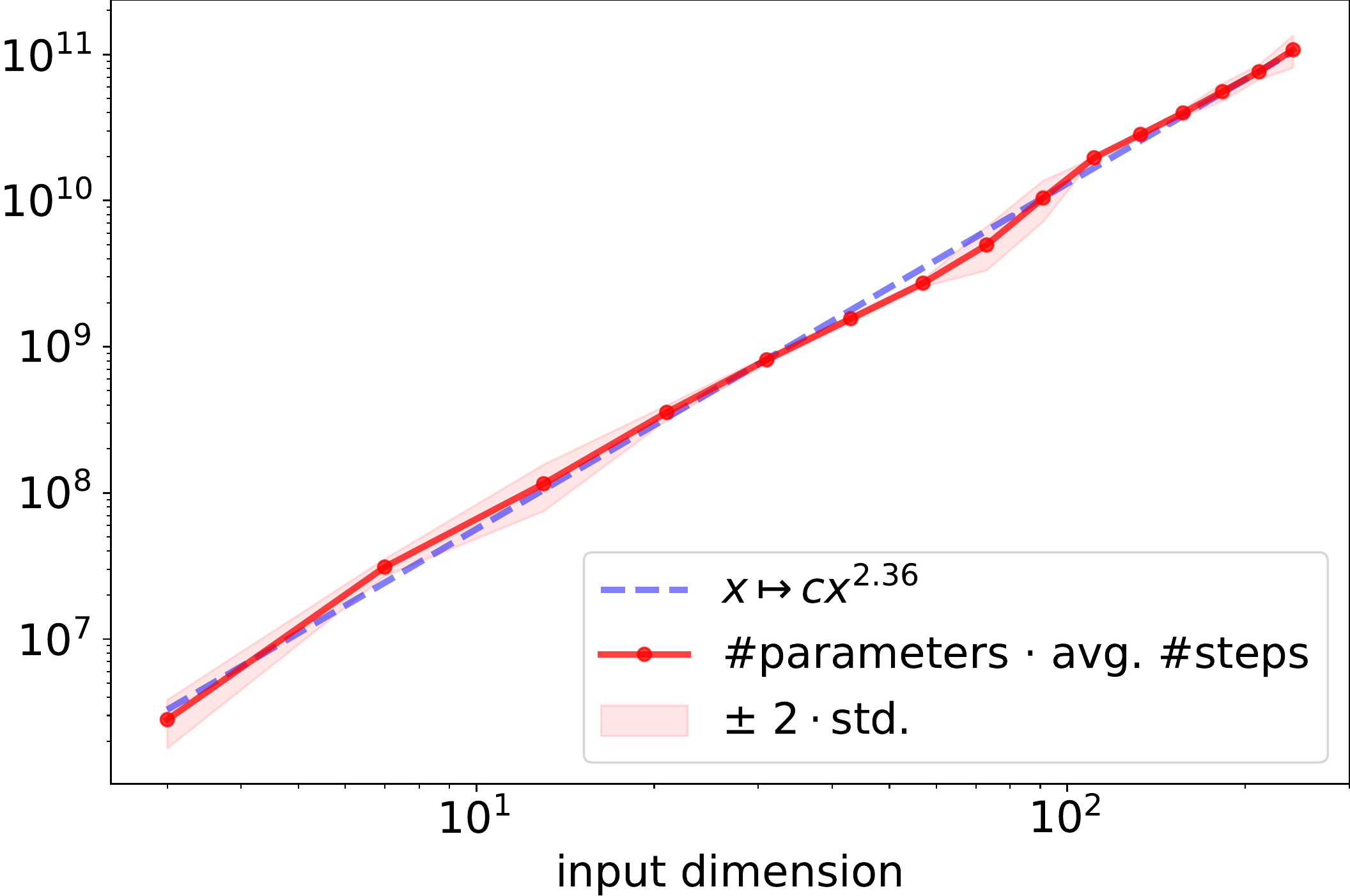}
\caption{Computational complexity as number of neural network parameters times number of SGD steps to solve heat equations of varying dimensions up to a specified precision. According to the fit above, the scaling is polynomial in the dimension~\cite{berner2020numerically}.}
\label{fig:curse}
\vspace{-1em}
\end{wrapfigure}
The Feynman--Kac representation~\eqref{eq:feynmankac} directly implies that $g(\cdot,t)$ can be computed as the Bayes optimal function of a regression task with input features $X\sim \mathcal{U}([0,1]^d)$ and labels $Y=\varphi(\mathcal{S}_{X,t})$, which allows for an analysis of the generalization error as well as implementations based on ERM algorithms~\cite{berner2020analysis, beck2021solving}. 

While it is in principle possible to analyze the approximation and generalization error, the analysis of the computational cost and/or convergence of corresponding SGD algorithms is completely open. Some promising numerical results exist, see, for instance, Figure~\ref{fig:curse}, but the stable training of NNs approximating PDEs to very high accuracy (that is needed in several applications such as quantum chemistry) remains very challenging. The recent work~\cite{grohs2021proof} has even proven several impossibility results in that direction.

\paragraph{Extensions and abstract idea:} Similar techniques may be used to prove expressivity results for nonlinear PDEs, for example, using nonlinear Feynman--Kac-type representations of~\cite{pardoux1992backward} in place of~\eqref{eq:feynmankac} and multilevel Picard sampling algorithms of~\cite{weinan2019multilevel} in place of~\eqref{eq:feynmankacMC}. 

We can also formulate the underlying idea in an abstract setting (a version of which has also been used in Subsection~\ref{subsec:RandSamp}). Assume that a high-dimensional function $g\colon \R^d \to\R$ admits a probabilistic representation of the form
\begin{equation}
\label{eq:stoch_representation}
g(x)=\E[Y_x], \quad x\in \R^d,
\end{equation}
for some random variable $Y_x$ which can be approximated 
by an iterative scheme 
\begin{equation*}
 \mathcal{Y}_x^{(L)} \approx Y_x \quad \text{and} \quad \mathcal{Y}_x^{(\ell)}=T_\ell(\mathcal{Y}_x^{(\ell-1)}), \quad \ell=1,\dots,L,
\end{equation*}
with dimension-independent convergence rate. If we can approximate realizations of the initial mapping $x\mapsto \mathcal{Y}^{0}_x$ and the maps $T_\ell$, $\ell\in[L]$, by NNs and the numerical scheme is stable enough, then we can also approximate $\mathcal{Y}_x^{(L)}$ using compositionality. Emulating a uniform Monte-Carlo approximator of~\eqref{eq:stoch_representation} then leads to approximation results for $g$ without curse of dimensionality. In addition, one can choose a $\R^d$-valued random variable $X$ as input features and define the corresponding labels by $Y_X$ to obtain a prediction task, which can be solved by means of ERM.
\paragraph{Other methods:} There exist a number of additional works related to the approximation capacities of NNs for high-dimensional PDEs, for example,~\cite{elbrachter2018dnn, li2019better, schwab2019deep}. In most of these works, the proof technique consists of emulating an existing method that does not suffer from the curse of dimensionality. For instance, in the case of first-order transport equations, one can show in some cases that NNs are capable of emulating the method of characteristics, which then also yields approximation results that are free of the curse of dimensionality~\cite{laakmann2021efficient}.

\section{Optimization of deep neural networks} \label{sec:optimizationNew}
We recall from Subsections~\ref{subsec:NewTheories} and~\ref{subsubsec:optimization} that the standard algorithm to solve the empirical risk minimization problem over the hypothesis set of NNs is stochastic gradient descent. This method would be guaranteed to converge to a global minimum of the objective if the empirical risk were convex, viewed as a function of the NN parameters. However, this function is severely nonconvex, may exhibit (higher-order) saddle points, seriously suboptimal local minima, and wide flat areas where the gradient is very small.

On the other hand, in applications, excellent performance of SGD is observed. This indicates that the trajectory of the optimization routine somehow misses suboptimal critical points and other areas that may lead to slow convergence. 
Clearly, the classical theory does not explain this performance. Below we describe some exemplary novel approaches that give partial explanations of this success.

In the flavor of this
\ifbook
\else%
article,
\fi%
the aim of this section is to present some selected ideas rather than giving an overview of the literature. To give at least some detail about the underlying ideas and to keep the length of this section reasonable, a selection of results had to be made and some ground-breaking results had to be omitted.

\subsection{Loss landscape analysis}

Given a NN $\Phi(\cdot, \theta)$ and training data $\cs\in \cZ^m$ the function $\theta \mapsto \prisk(\theta) \coloneqq \erisk{\cs}(\Phi(\cdot, \theta))$ describes, in a natural way, through its graph, a high-dimensional surface. This surface may have regions associated with lower values of $\erisk{\cs}$ which resemble valleys of a landscape if they are surrounded by regions of higher values. The analysis of the topography of this surface is called \emph{loss landscape analysis}. Below we shall discuss a couple of approaches that yield deep insights into the shape of this landscape.

\paragraph{Spin glass interpretation:}
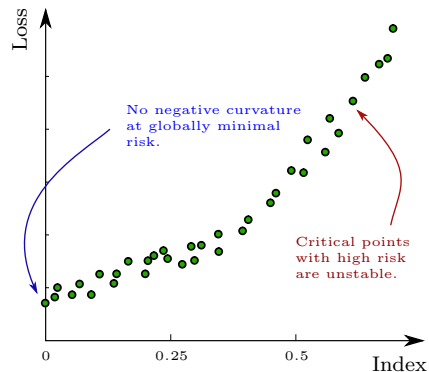
\begin{wrapfigure}{r}{0.34\textwidth}
\vspace{-0.5em}
\includegraphics[width = 0.95\linewidth, right]{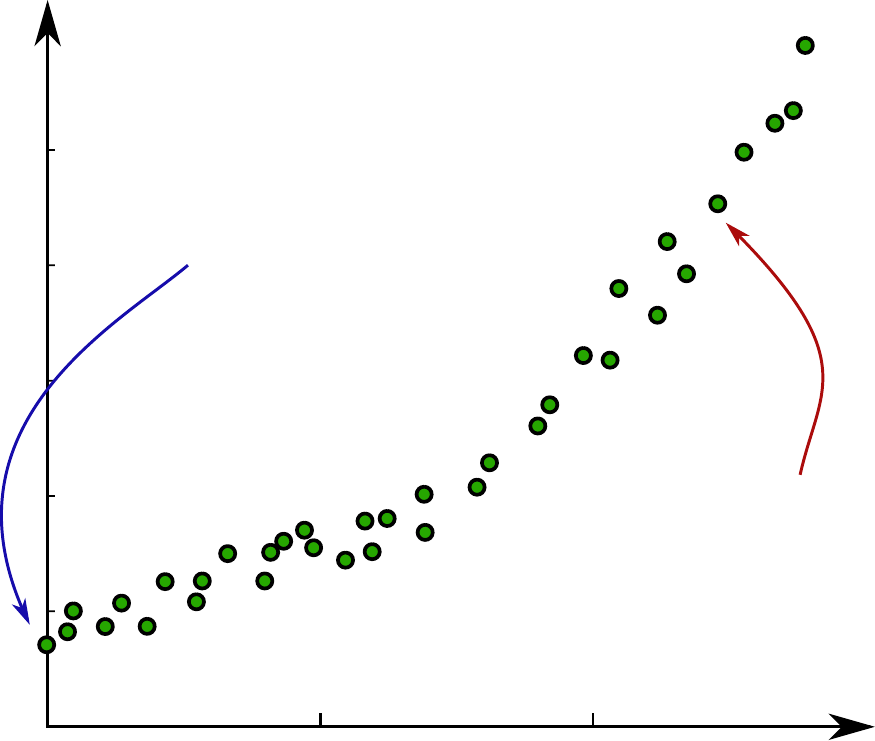}
\put(-20, -9){\footnotesize Index}
\put(-156, 112){\footnotesize \rotatebox{90}{Loss}}
\put(-113, 88){\tiny \color{blue} No negative curvature}
\put(-113, 82){\tiny \color{blue} at globally minimal}
\put(-113, 76){\tiny \color{blue} risk.}
\put(-49, 38){\tiny \color{darkred} Critical points}
\put(-49, 32){\tiny \color{darkred} with high risk}
\put(-49, 26){\tiny \color{darkred} are unstable.}
\put(-145, -5){\tiny $0$}
\put(-102, -5){\tiny $0.25$}
\put(-53, -5){\tiny $0.5$}
\caption{Sketch of the distribution of critical points of the Hamiltonian of a spin glass model.}
\label{fig:spinGlassCurve}
\end{wrapfigure}
One of the first discoveries about the shape of the loss landscape comes from deep results in statistical physics. The Hamiltonian of the \emph{spin glass model} is a random function on the $(n-1)$-dimensional sphere of radius $\sqrt{n}$. Making certain simplifying assumptions, it was shown in~\cite{choromanska2015loss} that the loss of a NN with random inputs can be considered as the Hamiltonian of a spin glass model, where the inputs of the model are the parameters of the NN.

This connection has far-reaching implications for the loss landscape of NNs because of the following surprising property of the Hamiltonian of spin glass models: Consider the set of critical points of this set, and associate to each point an \emph{index} that denotes the percentage of the eigenvalues of the Hessian at that point which are negative. This index corresponds to the relative number of directions in which the loss landscape has negative curvature. Then with high probability, a picture like we see in Figure~\ref{fig:spinGlassCurve} emerges~\cite{auffinger2013random}. More precisely, the further away from the optimal loss we are, the more unstable the critical points become. Conversely, if one finds oneself in a local minimum, it is reasonable to assume that the loss is close to the global minimum.

While some of the assumptions establishing the connection between the spin glass model and NNs are unrealistic in practice~\cite{choromanska2015open}, the theoretical distribution of critical points as in Figure~\ref{fig:spinGlassCurve} is visible in many practical applications~\cite{dauphin2014identifying}.

\paragraph{Paths and level sets:}

Another line of research is to understand the loss landscape by analyzing paths through the parameter space. In particular, the existence of paths in parameter space, such that the associated empirical risks are monotone along the path. Surely, should there exist a path of nonincreasing empirical risk from every point to the global minimum, then we can be certain that no non-global minima exist, since no such path can escape a minimum. An even stronger result holds. In fact, the existence of such paths shows that the loss landscape has connected level sets~\cite{freeman2017topology, venturi2019spurious}.

A crucial ingredient of the analysis of such paths are \emph{linear substructures}. Consider a biasless two-layer NN $\Phi$ of the form
\begin{equation}\label{eq:simpleNN}
\R^d \ni x \mapsto \Phi(x, \theta) \coloneqq \sum_{j = 1}^n \theta^{(2)}_j \varrho\big( \langle \theta^{(1)}_j , \begin{bmatrix}x \\  1 \end{bmatrix} \rangle\big),
\end{equation}
where $\theta^{(1)}_j \in \R^{d+1}$ for $j \in [n]$, $\theta^{(2)} \in \R^n$, $\varrho$ is a Lipschitz continuous activation function, and 
we augment the vector $x$ by a constant $1$ in the last coordinate as outlined in Remark~\ref{rem:nn_notation}. 
If we consider $\theta^{(1)}$ to be fixed, then it is clear that the space
\begin{equation}
    \widetilde{\cF}_{\theta^{(1)}} \coloneqq \{\Phi(\cdot, \theta) \colon \theta = (\theta^{(1)}, \theta^{(2)}), \  \theta^{(2)} \in \R^n\}
\end{equation}
is a linear space. If the risk\footnote{As most statements in this subsection are valid for the empirical risk $\prisk(\theta) = \erisk{\cs}(\Phi(\cdot, \theta))$ as well as the risk $\prisk(\theta) = \risk(\Phi(\cdot, \theta))$, given a suitable distribution of $Z$, we will just call $r$ the risk.} is convex, as is the case for the widely used quadratic or logistic loss, then this implies that 
$\theta^{(2)} \mapsto \prisk\big( (\theta^{(1)}, \theta^{(2)})\big)$ is a convex map and hence, for every parameter set $\cP \subset \R^n$ this map assumes its maximum on $\partial \cP$. Therefore, within the vast parameter space, there are many paths traveling along which does not increase the risk above the risk of the start and end points. 

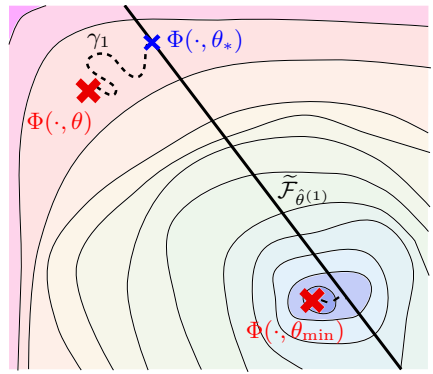
\begin{wrapfigure}{r}{0.34\textwidth}
\centering
\includegraphics[width =\linewidth]{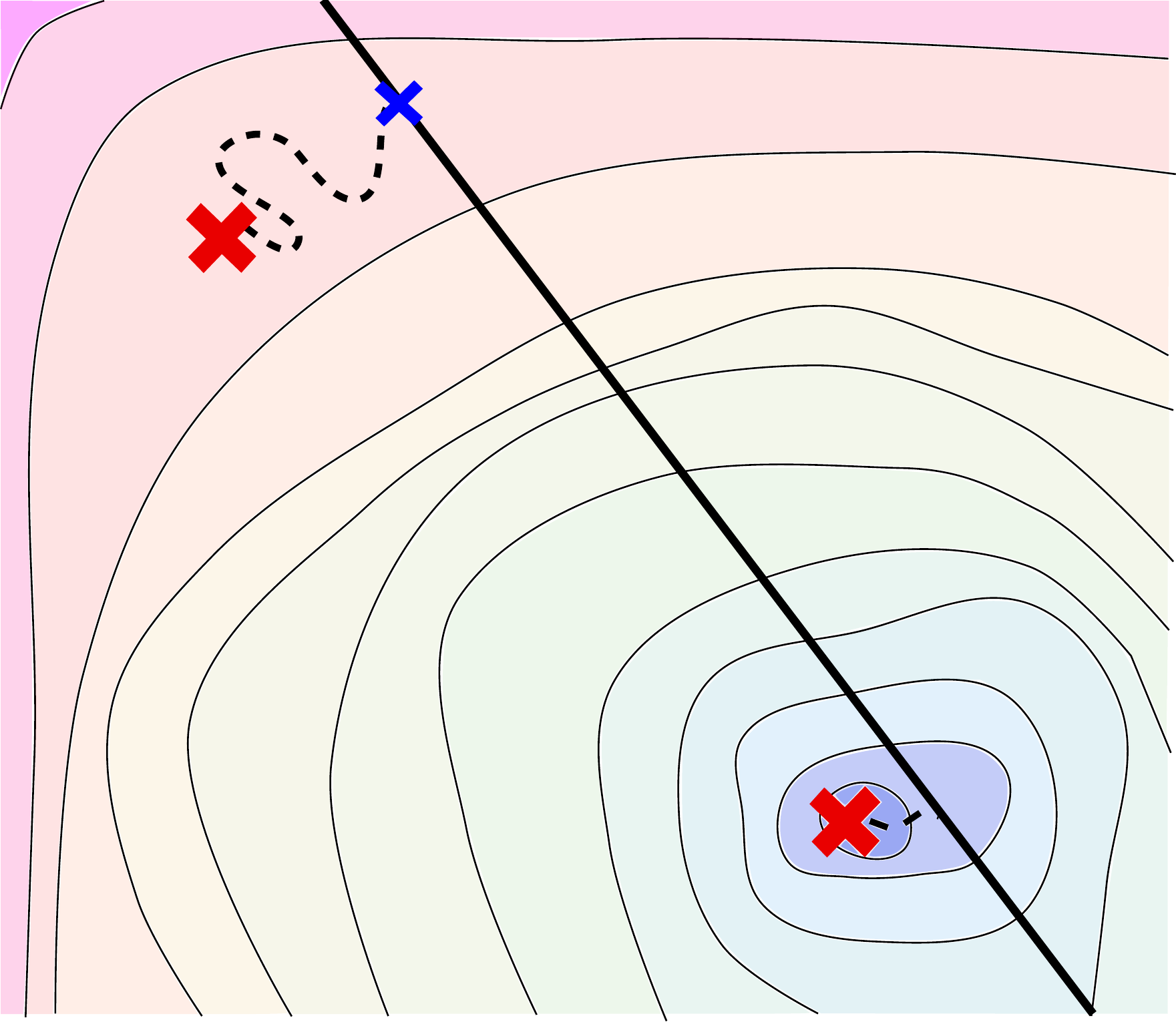}
\put(-70, 13){\small \color{red} $\Phi(\cdot, \theta_{\min })$}
\put(-153, 92){\small \color{red}$\Phi(\cdot, \theta)$}
\put(-58, 66){\small $\widetilde{\cF}_{\hat{\theta}^{(1)}}$}
\put(-100, 123){\small \color{blue} $\Phi(\cdot, \theta_{*})$}
\put(-130, 124){\small $\gamma_1$}
\caption{Construction of a path from an initial point $\theta$ to the global minimum $\theta_{\min}$ that does not have significantly higher risk than the initial point along the way. We depict here the landscape as a function of the neural network realizations instead of their parametrizations so that this landscape is convex.}
\label{fig:linpath}
\end{wrapfigure}

This idea was, for example, used in~\cite{freeman2017topology} in a way similar to the following simple sketch: Assume that, for two parameters $\theta$ and $\theta_{\min}$ there
exists a linear subspace of NNs $\widetilde{\cF}_{\hat{\theta}^{(1)}}$ such that there 
are paths $\gamma_1$ and $\gamma_2$ connecting $\Phi(\cdot, \theta)$ and $\Phi(\cdot, \theta_{\min})$ to $\widetilde{\cF}_{\hat{\theta}^{(1)}}$ respectively. Further assume that the paths are such that along $\gamma_1$ and $\gamma_2$ the risk does not significantly exceed $\max\{\prisk(\theta), \prisk( \theta_{\min})\}$. Figure~\ref{fig:linpath} shows a visualization of these paths. In this case, a path from $\theta$ to $\theta_{\min}$ not significantly exceeding $\prisk(\theta)$ along the way is found by concatenating the paths $\gamma_1$, a path along $\widetilde{\cF}_{\hat{\theta}^{(1)}}$, and $\gamma_2$. By the previous discussion, we know that only $\gamma_1$ and $\gamma_2$ determine the extent to which the combined path exceeds $\prisk(\theta)$ along its way. Hence, we need to ask about the existence of $\widetilde{\cF}_{\hat{\theta}^{(1)}}$ that facilitates the construction of appropriate $\gamma_1$ and $\gamma_2$. 

To understand why a good choice of $\widetilde{\cF}_{\hat{\theta}^{(1)}}$, so that the risk along $\gamma_1$ and $\gamma_2$ will not rise much higher than $\prisk(\theta)$, is likely possible, we set\footnote{We assume w.l.o.g.\@ that $n$ is a multiple of $2$.} 
\begin{equation}
\hat{\theta}^{(1)}_j \coloneqq \begin{cases}
     \theta^{(1)}_j & \text{for } j \in [n/2],
     \\
     (\theta_{\min }^{(1)})_j &\text{for } j\in [n]\setminus [n/2].
\end{cases}
\end{equation}

In other words, the first half of $\hat{\theta}^{(1)}$ is made from $\theta^{(1)}$ and the second from $\theta_{\min }^{(1)}$. If $\theta_j^{(1)}$, $j\in [N]$, are realizations of random variables distributed uniformly on the $d$-dimensional unit sphere, then by invoking standard covering bounds of 
spheres (e.g.,~\cite[Corollary 4.2.13]{vershynin2018high}), we expect that, for $\eps > 0$ 
and a sufficiently large number of neurons $n$, the vectors $(\theta^{(1)}_j)_{j = 1}^{n/2}$ already $\eps$-approximate all vectors $(\theta^{(1)}_j)_{j = 1}^n$. Replacing all vectors $(\theta^{(1)}_j)_{j = 1}^{n}$ by their nearest neighbor in $(\theta^{(1)}_j)_{j = 1}^{n/2}$ can be done with a linear path in the parameter space, and, given that $\prisk$ is locally Lipschitz continuous and $\|\theta^{(2)}\|_1$ is bounded, this operation will not increase the risk by more than $\cO(\eps)$. We denote the vector resulting from this replacement procedure by $\theta_{*}^{(1)}$. Since for all $j\in[n]\setminus [n/2]$ we now have that 
\begin{equation*}
    \varrho\big(\langle (\theta_{*}^{(1)})_j, \begin{bmatrix}\cdot \\  1 \end{bmatrix}\rangle\big) \in
    \left\{\varrho\big(\langle (\theta_*^{(1)})_k, \begin{bmatrix} \cdot \\  1 \end{bmatrix}\rangle\big)\colon k \in [n/2]\right\},
\end{equation*}
there exists a vector $\theta_{*}^{(2)}$ with $(\theta_{*}^{(2)})_j = 0$, $j\in[n]\setminus [n/2]$, so that 
\begin{equation*}
    \Phi(\cdot, (\theta_{*}^{(1)},\theta^{(2)}))= \Phi(\cdot, (\theta_{*}^{(1)}, \lambda \theta_{*}^{(2)}+ (1-\lambda)\theta^{(2)})), \quad \lambda\in[0,1].
\end{equation*}
In particular, this path does not change the risk between $(\theta_{*}^{(1)}, \theta^{(2)})$ and $(\theta_{*}^{(1)}, \theta_{*}^{(2)})$. Now, since $(\theta_{*}^{(2)})_j = 0$ for $j \in [n]\setminus [n/2]$, the realization $\Phi(\cdot, (\theta_{*}^{(1)},\theta_{*}^{(2)}))$ is computed by a sub-network consisting of the first $n/2$ hidden neurons and we can replace the parameters corresponding to the other neurons without any effect on the realization function. Specifically, it holds that 
\begin{equation*}
    \Phi(\cdot, (\theta_{*}^{(1)}, \theta_{*}^{(2)})) = \Phi(\cdot, ( \lambda\hat{\theta}^{(1)}+(1-\lambda)\theta_{*}^{(1)},\theta_{*}^{(2)})), \quad \lambda\in[0,1],
\end{equation*}
yielding a path of constant risk between $(\theta_{*}^{(1)}, \theta_{*}^{(2)})$ and $(\hat{\theta}^{(1)},\theta_{*}^{(2)})$. Connecting these paths completes the construction of $\gamma_1$ and shows that the risk along $\gamma_1$ does not exceed that at $\theta$ by more than $\cO(\eps)$. Of course, $\gamma_2$ can be constructed in the same way. The entire construction is depicted in Figure~\ref{fig:linpath}.

Overall, this derivation shows that for sufficiently wide NNs (appropriately randomly initialized) it is very likely possible to almost connect a random parameter value to the global minimum with a path which along the way does not need to climb much higher than the initial risk.

In~\cite{venturi2019spurious}, a similar approach is taken and the convexity in the last layer is used. However, the authors invoke the concept of intrinsic dimension to elegantly solve the non-linearity of $\prisk((\theta^{(1)}, \theta^{(2)}))$ with respect to $\theta^{(1)}$. Additionally,~\cite{safran2016quality} constructs a path of decreasing risk from random initializations. The idea here is that if one starts at a point of sufficiently high risk, one can always find a path to the global optimum with strictly decreasing risk. The intriguing insight behind this result is that if the initialization is sufficiently bad, i.e., worse than that of a NN outputting only zero, then there exist two operations that influence the risk directly. Multiplying the last layer with a number smaller than one will decrease the risk, whereas the opposite will increase it. Using this tuning mechanism, any given potentially non-monotone path from the initialization to the global minimum can be modified so that it is strictly monotonically decreasing. 
In a similar spirit,~\cite{nguyen2017loss} shows that if a deep NN has a layer with more neurons than training data points, then under certain assumptions the training data will typically be mapped to linearly independent points in that layer. Of course, this layer could then be composed with a linear map that maps the linearly independent points to any desirable output, in particular one that achieves vanishing empirical risk, see also Proposition~\ref{prop:interpolation}. As for two-layer NNs, the previous discussion on linear paths immediately shows that in this situation a monotone path to the global minimum exists. 

\subsection{Lazy training and provable convergence of stochastic gradient descent}
\label{subsec:lazy_training}
\begin{figure}[t]
    \centering
    \includegraphics[width = 0.99 \textwidth]{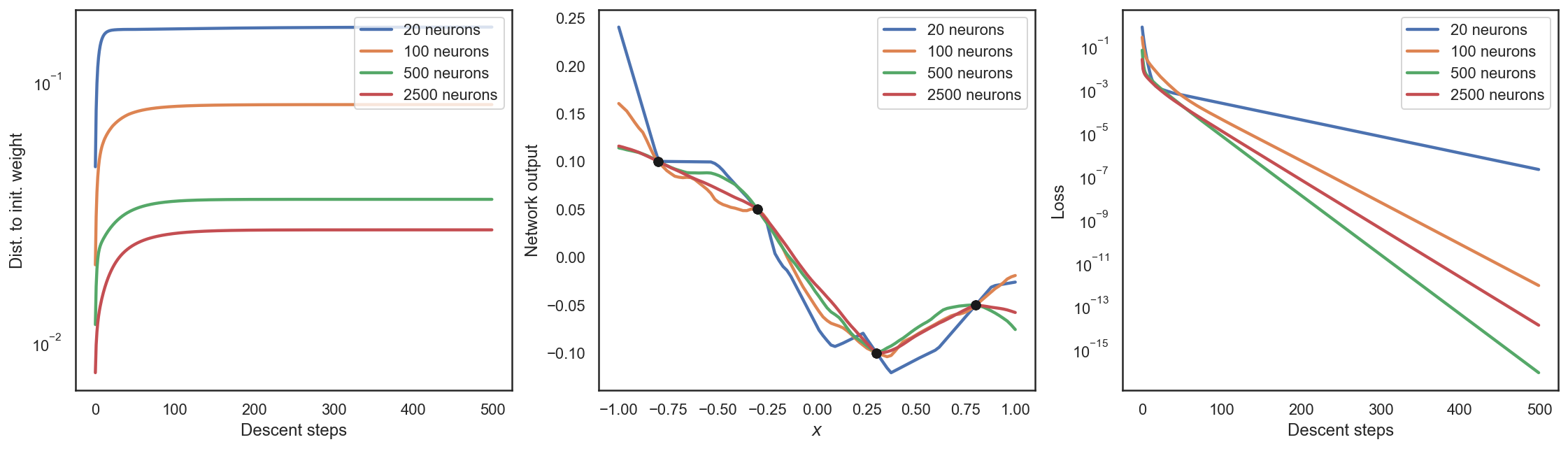}
    \caption{Four networks with architecture $((1,n,n,1),\ReLU)$ and $n \in \{20, 100, 500, 2500\}$ neurons per hidden layer were trained by gradient descent to fit four points that are shown in the middle figure as black dots.
    We depict on the left the relative Euclidean distance of the parameters from the initialization through the training process. In the middle, we show the final trained NNs. On the right we show the behavior of the training error.}
    \label{fig:trainingOfOverparametrisedNNs}
\end{figure}

When training highly overparametrized NNs, one often observes that the parameters of the NNs barely change during training. In Figure~\ref{fig:trainingOfOverparametrisedNNs}, we show the relative distance that the parameters travel through the parameter space during the training of NNs of varying numbers of neurons per layer. 

The effect described above has been observed repeatedly and theoretically explained, see, e.g.,~\cite{du2018gradient, li2018learning, allen2019convergence, du2019gradient, zou2020gradient}.
In Subsection~\ref{subsec:ntk}, we have already seen a high-level overview and, in particular, the function space perspective of this phenomenon in the infinite width limit.
Below we present a short and highly simplified derivation of this effect and show how it leads to provable convergence of gradient descent for sufficiently overparametrized deep NNs.

\paragraph{A simple learning model:} 
We consider again the simple NN model of~\eqref{eq:simpleNN} with a smooth activation function $\varrho$ which is not affine linear. For the quadratic
loss and training data $\cs = ((x^{(i)}, y^{(i)}))_{i=1}^m \in (\R^d \times \R)^m$, where $x_i \neq x_j$ for all $i \neq j$, the empirical risk is given by
\begin{equation*}
    \prisk(\theta)=\erisk{\cs}(\theta) = \frac{1}{m} \sum_{i=1}^m ( \Phi(x^{(i)}, \theta) - y^{(i)} )^2.
\end{equation*}
Let us further assume that $\Theta^{(1)}_{j} \sim \cN(0,1/n)^{d+1}$, $j \in [n]$, and $\Theta^{(2)}_j \sim \cN(0,1/n)$, $j \in [n]$, are independent random variables.

\paragraph{A peculiar kernel:} Next, we would like to understand how the gradient $\nabla_\theta \prisk(\Theta)$ looks like with high probability over the initialization $\Theta=(\Theta^{(1)}, \Theta^{(2)})$. Similar to~\eqref{eq:risk_flow}, we have by restricting the gradient to $\theta^{(2)}$ and applying the chain rule that
\begin{equation}
\begin{split}
    \|\nabla_\theta \prisk(\Theta)\|_2^2 
    &\geq \frac{4}{m^2} \Big\| \sum_{i = 1}^m  \nabla_{\theta^{(2)}} \Phi(x^{(i)}, \Theta)(\Phi(x^{(i)}, \Theta) - y^{(i)} ) \Big\|_2^2 \\
    & = \frac{4}{m^2} \big((\Phi(x^{(i)}, \Theta) - y^{(i)} )_{i=1}^m\big)^T \bar{K}_\Theta (\Phi(x^{(j)}, \Theta) - y^{(j)})_{j=1}^m, \label{eq:applySomethingToThis}
\end{split}
\end{equation}
where $\bar{K}_\Theta$ is a random $\R^{m\times m}$-valued kernel given by
\begin{equation*}
    (\bar{K}_\Theta)_{i,j} \coloneqq \big(\nabla_{\theta^{(2)}} \Phi(x^{(i)}, \Theta)\big)^T \nabla_{\theta^{(2)}} \Phi(x^{(j)}, \Theta), \quad i,j\in [m].
\end{equation*}
This kernel is closely related to the neural tangent kernel in~\eqref{eq:ntk} evaluated at the features $(x^{(i)})_{i=1}^m$ and the random initialization $\Theta$. It is a slightly simplified version thereof, as in~\eqref{eq:ntk} the gradient is taken with respect to the full vector $\theta$. This can also be regarded as the kernel associated with a \emph{random features model}~\cite{rahimi2007random}.

Note that for our two-layer NN we have that
\begin{equation}
    \big(\nabla_{\theta^{(2)}} \Phi(x, \Theta)\big)_k= \varrho\left(\left\langle \Theta^{(1)}_k, \begin{bmatrix}x \\  1 \end{bmatrix}\right\rangle\right), \quad x\in\R^d, \  k\in[n].
\end{equation}
Thus, we can write $\bar{K}_\Theta$ as the following sum of (random) rank one matrices:
\begin{equation}
\label{eq:kerneliidsum}
    \bar{K}_\Theta = \sum_{k = 1}^n v_k v_k^T \qquad \text{with} \quad v_k = \left(\varrho\left(\left\langle \Theta^{(1)}_k, \begin{bmatrix}x^{(i)} \\  1 \end{bmatrix}\right\rangle\right)\right)_{i =1}^m \in \R^m, \quad k\in[n].
\end{equation}
The kernel $\bar{K}_\Theta$ are symmetric and positive semi-definite by construction. 
It is positive definite if it is non-singular, i.e., if at least $m$ of the $n$ vectors $v_k$, $k \in [n]$, are linearly independent.
Proposition~\ref{prop:interpolation} shows that for $n = m$ the probability of that event is not zero, say $\delta$, and is therefore at least $1-(1-\delta)^{\lfloor n/m \rfloor}$ for arbitrary $n$. In other words, the probability increases rapidly with $n$. It is also clear from~\eqref{eq:kerneliidsum} that $\E[\bar{K}_\Theta]$ scales linearly with $n$.

From this intuitive derivation, we conclude that for sufficiently large $n$, with high probability $\bar{K}_\Theta$ is a positive definite kernel with smallest eigenvalue $\lambda_{\min}(\bar{K}_\Theta)$ scaling linearly with $n$. The properties of $\bar{K}_\Theta$, in particular its positive definiteness, have been studied much more rigorously as already described in Subsection~\ref{subsec:ntk}.

\paragraph{Control of the gradient:} Applying the expected behavior of the smallest eigenvalue $\lambda_{\min}(\bar{K}_\Theta)$ of $\bar{K}_\Theta$ to~\eqref{eq:applySomethingToThis}, we conclude that with high probability
\begin{equation} \label{eq:ThisRules}
    \|\nabla_\theta \prisk(\Theta)\|_2^2 \geq \frac{4}{m^2} \lambda_{\min}(\bar{K}_\Theta) \|(\Phi(x^{(i)}, \Theta) - y^{(i)} )_{i=1}^m\|_2^2 \gtrsim \frac{n}{m} \prisk(\Theta). 
\end{equation}

To understand what will happen when applying gradient descent, we first need to understand how the situation changes in a neighborhood of $\Theta$. We fix $x\in\R^d$ and observe that by the mean value theorem for all $\bar\theta \in B_1(0)$ we have 
\begin{equation}
    \left\|\nabla_\theta \Phi(x, \Theta) - \nabla_{\theta} \Phi(x, \Theta+\bar\theta)\right\|_2^2 \lesssim \sup_{\hat\theta \in B_1(0)} \big\| \nabla^2_{\theta} \Phi(x, \Theta+\hat{\theta})\big\|_{\mathrm{op}}^2,\label{eq:LossAtDifferentPoints}
\end{equation}
where $\| \nabla^2_\theta \Phi(x, \Theta+ \hat\theta)\|_{\mathrm{op}}$ denotes the operator norm of the Hessian of $\Phi(x, \cdot)$ at $\Theta+\hat\theta$. From inspection of~\eqref{eq:simpleNN}, it is not hard to see that for all $i, j \in [n]$ and $k, \ell \in [d + 1]$ 
\begin{equation*}
\E\Bigg[\Big(\frac{\partial^2 \Phi(x, \Theta)}{\partial \theta^{(2)}_i \partial \theta^{(2)}_j}  \Big)^2 \Bigg] = 0, \quad \E\Bigg[\Big(\frac{\partial^2  \Phi(x, \Theta)}{\partial \theta^{(2)}_i \partial (\theta^{(1)}_j)_k} \Big)^2 \Bigg] \lesssim \delta_{i,j}, \quad \text{and}  \quad \E\Bigg[ \Big(\frac{\partial^2\Phi(x, \Theta)}{\partial (\theta^{(1)}_i)_k \partial (\theta^{(1)}_j)_\ell}  \Big)^2 \Bigg] \lesssim \frac{\delta_{i,j}}{n},
\end{equation*}
where $\delta_{i,j} = 0$ if $i \neq j$ and $\delta_{i,i} = 1$ for all $i,j \in [n]$. For sufficiently large $n$, we have that $\nabla^2_\theta \Phi(x, \Theta)$ is in expectation approximately a block band matrix with band-width $d+1$. Therefore, we conclude that $\E \big[\| \nabla^2_\theta \Phi(x, \Theta)\|_{\mathrm{op}}^2 \big]\lesssim 1$. Hence, we obtain by concentration of Gaussian random variables that with high probability $\| \nabla^2_\theta \Phi(x, \Theta)\|_{\mathrm{op}}^2 \lesssim 1$. 
By the block-banded form of $\nabla^2_\theta \Phi(x, \Theta)$ we have that, even after perturbation of $\Theta$ by a vector $\hat\theta$ with norm bounded by $1$, the term $\| \nabla^2_\theta \Phi(x, \Theta+\hat\theta)\|_{\mathrm{op}}^2$ is bounded, which yields that the right-hand side of~\eqref{eq:LossAtDifferentPoints} is bounded with high probability. 

Using~\eqref{eq:LossAtDifferentPoints}, we can extend~\eqref{eq:ThisRules}, which holds with high probability, to a neighborhood of $\Theta$
by the following argument: Let $\bar{\theta} \in B_1(0)$, then
\begin{equation}
\begin{split}
    \|\nabla_\theta \prisk(\Theta+\bar{\theta})\|_2^2 
    &\ge \frac{4}{m^2} \Big\| \sum_{i = 1}^m  \nabla_{\theta^{(2)}} \Phi(x^{(i)}, \Theta+\bar{\theta})(\Phi(x^{(i)}, \Theta+\bar{\theta}) - y^{(i)} ) \Big\|_2^2 \\
    &\! \! \underset{ \eqref{eq:LossAtDifferentPoints}}{=}  \frac{4}{m^2}\Big\|\sum_{i=1}^m(\nabla_{\theta^{(2)}} \Phi(x^{(i)}, \Theta) + \cO(1)) (\Phi(x^{(i)}, \Theta+\bar{\theta}) - y^{(i)})  \Big\|_2^2 \label{eq:lowerPerturbationBound}\\
    & \underset{(*)}{\gtrsim}
    \frac{1}{m^2} (\lambda_{\min}(\bar{K}_\Theta)  +  \cO(1)) \|(\Phi(x^{(i)}, \Theta+\bar{\theta}) - y^{(i)} )_{i=1}^m\|_2^2\\
     &\gtrsim \frac{n}{m} \prisk(\Theta+\bar{\theta}),
\end{split}
\end{equation}
where the estimate marked by $(*)$ uses the positive definiteness of $\bar{K}_\Theta$ again and only holds for sufficiently large $n$, so that the $\cO(1)$ term is negligible.

We conclude that, with high probability over the initialization $\Theta$, on a ball of fixed radius around $\Theta$ the squared Euclidean norm of the gradient of the empirical risk is lower bounded by $\frac{n}{m}$ times the empirical risk. 

\paragraph{Exponential convergence of gradient descent:} For sufficiently small step sizes $\eta$, the observation in the previous paragraph yields the following convergence rate for gradient descent as in Algorithm~\ref{alg:1}, specifically~\eqref{eq:SGDForERM}, with $m'=m$ and $\Theta^{(0)} = \Theta$:
If $\|\Theta^{(k)} - \Theta\|\le 1$ for all $k \in [K+1]$, then\footnote{Note that the step-size $\eta$ needs to be small enough to facilitate the approximation step in~\eqref{eq:ExponentialConvergence}. Hence, we cannot simply put $\eta = m/(cn)$ in~\eqref{eq:ExponentialConvergence} and have convergence after one step.} 
\begin{equation}
\prisk(\Theta^{(K+1)}) \approx \prisk(\Theta^{(K)}) - \eta \| \nabla_{\theta}\prisk(\Theta^{(K)})\|_2^2 \leq \Big(1 -  \frac{c \eta n}{m} \Big) \prisk(\Theta^{(K)}) \lesssim \Big(1 -  \frac{c \eta n}{m} \Big)^K,  \label{eq:ExponentialConvergence}
\end{equation}
for $c\in(0,\infty)$ so that $\|\nabla_{\theta} \prisk(\Theta^{(k)})\|_2^2 \geq \frac{c n}{m} \prisk(\Theta^{(k)})$ for all $k\in[K]$. 

Let us assume without proof that the estimate~\eqref{eq:lowerPerturbationBound} could be extended to an equivalence. In other words, we assume that we additionally have that $\|\nabla_\theta \prisk(\Theta+\bar{\theta})\|_2^2  \lesssim \frac{n}{m} \prisk(\Theta+\bar{\theta})$. This, of course, could be shown with similar tools as were used for the lower bound. Then we have that $\|\Theta^{(k)} - \Theta\|_2\le 1$ for all $k \lesssim \sqrt{m/(\eta^2 n)}$. Setting $t = \sqrt{m/(\eta^2 n)}$ and using the limit definition of the exponential function, i.e., $\lim_{t\to \infty} (1 - x/t)^t = e^{-x}$, yields for sufficiently small $\eta$ that~\eqref{eq:ExponentialConvergence} is bounded by $e^{-c \sqrt{n/m}}$. 

We conclude that, with high probability over the initialization,
\emph{gradient descent converges with an exponential rate to an arbitrary small empirical risk if the width $n$ is sufficiently large. In addition, the iterates of the descent algorithm even stay in a small fixed neighborhood of the initialization during training}.
Because the parameters only move very little, this type of training has also been coined lazy training~\cite{chizat2019lazy}. 

Similar ideas as above, have led to groundbreaking convergence results of SGD for overparametrized NNs in much more complex and general settings, see, e.g.,~\cite{du2018gradient, li2018learning, allen2019convergence}.

In the infinite width limit, NN training is practically equivalent to kernel regression, see Subsection~\ref{subsec:ntk}. If we look at Figure~\ref{fig:trainingOfOverparametrisedNNs} we see that the most overparametrized NN interpolates the data like a kernel-based interpolator would.
In a sense, which was also highlighted in~\cite{chizat2019lazy}, this shows that, while overparametrized NNs in the lazy training regime have very nice properties, they essentially act like linear methods.

\section{Tangible effects of special architectures}
\label{sec:architectures}
In this section, we describe results that isolate the effects of certain aspects of NN architectures. As we have discussed in Subsection~\ref{subsec:NewTheories}, typically only either the depth or the number of parameters are used to study theoretical aspects of NNs. We have seen instances of this throughout Sections~\ref{sec:expressivity} and~\ref{sec:CurseOfDimension}. Moreover, also in Section~\ref{sec:optimizationNew}, we saw that wider NNs enjoy certain very favorable properties from an optimization point of view. 

Below, we introduce certain specialized NN architectures. We start with one of the most widely used types of NNs, the \emph{convolutional neural network} (CNN). In Subsection~\ref{subsec:ResNet} we introduce \emph{skip connections} and in Subsection~\ref{subsec:framelet} we discuss a specific class of CNNs equipped with an encoder-decoder structure that are frequently used in image processing techniques. We introduce the \emph{batch normalization block} in Subsection~\ref{subsec:BatchNorm}. Then, we discuss \emph{sparsely connected} NNs that typically result as an extraction from fully connected NNs in Subsection~\ref{subsec:sparse}. Finally, we briefly comment on {recurrent neural networks} in Subsection~\ref{subsec:RNNs}.

As we have noted repeatedly throughout this
\ifbook
\else%
manuscript,
\fi%
it is impossible to give a full account of the literature in a short introductory
\ifbook
\else%
article.
\fi%
In this section, this issue is especially severe since the number of special architectures studied in practice is enormous. Therefore, we had to omit many very influential and widely used neural network architectures. Among those are \emph{graph neural networks}, which handle data from non-Euclidean input spaces. We refer to the survey articles~\cite{bronstein2017geometric, wu2021comprehensive} for a discussion. Another highly successful type of architectures are \emph{(variational) autoencoders}~\cite{ackley1985learning,hinton1994autoencoders}. These are neural networks with a bottleneck that enforce a more efficient representation of the data. Similarly, \emph{generative adversarial networks}~\cite{goodfellow2014generative} which are composed of two neural networks, one generator and one discriminator, could not be discussed here. Another widely used component of architectures used in practice is the so-called \emph{dropout layer}. This layer functions through removing some neurons randomly during training. This procedure empirically prevents overfitting. An in-detail discussion of the mathematical analysis behind this effect is beyond the scope of this
\ifbook
\else%
manuscript.
\fi%
We refer to~\cite{wan2013regularization, srivastava2014dropout, haeffele2017global, mianjy2018implicit} instead. Finally, the very successful \emph{attention mechanism}~\cite{bahdanau2014neural, vaswani2017attention}, that is the basis of \emph{transformer neural networks}, had to be omitted.

Before we start describing certain effects of special NN architectures, a word of warning is required. The special building blocks, which will be presented below, have been developed based on a specific need in applications and are used and combined in a very flexible way. To describe these tools theoretically without completely inflating the notational load, some simplifying assumptions need to be made. It is very likely that the simplified building blocks do not accurately reflect the practical applications of these tools in all use cases. 

\subsection{Convolutional neural networks}
\label{subsec:convnets}

Especially for very high-dimensional inputs where the input dimensions are spatially related, fully connected NNs seem to require unnecessarily many parameters. For example, in image classification problems, neighboring pixels very often share information and the spatial proximity should be reflected in the architecture.
Based on this observation, it appears reasonable to have NNs that have local receptive fields in the sense that they collect information jointly from spatially close inputs. 
In addition, in image processing, we are not necessarily interested in a universal hypothesis set. A good classifier is invariant under many operations, such as translation or rotation of images. It seems reasonable to hard-code such invariances into the architecture. 

These two principles suggest that the receptive field of a NN should be the same on different translated patches of the input. In this sense, parameters of the architecture can be reused. Together, these arguments make up the three fundamental principles of convolutional NNs: local receptive fields, parameter sharing, and equivariant representations, as introduced in~\cite{lecun1989backpropagation}. We will provide a mathematical formulation of convolutional NNs below and then revisit these concepts.

A convolutional NN corresponds to multiple convolutional blocks, which are special types of layers. 
For a group $\gr$, which typically is either $[d] \cong \Z/(d\Z)$ or $[d]^2 \cong (\Z/(d\Z))^2$ for $d \in \N$, depending on whether we are performing one-dimensional or two-dimensional convolutions, the convolution of two vectors $a,b \in \R^\gr$ is defined as 
\begin{equation*}
    (a*b)_i = \sum_{j \in \gr} a_j b_{j^{-1} i}, \quad i\in \gr.
\end{equation*}
Now we can define a \emph{convolutional block} as follows: Let $\widetilde{\gr}$ be a subgroup of $\gr$, let $p: \gr \to \widetilde{\gr}$ be a so-called \emph{pooling-operator}, and let $C\in\N$ denote the number of channels. Then, for a series of kernels $\kappa_i\in \R^\gr$, $i\in[C]$, the output of a convolutional block is given by 
\begin{equation}\label{eq:ConvBlock}
    \R^\gr \ni x \mapsto x' \coloneqq (p(x * \kappa_i))_{i=1}^C \in (\R^{\widetilde{\gr}})^{C}.
\end{equation}
A typical example of a pooling operator is for $\gr = (\Z/(2d\Z))^2$ and $\widetilde{\gr} = (\Z/(d\Z))^2$ the $2 \times 2$ subsampling operator 
\begin{equation*}
    p: \R^\gr \to \R^{\widetilde{\gr}}, \quad x 
    \mapsto (x_{2i-1,2j-1})_{i,j =1}^d.
\end{equation*}
Popular alternatives are average pooling or max pooling. These operations then either pass the average or the maximum over patches of similar size. The convolutional kernels correspond to the aforementioned receptive fields. They can be thought of as local if they have small supports, i.e., few nonzero entries.

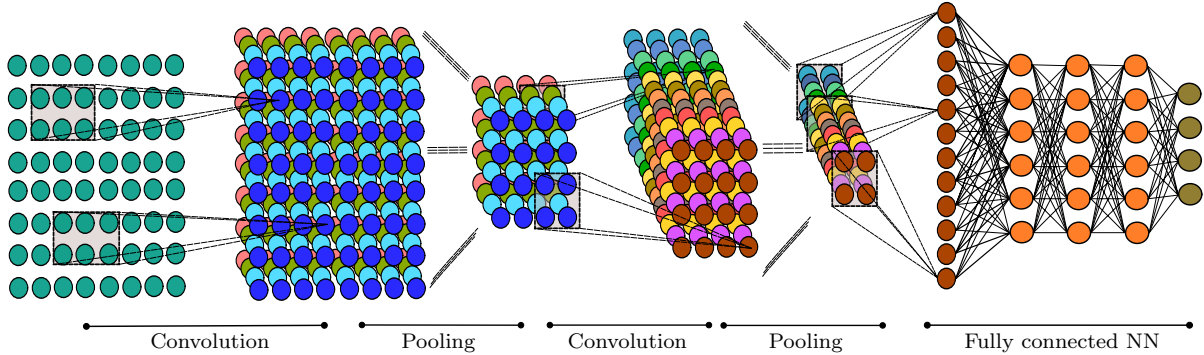
\begin{figure}[t]
    \centering
    \includegraphics[height= 0.2\textheight]{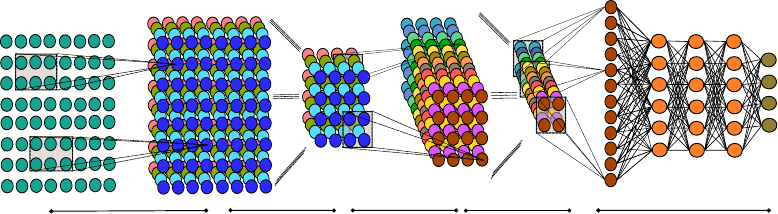}
    \put(-398, -7){\footnotesize Convolution}
    \put(-303, -7){\footnotesize Pooling}
    \put(-240, -7){\footnotesize Convolution}
    \put(-164, -7){\footnotesize Pooling}
    \put(-90, -7){\footnotesize Fully connected NN}
    \caption{Illustration of a convolutional neural network with two-dimensional convolutional blocks and $2 \times 2$ subsampling as pooling operation.}
    \label{fig:CNN}
\end{figure}

As explained earlier, a convolutional NN is built by stacking multiple convolutional blocks after another\footnote{We assume that the definition of a convolutional block is suitably extended to input data in the Cartesian product $(\R^G)^C$. For instance, one can take an affine linear combination of $C$ mappings as in~\eqref{eq:ConvBlock} acting on each coordinate. Moreover, one may also interject an activation function between the blocks.}. At some point, the output can be \emph{flattened}, i.e., mapped to a vector and is then fed into a FC NN (see Definition~\ref{def:classical_nns}). We depict this setup in Figure~\ref{fig:CNN}.

Owing to the fact that convolution is a linear operation, depending on the pooling operation, one may write a convolutional block~\eqref{eq:ConvBlock} as a FC NN. For example, if $\gr = (\Z/(2d\Z))^2$ and the $2\times 2$ subsampling pooling operator is used, then the convolutional block could be written as 
$x \mapsto W x$ for a block circulant matrix $W \in \R^{ (C d^2) \times (2d)^2}$. Since we require $W$ to have a special structure, we can interpret a convolutional block as a special, restricted feed-forward architecture.

After these considerations, it is natural to ask how the restriction of a NN to a pure convolutional structure, i.e., consisting only of convolutional blocks, will affect the resulting hypothesis set. The first natural question is whether the set of such NNs is still universal in the sense of Theorem~\ref{theo:universal}.
The answer to this question depends strongly on the type of pooling and convolution that is allowed. If the convolution is performed with padding, then the answer is yes~\cite{oono2019approximation, zhou2020universality}. On the other hand, for circular convolutions and without pooling, universality does not hold, but the set of translation equivariant functions can be universally approximated~\cite{yarotsky2018universal, petersen2020equivalence}. Furthermore,~\cite{yarotsky2018universal} illuminates the effect of subsample pooling by showing that, if no pooling is applied, then universality cannot be achieved, whereas if pooling is applied then universality is possible. The effect of subsampling in CNNs from the viewpoint of approximation theory is further discussed in~\cite{zhou2020theory}. The role of other types of pooling in enhancing invariances of the hypothesis set will be discussed in Subsection~\ref{subsec:wavelets} below. 

\subsection{Residual neural networks} \label{subsec:ResNet}
Let us first illustrate a potential obstacle when training deep NNs.
Consider for $L \in \N$ the product operation 
\begin{equation*}
\R^L \ni x \mapsto \pi(x) = \prod_{\ell=1}^L x_{\ell}.
\end{equation*}
It is clear that 
\begin{equation*}
\frac{\partial}{\partial x_k} \pi(x) = \prod_{\ell\neq k}^L x_{\ell}, \quad x\in \R^L.
\end{equation*}
Therefore, for sufficiently large $L$, we expect that $\big|\frac{\partial \pi }{\partial x_k}\big|$ will be exponentially small, if $|x_\ell|< \lambda < 1$ for all $\ell \in [L]$ or exponentially large, if $|x_\ell| > \lambda > 1$ for all $\ell \in [L]$. The output of a general NN, considered as a directed graph, is found by repeatedly multiplying the input with parameters in every layer along the paths that lead from the input to the output neuron. Due to the aforementioned phenomenon, it is often observed that training NNs suffers from either the exploding or the vanishing gradient problem, which may prevent lower layers from training at all. The presence of an activation function is likely to exacerbate this effect. The exploding or vanishing gradient problem seems to be a serious obstacle towards efficient training of deep NNs.

In addition to the vanishing and exploding gradient problems, there is an empirically observed \emph{degradation problem}~\cite{he2016deep}. This phrase describes the phenomenon that FC NNs seem to achieve lower accuracy on both the training and test data when increasing their depth.

From an approximation theoretic perspective, deep NNs should always be superior to shallow NNs. The reason for this is that NNs with two layers can either exactly represent the identity map or approximate it arbitrarily well. 
Concretely, for the ReLU activation function $\ReLU$ we have that $x = \ReLU(x + b) - b$ for $x \in \R^d$ with $x_i > - b_i$, where $b\in \R^d$. In addition, for any activation function $\varrho$ which is continuously differentiable on a neighborhood of some point $\lambda\in\R$ with $\varrho'(\lambda)\neq 0$ one can approximate the identity arbitrary well, see~\eqref{eq:representationofMonomials}.
Because of this, extending a NN architecture by one layer can only enlarge the associated hypothesis set. 

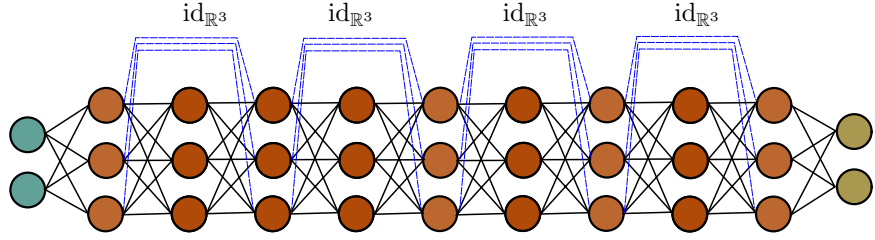
\begin{figure}[t]
\centering
\includegraphics[height= 0.12\textheight]{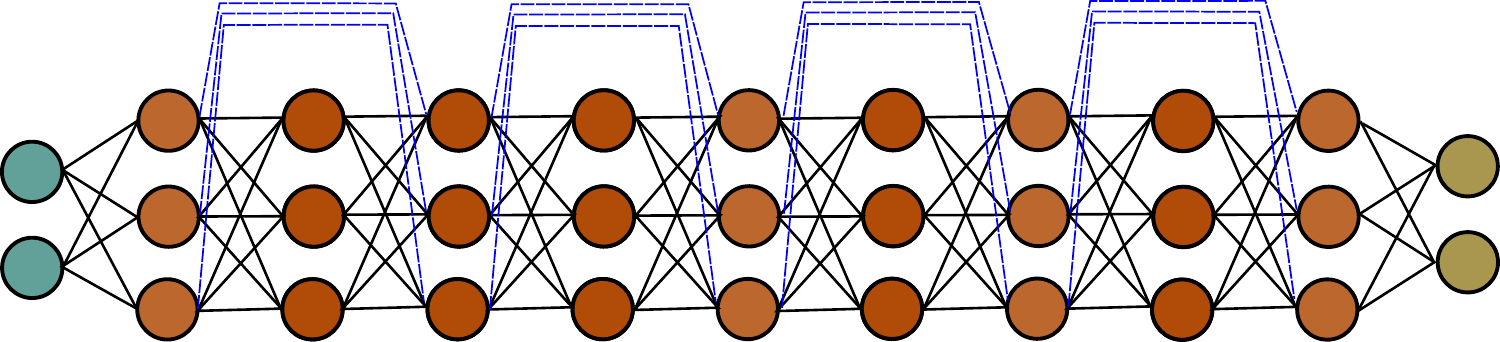}
\put(-261, 80){$\mathrm{id}_{\R^3}$}
\put(-206, 80){$\mathrm{id}_{\R^3}$}
\put(-140, 80){$\mathrm{id}_{\R^3}$}
\put(-75, 80){$\mathrm{id}_{\R^3}$}
\caption{Illustration of a neural network with residual blocks.}
\label{fig:ResNet}
\end{figure}

Therefore, one may expect that the degradation problem is more associated with the optimization aspect of learning. This problem is addressed by a small change to the architecture of a feed-forward NN in~\cite{he2016deep}. Instead of defining a FC NN $\Phi$ as in~\eqref{eq:def_nn_classical}, one can insert a residual block in the $\ell$-th layer by redefining\footnote{One can also skip multiple layers, e.g., in~\cite{he2016deep} two or three layers skipped, use a simple transformation instead of the identity~\cite{srivastava2015training}, or randomly drop layers~\cite{huang2016deep}.}
\begin{equation}
\label{eq:resBlock}
    \bar{\Phi}^{(\ell)}(x,\theta) = \varrho(\Phi^{(\ell)}(x,\theta))+ \bar{\Phi}^{(\ell-1)}(x,\theta),
\end{equation}
where we assume that $N_{\ell}=N_{\ell-1}$.
Such a block can be viewed as the sum of a regular FC NN and the identity which is referred to as skip connection or \emph{residual connection}. A sketch of a NN with residual blocks is shown in Figure~\ref{fig:ResNet}. 
Inserting a residual block in all layers leads to a so-called \emph{residual NN}. 

A prominent approach to analyze residual NNs is by establishing a connection with optimal control problems and dynamical systems~\cite{weinan2017proposal, thorpe2018deep, weinan2019mean, li2019deep, ruthotto2019deep, lu2020mean}. Concretely, if each layer of a NN $\Phi$ is of the form~\eqref{eq:resBlock}, then we have that
\begin{equation*}
    \bar{\Phi}^{(\ell)} - \bar{\Phi}^{(\ell-1)} = \varrho(\Phi^{(\ell)}) \eqqcolon h(\ell, \Phi^{(\ell)}),
\end{equation*}
where we abbreviate $\bar{\Phi}^{(\ell)}=\bar{\Phi}^{(\ell)}(x,\theta)$ and set $\bar{\Phi}^{(0)}=x$.
Hence, $(\bar{\Phi}^{(\ell)})_{\ell=0}^{L-1}$ corresponds to an Euler discretization of the ODE 
\begin{equation*}
\dot{\phi}(t) = h(t, \phi(t)), \qquad \phi(0) = x,
\end{equation*}
where $t \in [0,L-1]$ and $h$ is an appropriate function.

Using this relationship, deep residual NNs can be studied in the framework of the well-established theory of dynamical systems, where strong mathematical guarantees can be derived.

\subsection{Framelets and U-Nets} \label{subsec:framelet}

\begin{figure}[t]
    \centering
    \includegraphics[width= 0.99\textwidth]{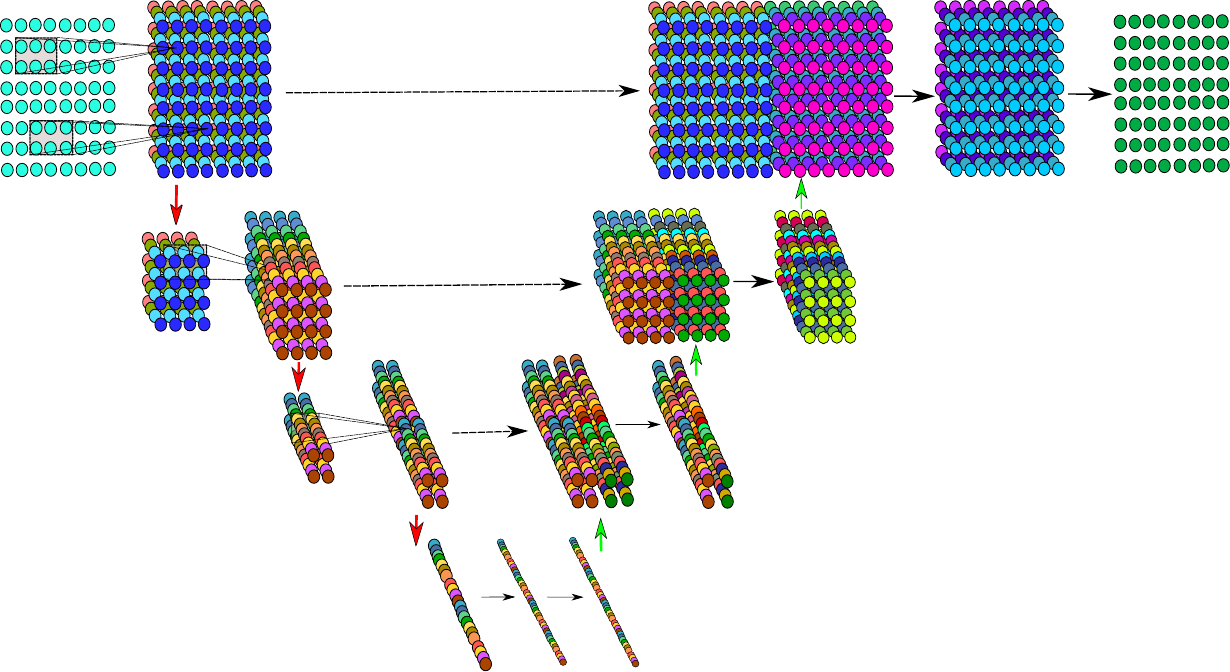}
    \caption{Illustration of a simplified U-net neural network. Down-arrows stand for pooling, up arrows for deconvolution or upsampling, right arrows for convolution or fully connected steps. Dashed lines are skip connections.}
    \label{fig:UNet}
\end{figure}

One of the most prominent application areas of deep NNs are inverse problems, particularly those in the field of imaging science, see also Subsection~\ref{subsec:inverseproblems}. A specific architectural design of CNNs, namely so-called \emph{U-nets} introduced in~\cite{Unet2015}, seems to perform best for this range of problems. We depict a sketch of a U-net in Figure~\ref{fig:UNet}. 
However, a theoretical understanding of the success of this architecture was lacking. 

Recently, an innovative approach called \emph{deep convolutional framelets} was suggested in~\cite{Ye2018Framelets}, which we now briefly explain. The core idea is to take a frame-theoretic viewpoint, see, e.g.,~\cite{CKPIntroFrames}, and regard the forward pass of a CNN as a decomposition in terms of a frame (in the sense of a generalized basis). A similar approach will be taken in Subsection~\ref{subsec:sparsecoding} for understanding the learned kernels using sparse coding. However, based on the analysis and synthesis operators of the corresponding frame, the usage of deep convolutional framelets naturally leads to a theoretical understanding of encoder-decoder architectures, such as U-nets.

Let us describe this approach for one-dimensional convolutions on the group $G\coloneqq\Z/(d\Z)$ with kernels defined on the subgroup $H\coloneqq \Z/(n\Z)$, where $d,n\in\N$ with $n <d$, see also Subsection~\ref{subsec:convnets}. 
We define the convolution between $u\in\R^G$ and $v\in\R^H$ by zero-padding $v$, i.e., $g\ast_{\circ} v \coloneqq g\ast \bar{v}$, where $\bar{v}\in\R^G$ is defined by $\bar{v}_i= v_i$ for $i\in H$ and $\bar{v}_i= 0$ else.
As an important tool, we consider the
Hankel matrix
$
\mathbb{H}_n(x) = (x_{i+j})_{i\in G, j\in H} \in \R^{d \times n} $
associated with $x \in \mathbb{R}^G$.
As one key property, matrix-vector multiplications with Hankel matrices are translated to convolutions via\footnote{Here and in the following we naturally identify elements in $\R^G$ and $\R^H$ with the corresponding vectors in $\R^d$ and $\R^n$.}
\begin{equation}
\label{eq:hankel_conv}
    \langle e^{(i)} , \mathbb{H}_n(x)v\rangle = \sum_{j\in H} x_{i+j} v_j  = \langle x, e^{(i)} \ast_{\circ} v \rangle, \quad i \in G,
\end{equation}
where $e^{(i)}\coloneqq \mathds{1}_{\{i\}}\in\R^G$ and $v\in\R^H$, see~\cite{yin2017tale}. 
Further, we can recover the $k$-th coordinate of $x$ by the Frobenius inner product between $\mathbb{H}_n(x)$ and the Hankel matrix associated with $e^{(k)}$, i.e.,
\begin{equation}
\label{eq:hankel_coord}
    \frac{1}{n}\trace\big(\mathbb{H}_n(e^{(k)})^T\mathbb{H}_n(x)\big)=\frac{1}{n} \sum_{j \in H} \sum_{i\in G} e^{(k)}_{i+j} x_{i+j}= \frac{1}{n} |H|  x_{k} = x_k.
\end{equation}
This allows us to construct global and local bases as follows: 
Let $p,q \in \N$, let $U=\begin{bmatrix} u_1 \cdots u_p \end{bmatrix} \in \mathbb{R}^{d \times p}$, $V=\begin{bmatrix} v_1 \cdots v_q \end{bmatrix} \in \mathbb{R}^{n \times q}$,  $\widetilde{U}=\begin{bmatrix} \tilde{u}_1 \cdots \tilde{u}_p\end{bmatrix} \in \mathbb{R}^{d \times p}$, and $\widetilde{V}=\begin{bmatrix}\tilde{v}_1 \cdots \tilde{v}_q \end{bmatrix} \in \mathbb{R}^{n \times q}$, and assume that
\begin{equation}
\label{eq:identity_hankel}
    \mathbb{H}_n(x) = \widetilde{U}U^T \mathbb{H}_n(x)V \widetilde{V}^T.
\end{equation}
For $p\ge d$ and $q\ge n$, this is, for instance, satisfied if $U$ and $V$ constitute frames with $\widetilde{U}$ and $\widetilde{V}$ being their respective dual frames, i.e., $\widetilde{U}U^T = \mathrm{I}_d$ and $V \widetilde{V}^T = \mathrm{I}_n$. As a special case, one can consider orthonormal bases $U=\widetilde{U}$ and $V=\widetilde{V}$ with $p=d$ and $q=n$. In the case $p=q=r \le n$, where $r$ is the rank of $\mathbb{H}_n(x)$, one can establish~\eqref{eq:identity_hankel} by choosing the left and right singular vectors of $\mathbb{H}_n(x)$ as $U=\widetilde{U}$ and $V=\widetilde{V}$, respectively.
The identity in~\eqref{eq:identity_hankel}, in turn, ensures the following decomposition:
\begin{equation}
\label{eq:PR}
x = \frac{1}{n} \sum_{i=1}^p \sum_{j=1}^q \langle x , u_i \ast_{\circ} v_j \rangle  \tilde{u}_i \ast_{\circ} \tilde{v}_j.
\end{equation}
Observing that the vector $v_j \in \mathbb{R}^H$ interacts locally with $x\in\R^G$ due to the fact that $H \subset G$, whereas $u_i\in \mathbb{R}^G$ acts on the entire vector $x$, we refer to $(v_j)_{j=1}^q$ as local and $(u_i)_{i=1}^p$ as global bases. In the context of CNNs, $v_i$ can be interpreted as local convolutional kernel and $u_i$ as pooling operation\footnote{Note that $\langle x, u_i \ast_{\circ} v_j \rangle$ can also be interpreted as $\langle u_i, v_j \star x \rangle$, where $\star$ denotes the cross-correlation between the zero-padded $v_j$ and $x$. This is in line with software implementations for deep learning applications, e.g., TensorFlow and PyTorch, where typically cross-correlations are used instead of convolutions.}.
The proof of~\eqref{eq:PR} follows directly from properties~\eqref{eq:hankel_conv},~\eqref{eq:hankel_coord}, and~\eqref{eq:identity_hankel} as
\begin{equation*}
    x_k = \frac{1}{n} \trace\big(\mathbb{H}_n(e^{(k)})^T\mathbb{H}_n(x)\big) = \frac{1}{n} \trace\big(\mathbb{H}_n(e^{(k)})^T \widetilde{U}U^T \mathbb{H}_n(x)V \widetilde{V}^T\big) = \frac{1}{n}\sum_{i=1}^p \sum_{j=1}^q \langle u_i , \mathbb{H}_n(x) v_j \rangle  \langle \tilde{u}_i  , \mathbb{H}_n(e^{(k)}) \tilde{v}_j \rangle. 
\end{equation*}

The decomposition in~\eqref{eq:PR} can now be interpreted as a composition of an encoder and a decoder,
\begin{equation}
\label{eq:encoder_decoder}
    x \mapsto C = (\langle x , u_i \ast_{\circ} v_j \rangle)_{ i\in[p], j \in [q]} \quad \text{and} \quad \quad C \mapsto \frac{1}{n} \sum_{i=1}^p \sum_{j=1}^q C_{i,j} \tilde{u}_i \ast_{\circ} \tilde{v}_j,
\end{equation}
which relates it to CNNs equipped with an encoder-decoder structure such as U-nets, see Figure~\ref{fig:UNet}. Generalizing this approach to multiple channels, it is possible to stack such encoders and decoders which leads to a layered version of~\eqref{eq:PR}. In~\cite{Ye2018Framelets} it is shown that one can make an informed decision on the number of layers based on the rank of $\mathbb{H}_n(x)$, i.e., the complexity of the input features $x$. Moreover, also an activation function such as the ReLU or bias vectors can be included. 
 The key question one can then ask is how the kernels can be
chosen to obtain sparse coefficients $C$ in~\eqref{eq:encoder_decoder} and a decomposition such as~\eqref{eq:PR}, i.e., perfect reconstruction.
If $U$ and $V$ are chosen as the left and right singular vectors of $\mathbb{H}_n(x)$, one obtains a very sparse, however input-dependent, representation in~\eqref{eq:PR} due to the fact that
\begin{equation*}
    C_{i,j} = \langle x , u_i \ast_{\circ} v_j \rangle =  \langle u_i, \mathbb{H}_n(x)v_j \rangle = 0, \quad i \neq j.
\end{equation*}
Finally, using the framework of deep convolutional framelets, theoretical
reasons for including skip connections can be derived, since they aid to obtain a perfect reconstruction.
\subsection{Batch normalization}\label{subsec:BatchNorm}

\begin{figure}[t]
    \centering
    \includegraphics[height= 0.22\textheight]{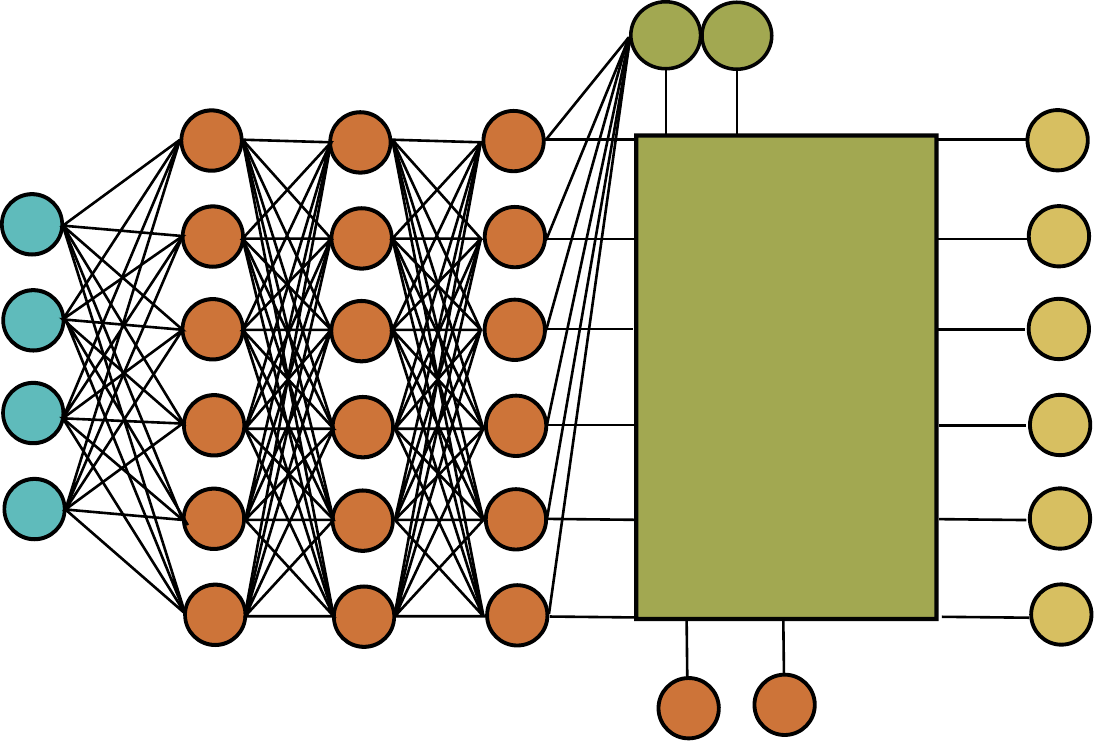}
    \put(-83, 128){\footnotesize $\mu_\batch$}
    \put(-70, 128){\footnotesize$\sigma_\batch$}
    \put(-77, 4){\footnotesize$\beta$}
    \put(-59.5, 4){\footnotesize$\gamma$} 
    \put(-82, 75){$\widehat{y} = \frac{y - \mu_\batch}{ \sigma_\batch}$}
    \put(-82, 55){$z = \gamma \widehat{y} + \beta$}
    \caption{A batch normalization block after a fully connected neural network. The parameters $\mu_\batch, \sigma_\batch$ are the mean and the standard deviation of the output of the fully connected network computed over a batch $\cs$, i.e., a set of inputs. The parameters $\beta,\gamma$ are learnable parts of the batch normalization block.}
\label{fig:batchnorm}
\end{figure}
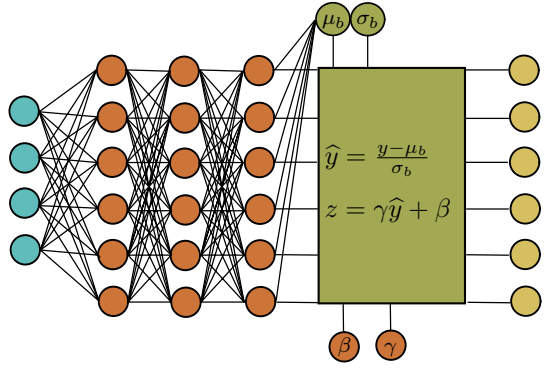

Batch normalization is a building block of NNs that was invented in~\cite{ioffe2015batch} with the goal to reduce so-called \emph{internal covariance shift}. In essence, this phrase describes the (undesirable) situation where during training each layer receives inputs with different distribution. A batch normalization block is defined as follows:
For points $\batch = (y^{(i)})_{i=1}^m \in (\R^n)^m$ and $\beta, \gamma \in \R$, we define 
\begin{equation}
\label{eq:batch_norm_block}
    \mathrm{BN}_{\batch}^{(\beta, \gamma)}(y) \coloneqq \gamma \, \frac{y - \mu_\batch}{\sigma_\batch} + \beta, \quad y\in\R^n, \quad \text{with} \quad \mu_\batch = \frac{1}{m}\sum_{i=1}^m y^{(i)} \quad \text{and} \quad \sigma_\batch^2 = \frac{1}{m}\sum_{i=1}^m (y^{(i)} - \mu_\batch)^2,
\end{equation}
where all operations are to be understood componentwise, see Figure~\ref{fig:batchnorm}. 

Such a batch normalization block can be added into a NN architecture. Then $\batch$ is the output of the previous layer over a batch or the whole training data\footnote{In practice, one typically uses a moving average to estimate the mean $\mu$ and the standard deviation $\sigma$ of the output of the previous layer over the whole training data by only using batches.}. Furthermore, the parameters $\beta, \gamma$ are variable and can be learned during training. Note that, if one sets $\beta = \mu_\batch$ and $\gamma = \sigma_\batch$, then $\mathrm{BN}_{\batch}^{(\beta, \gamma)}(y)= y$ for all $y \in \R^n$. Therefore, a batch normalization block does not negatively affect the expressivity of the architecture. 
On the other hand, batch normalization does have a tangible effect on the optimization aspects of deep learning. Indeed, in~\cite[Theorem 4.1]{santurkar2018does}, the following observation was made:
\begin{proposition}[Smoothening effect of batch normalization]\label{prop:reductionOfDerivative}
Let $m \in \N$ with $m\ge 2$ and for every $\beta,\gamma \in \R$ define $\mathcal{B}^{(\beta,\gamma)} \colon \R^m \to \R^m$ by
\begin{equation}
\label{eq:batch_norm}
   \mathcal{B}^{(\beta,\gamma)}(\batch) = (\mathrm{BN}_{\batch}^{(\beta, \gamma)}(y^{(1)}), \dots, \mathrm{BN}_{\batch}^{(\beta, \gamma)}(y^{(m)})), \quad \batch = (y^{(i)})_{i=1}^m \in \R^m,
\end{equation}
where $\mathrm{BN}_{\batch}^{(\beta,\gamma)}$ is given as in~\eqref{eq:batch_norm_block}.
Let $\beta,\gamma \in \R$ and let $r \colon \R^m \to \R$ be a differentiable function.
Then it holds for every $\batch\in\R^m$ that
\begin{equation*}
    \| \nabla(r\circ \mathcal{B}^{(\beta,\gamma)}) (\batch) \|_2^2 = \frac{\gamma^2}{\sigma_\batch^2} \big( \left\| \nabla r (\batch) \right\|^2 - \frac{1}{m} \langle \mathbf{1}, \nabla r (\batch) \rangle^2    - \frac{1}{m} \langle \mathcal{B}^{(0,1)}(\batch), \nabla r (\batch) \rangle^2\big),
\end{equation*}
where $\mathbf{1}=(1,\dots,1)\in\R^m$ and
$
    \sigma_\batch^2
$ is given as in~\eqref{eq:batch_norm_block}. 
\end{proposition}

For multi-dimensional $y^{(i)}\in\R^n$, $i\in [m]$, the same statement holds for all components as, by definition, the batch normalization block acts componentwise. Proposition~\ref{prop:reductionOfDerivative} follows from a convenient representation of the Jacobian of the mapping $\mathcal{B}^{(\beta,\gamma)}$, given by
\begin{equation*}
    \frac{\partial \mathcal{B}^{(\beta,\gamma)}(\batch)}{\partial \batch} = \frac{\gamma}{\sigma_\batch}\Big(\mathrm{I}_m - \frac{1}{m}\mathbf{1}\mathbf{1}^T - \frac{1}{m} \mathcal{B}^{(0,1)}(\batch)(\mathcal{B}^{(0,1)}(\batch))^T\Big), \quad \batch\in\R^m,
\end{equation*}
and the fact that $\{ \frac{\mathbf{1}}{\sqrt{m}}, \frac{1}{\sqrt{m}} \mathcal{B}^{(0,1)}(\batch)\}$ constitutes an orthonormal set.

Choosing $r$ to mimic the empirical risk of a learning task, Proposition~\ref{prop:reductionOfDerivative} shows that, in certain situations---for instance, if $\gamma$ is smaller than $\sigma_\batch$ or if $m$ is not too large---a batch normalization block can considerably reduce the magnitude of the derivative of the empirical risk with respect to the input of the batch normalization block. By the chain rule, this implies that also the derivative of the empirical risk with respect to NN parameters influencing the input of the batch normalization block is reduced. 

Interestingly, a similar result holds for second derivatives~\cite[Theorem 4.2]{santurkar2018does} if $r$ is twice differentiable. One can conclude that adding a batch normalization block increases the smoothness of the optimization problem. Since the parameters $\beta$ and $\gamma$ were introduced, including a batch normalization block also increases the dimension of the optimization problem by two.

\subsection{Sparse neural networks and pruning}\label{subsec:sparse}
\begin{wrapfigure}{r}{0.34\textwidth}
    \includegraphics[width= \linewidth]{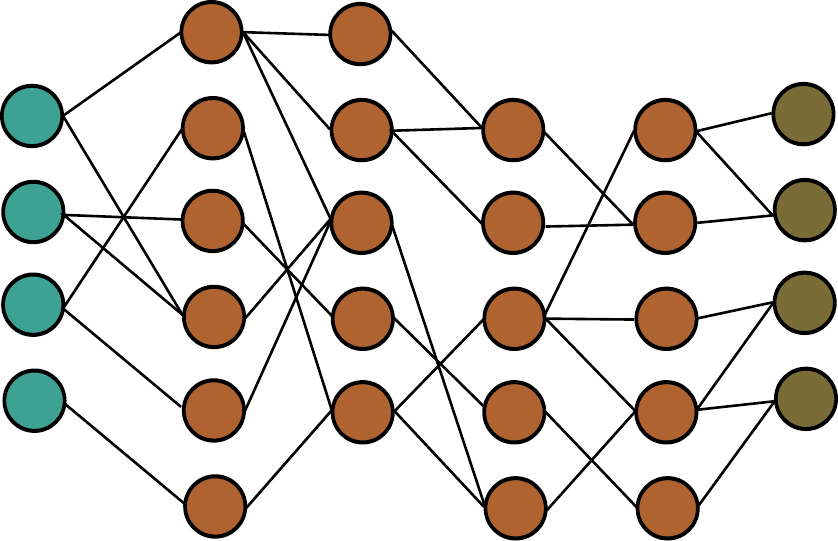}
    \caption{A neural network with sparse connections.}
    \label{fig:sparse}
\end{wrapfigure}

For deep FC NNs, the number of trainable parameters usually scales like the square of the number of neurons.
For reasons of computational complexity and memory efficiency, it appears sensible to seek for techniques to reduce the number of parameters or extract \emph{sparse subnetworks} (see Figure~\ref{fig:sparse}) without affecting the output{
\parfillskip=0pt
\parskip=0pt
\par}
\noindent of a NN much.
One way to do this is by \emph{pruning}~\cite{le1989optimal, HanMD15}.
Here, certain parameters of a NN are removed after training. This is done, for example, by setting these parameters to zero. 

In this context, the \emph{lottery ticket hypothesis} was formulated in~\cite{frankle2018lottery}. It states: \enquote{A randomly-initialized, dense NN contains a subnetwork that is initialized such that---when trained in isolation---it can match the test accuracy of the
original NN after training for at most the same number of iterations}. In~\cite{ramanujan2020s} a similar hypothesis was made and empirically studied. There, it is claimed that, for a sufficiently overparametrized NN, there exists a subnetwork that matches the performance of the large NN after training without being trained itself, i.e., already at initialization.

Under certain simplifying assumptions, the existence of favorable subnetworks is quite easy to prove. We can use a technique that was previously indirectly used in Subsection~\ref{subsec:RandSamp}---the Carathéodory Lemma. This result states the following: Let $n\in\N$, $C\in (0,\infty)$, and let $(\mathcal{H},\|\cdot\|)$ be a Hilbert space. Let $F\subset \mathcal{H}$ with $\sup_{f\in F} \|f\|\le C$ and let $g\in \mathcal{H}$ be in the convex hull of $F$.
Then there exist $f_{i}\in F$, $i\in [n]$, and $c \in [0,1]^n$ with $\|c\|_1 =1$ such that

\begin{equation*}
    \left\| g - \sum_{i=1}^n c_i f_i\right\| \leq \frac{C}{\sqrt{n}},
\end{equation*}
see, e.g.,~\cite[Theorem 0.0.2]{vershynin2018high}. 

\begin{proposition}[Carathéodory pruning]\label{prop:CaratheordoryPruning}
Let $d,n\in\N$, with $n\ge 100$ and let $\mu$ be a probability measure on the unit ball $B_1(0)\subset\R^d$. Let $\arch=((d,n,1),\ReLU)$ be the architecture of a two-layer ReLU network and let $\theta\in \R^{P((d,n,1))}$ be corresponding parameters such that
\begin{equation*}
    \Phi_a(\cdot,\theta) = \sum_{i=1}^n w^{(2)}_i \ReLU( \langle(w_i^{(1)}, \cdot \rangle +b_i^{(1)})),
\end{equation*}
where $(w_i^{(1)}, b_i^{(1)})\in \R^d\times\R$, $i\in [n]$, and $w^{(2)}\in\R^n $. 
Assume that for every $i\in [n]$ it holds that $\|w_i^{(1)}\|_2\le 1/2$ and $b_i^{(1)} \le 1/2$.
Then there exists a parameter $\tilde{\theta}\in \R^{P((d,n,1))}$ with at least $99\%$ of its entries being zero such that
\begin{equation*}
    \|\Phi_a(\cdot,\theta) - \Phi_{a}(\cdot, \tilde{\theta})\|_{L^2(\mu)} \le \frac{15\|w^{(2)}\|_1}{\sqrt{n}}.
\end{equation*}
Specifically, there exists an index set $I\subset [n]$ with $|I|\le n/100$ such that $\tilde{\theta}$ satisfies that
\begin{equation*}
    \widetilde{w}_i^{(2)}= 0, \quad \text{if }i\notin I, \qquad \text{and} \qquad (\widetilde{w}_i^{(1)}, \tilde{b}_i^{(1)})=\begin{cases} (w_i^{(1)},b_i^{(1)}), &\text{if }i\in I, \\ (0,0), &\text{if }i\notin I. \end{cases}
\end{equation*}
\end{proposition}

The result is clear if $w^{(2)} = 0$.
Otherwise, define
\begin{equation}
    f_i \coloneqq  \|w^{(2)}\|_1\ReLU( \langle w_i^{(1)}, \cdot\rangle +  b^{(1)}_i), \quad i \in [n],
\end{equation}
and observe that $\Phi_a(\cdot,\theta)$ is in the convex hull of $\{f_i\}_{i=1}^n \cup \{-f_i\}_{i=1}^n$.
Moreover, by the Cauchy--Schwarz inequality, it holds that 
\begin{equation*}
    \|f_i\|_{L^2(\mu)} \leq \|w^{(2)}\|_1 \|f_i\|_{L^\infty(B_1(0))} \leq \|w^{(2)}\|_1.
\end{equation*}
We conclude with the Carathéodory Lemma that there exists $I\subset [n]$ with $|I|= \lfloor n /100 \rfloor \ge n /200$ and $c_i\in [-1,1]$, $i\in I$, such that
\begin{equation*}
    \left\|\Phi_a(\cdot, \theta) - \sum_{i\in I}c_i f_i \right\|_{L^2(\mu)} \leq \frac{\|w^{(2)}\|_1}{\sqrt{|I|}} \le \frac{\sqrt{200}\|w^{(2)}\|_1}{\sqrt{n}},    
\end{equation*}
which yields the result.

Proposition~\ref{prop:CaratheordoryPruning} shows that certain, very wide NNs can be approximated very well by sparse subnetworks where only the output weight matrix needs to be changed. The argument of Proposition~\ref{prop:CaratheordoryPruning} is inspired by~\cite{barron2018approximation}, where a much more refined result is shown for deep NNs.

\subsection{Recurrent neural networks}\label{subsec:RNNs}
\begin{wrapfigure}{r}{0.34\textwidth}
    \includegraphics[width=\linewidth]{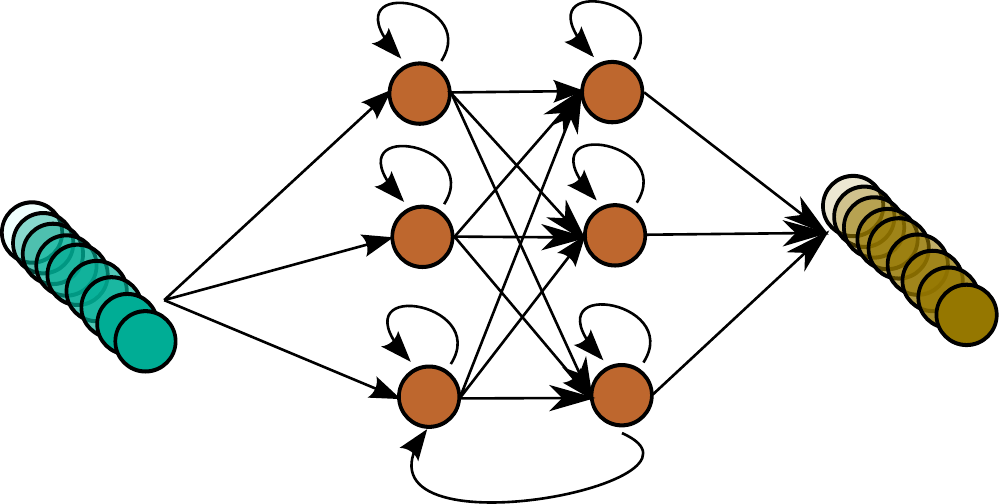}
    \caption{Sketch of a recurrent neural network. Cycles in the computational graph incorporate the sequential structure of the input and output.}
    \label{fig:RNN}
\end{wrapfigure}
Recurrent NNs are NNs where the underlying graph is allowed to exhibit cycles as in Figure~\ref{fig:RNN}, see~\cite{hopfield1982neural, rumelhart1986learning, elman1990finding, jordan1990attractor}. Previously, we had excluded cyclic computational graphs. For a feed-forward NN, the computation of internal states is naturally performed step by step through the layers. Since the output of a layer does not affect previous layers, the order in which the computations of the NN are performed corresponds to the order of the layers. For recurrent NNs, the concept of layers does not exist, and the order of operations is much more delicate. Therefore, one considers time steps. In each time step, all possible computations of the graph are applied to the current state of the NN. This yields a new internal state.
Given that time steps arise naturally from the definition of recurrent NNs, this NN type is typically used for sequential data. 

If the input to a recurrent NN is a sequence, then every input determines the internal state of the recurrent NN for the following inputs. Therefore, one can claim that these NNs exhibit a memory. This fact is extremely desirable in natural language processing, which is why recurrent NNs are widely used in this application.

Recurrent NNs can be trained similarly to regular feed-forward NNs by an algorithm called \emph{backpropagation through time}~\cite{marvin1969perceptrons, werbos1988generalization, williams1995gradient}. This procedure essentially unfolds the recurrent structure yielding a classical NN structure. However, the algorithm may lead to very deep structures. Due to the vanishing and exploding gradient problem discussed earlier, very deep NNs are often hard to train. Because of this, special recurrent structures were introduced that include gates which prohibit too many recurrent steps; these include the widely used LSTMs~\cite{hochreiter1997long}.

The application area of recurrent NNs is typically quite different from that of regular NNs since they are specialized on sequential data. Therefore, it is hard to quantify the effect of a recurrent connection on a fully connected NN. However, it is certainly true that with recurrent connections certain computations can be performed much more efficiently than with feed-forward NN structures. A particularly interesting construction can be found in~\cite[Theorem 4.4]{bohn2019recurrent}, where it is shown that a fixed size, recurrent NN with ReLU activation function, can approximate the function $x \mapsto x^2$ to any desired accuracy. The reason for this efficient representation can be seen when considering the self-referential definition of the approximant to $x-x^2$ shown in Figure~\ref{fig:squaring}.

On the other hand, with feed-forward NNs, it transpires from Theorem~\ref{thm:lowerboundpieces} that the approximation error of fixed-sized ReLU NNs for any non-affine function is greater than a positive lower bound.

\section{Describing the features a deep neural network learns}\label{sec:features}

This section presents two viewpoints which help in understanding the nature of features that can be described by NNs. Section~\ref{subsec:wavelets} summarizes aspects of the so-called \emph{scattering transform} which constitutes a specific NN architecture that can be shown to satisfy desirable properties, such as translation and deformation invariance. Section~\ref{subsec:sparsecoding} relates NN features to the current paradigm of \emph{sparse coding}. 

\subsection{Invariances and the scattering transform}
\label{subsec:wavelets}

One of the first theoretical contributions to the understanding of the mathematical properties of CNNs is~\cite{mallat2012group}. The approach taken in that work is to consider specific CNN architectures with \emph{fixed} parameters that result in a stand-alone feature descriptor whose output may be fed into a subsequent classifier (for example, a kernel support vector machine or a trainable FC NN). 
From an abstract point of view, a feature descriptor is a function $\Psi$ mapping from a signal space, such as $L^2(\mathbb{R}^d)$ or the space of piecewise smooth functions, to a feature space. In an ideal world, such a classifier should \enquote{factor} out invariances that are irrelevant to a subsequent classification problem while preserving all other information of the signal. A very simple example of a classifier which is invariant under translations is the Fourier modulus $\Psi\colon L^2(\mathbb{R}^d) \to L^2(\mathbb{R}^d)$, $u\mapsto |\hat{u}|$. This follows from the fact that a translation of a signal $u$ results in a modulation of its Fourier transform, i.e., $\widehat{u(\cdot - \tau)}(\omega) = e^{-2\pi i \langle \tau, \omega\rangle}\hat{u}(\omega)$, $\tau,\omega\in\R^d$. 
Furthermore, in most cases (for example, if $u$ is a generic compactly supported function~\cite{grohs2020phase}), $u$ can be reconstructed up to a translation from its Fourier modulus~\cite{grohs2020phase} and an energy conservation property of the form $\|\Psi(u)\|_{L^2} = \|u\|_{L^2}$ holds true. 
Translation invariance is, for example, typically exhibited by image classifiers, where the label of an image does not change if it is translated.

In practical problems many more invariances arise. Providing an analogous representation that factors out general invariances would lead to a significant reduction in the problem dimensionality and constitutes an extremely promising route towards dealing with the very high dimensionality that is commonly encountered in practical problems~\cite{mallat2016understanding}. This program is carried out in~\cite{mallat2012group} for additional invariances with respect to deformations $u\mapsto u_\tau:=u(\cdot - \tau(\cdot))$, where $\tau\colon\R^d\to \R^d$ is a smooth mapping. Such transformations may occur in practice, for instance, as image warpings. In particular, a feature descriptor $\Psi$ is designed that, with a suitable norm $\|\cdot \|$ on the image of $\Psi$, 
\begin{enumerate}[label=(\alph*)]
    \item \label{it:inv1} is Lipschitz continuous with respect to deformations in the sense that $\|\Psi(u) - \Psi(u_\tau)\|\lesssim K(\tau,\nabla \tau, \nabla^2\tau)$ holds for some $K$ that only mildly depends on $\tau$ and essentially grows linearly in $\nabla \tau$ and $\nabla^2\tau$,
    \item is almost (i.e., up to a small and controllable error) invariant under translations of the input data, and
    \item \label{it:inv3} contains all relevant information on the input data in the sense that an energy conservation property 
    \begin{equation*}
    \|\Psi(u)\|\approx \| u\|_{L^2} 
    \end{equation*}
    holds true.
\end{enumerate} 

Observe that, while the action of translations only represents a $d$-parameter group, the action of deformations/warpings represents an infinite-dimensional group. Hence, a deformation invariant feature descriptor represents a big potential for dimensionality reduction.  
Roughly speaking, the feature descriptor $\Psi$ of~\cite{mallat2012group} (also coined the \emph{scattering transform}) is defined by collecting features that are computed by iteratively applying a wavelet transform followed by a pointwise modulus non-linearity and a subsequent low-pass filtering step, i.e.,
\begin{equation*}
|||u\ast \psi_{j_1}|\ast \psi_{j_2} \ast \dots |\ast \psi_{j_{\ell}}| \ast \varphi_J, 
\end{equation*}
where $\psi_{j}$ refers to a wavelet at scale $j$ and $\varphi_J$ refers to a scaling function at scale $J$.
The collection of all these so-called \emph{scattering coefficients} can then be shown to satisfy the properties in~\ref{it:inv1}--\ref{it:inv3} above in a suitable (asymptotic) sense. The proof of this result relies on a subtle interplay between a \enquote{deformation covariance} property of the wavelet transform and a \enquote{regularizing} property of the operation of convolution with the modulus of a wavelet.
%
We remark that similar results can be shown also for different systems, such as Gabor frames~\cite{wiatowski2017energy, czaja2019analysis}.

\subsection{Hierarchical sparse representations}
\label{subsec:sparsecoding}

The previous approach modeled the learned features by a specific dictionary, namely wavelets. It is well known that one of the striking properties of wavelets is to provide sparse representations for functions belonging to certain function classes. More generally, we speak of sparse representations with respect to a representation system. For a vector $x \in \mathbb{R}^d$, a sparsifying representation system $D \in \mathbb{R}^{d \times p}$---also called \emph{dictionary}---is such that $x = D\phi$ with the coefficients $\phi \in \mathbb{R}^p$ being sparse in the sense that $\|\phi\|_0 \coloneqq |\supp(\phi)|= |\{i \in [p]\colon \phi_i \neq 0\}|$ is small compared to $p$. A similar definition can be made for signals in infinite-dimensional spaces. Taking sparse representations into account, the theory of sparse coding provides an approach to a theoretical understanding of the features a deep NN learns. 

One common method in image processing is the utilization of not the entire image but overlapping patches of it, coined \emph{patch-based image processing}. Thus of particular interest are local dictionaries which sparsify those patches, but presumably not the global image. This led to the introduction of the \emph{convolutional sparse coding} model (CSC model), which links such local and global behaviors. Let us describe this model for one-dimensional convolutions on the group $G:=\Z / (d\Z)$ with kernels supported on the subgroup $H:=\Z / (n\Z)$, where $d,n\in\N$ with $n <d$, see also Subsection~\ref{subsec:convnets}.
The corresponding CSC model is based on a decomposition of a global signal $x \in (\mathbb{R}^G)^{c}$ with $c\in\N$ channels as 
\begin{equation} \label{eq:CSCmodel}
x_i =  \sum_{j=1}^{C} \kappa_{i,j} \ast \phi_j, \quad i\in [c],
\end{equation}
where $\phi \in (\mathbb{R}^G)^{C}$ is supposed to be a sparse representation with $C\in\N$ channels and $\kappa_{i,j}\in \R^G$, $i\in[c]$, $j\in[C]$, are local kernels with $\supp (\kappa_{i,j}) \subset H$.
Let us consider a patch $((x_i)_{g+h})_{i \in [c], h \in H}$ of $n$ adjacent entries, starting at position $g\in G$, in each channel of $x$. The condition on the support of the kernels $\kappa_{i,j}$ and the representation in~\eqref{eq:CSCmodel} imply that this patch is only affected by a stripe of at most $(2n-1)$ entries in each channel of $\phi$.
The local, patch-based sparsity of the representation $\phi$ can thus be appropriately measured via
\begin{equation*}
    \|\phi\|^{(n)}_{0,\infty} \coloneqq \max_{g\in G} \|((\phi_{j})_{g+k})_{j \in [C], k \in [2n-1]}\|_0,
\end{equation*}
see~\cite{papyan2017working}. Furthermore, note that we can naturally identify $x$ and $\phi$ with vectors in $\R^{d c}$ and $\R^{d C}$ and write
$x = D \phi$, where $D \in \R^{d c \times dC}$ is a matrix consisting of circulant blocks, typically referred to as a \emph{convolutional dictionary}.

The relation between the CSC model and deep NNs is revealed by applying the CSC model in a layer-wise fashion~\cite{papyan2017convolutional,sulam2018multilayer,papyan2018theoretical}.
To see this, let $C_0\in \N$ and for every $\ell\in[L]$ let $C_\ell, k_\ell \in \N$ and let $D^{(\ell)}\in \mathbb{R}^{dC_{\ell-1} \times dC_{\ell}}$ be a convolutional dictionary with kernels supported on $\Z/(n_\ell\Z)$. A signal $x=\phi^{(0)} \in \mathbb{R}^{dC_0}$ is said to belong to the corresponding \emph{multi-layered CSC model} (ML-CSC model) if there exist coefficients $\phi^{(\ell)}\in \R^{dC_{\ell}}$ with  
\begin{equation}
\label{eq:ml_csc}
    \phi^{(\ell-1)} = D^{(\ell)} \phi^{(\ell)} \quad \text{and} \quad  \|\phi^{(\ell)}\|^{(n_\ell)}_{0,\infty} \le k_\ell, \quad \ell\in[L].
\end{equation}
We now consider the problem of reconstructing the sparse coefficients $(\phi^{(\ell)})_{\ell=1}^L$ from a noisy signal $\tilde{x} \coloneqq x + \nu$, where the noise $\nu\in\R^{dC_0}$ is assumed to have small $\ell^2$-norm and $x$ is assumed to follow the ML-CSC model in~\eqref{eq:ml_csc}. In general, this problem is NP-hard. However, under suitable conditions on the ML-CSC model, it can be approximately solved, for instance, by a layered thresholding algorithm. 

More precisely, for $D \in \R^{d c \times dC}$ and $b \in \R^{dC}$, we define a \emph{soft-thresholding operator} by  
\begin{equation}
\label{eq:soft_thresh}
    \mathcal{T}_{D,b}(x) \coloneqq \ReLU(D^Tx - b)-\ReLU(-D^Tx - b), \quad x\in \R^{d c},
\end{equation}
where $\ReLU(x)= \max\{0,x\}$ is applied componentwise. If $x=D\phi$ as in~\eqref{eq:CSCmodel}, we obtain $\phi\approx \mathcal{T}_{D,b}(x)$ roughly under the following conditions: The distance of $\phi$ and $\psi\coloneqq D^Tx=D^TD\phi$ can be bounded using the local sparsity of $\phi$ and the mutual coherence and locality of the kernels of the convolutional dictionary $D$. For a suitable threshold $b$, the mapping $\psi \mapsto \ReLU(\psi - b)-\ReLU(-\psi - b)$ further recovers the support of $\phi$ by nullifying entries of $\psi$ with $\psi_i \le |b_i|$. Utilizing the soft-thresholding operator~\eqref{eq:soft_thresh} iteratively for corresponding vectors $b^{(\ell)} \in \R^{d C_\ell}$, $\ell \in [L]$, then suggests the following approximations:
\begin{equation*}
    \phi^{(\ell)} \approx (\mathcal{T}_{D^{(\ell)},b^{(\ell)}} \circ \dots \circ \mathcal{T}_{D^{(1)},b^{(1)}})(\tilde{x}), \quad \ell \in [L].
\end{equation*}
The resemblance with the realization of a CNN with ReLU activation function is evident. The transposed dictionary $(D^{(\ell)})^T$ can be regarded as modeling the learned convolutional kernels, the threshold $b^{(\ell)}$ models the bias vector, and the soft-thresholding operator $\mathcal{T}_{D^{(\ell)},b^{(\ell)}}$ mimics the application of a convolutional block with a ReLU non-linearity in the $\ell$-th layer. 

Using this model, a theoretical understanding of CNNs from the perspective of sparse coding is now at hand. This novel perspective gives a precise mathematical meaning of the kernels in a CNN as sparsifying dictionaries of an ML-CSC model. Moreover, the forward pass of a CNN can be understood as a layered thresholding algorithm for decomposing a noisy signal $\tilde{x}$.
The results derived are then of the following flavor: Given a suitable reconstruction procedure such as thresholding or $\ell_1$-minimization, the sparse coefficients $(\phi^{(\ell)})_{\ell=1}^L$ of a signal $x$ following a ML-CSC model can be stably recovered from the noisy signal $\tilde{x}$ under certain hypotheses on the ingredients of the ML-CSC model.

\section{Effectiveness in natural sciences}
\label{sec:effectiveness}

The theoretical insights of the previous sections do not always accurately describe the performance of NNs in applications. Indeed, there often exists a considerable gap between the predictions of approximation theory and the practical performance of NNs~\cite{adcock2020gap}. 

In this section, we consider concrete applications which have been very successfully solved with deep-learning-based methods. In Section~\ref{subsec:inverseproblems} we present an overview of deep-learning-based algorithms applied to inverse problems. Section~\ref{subsect:pdebasedmodels} then continues by describing how NNs can be used as a numerical ansatz for solving PDEs, highlighting their use in the solution of the multi-electron Schr\"odinger equation. 
\subsection{Deep neural networks meet inverse problems}
\label{subsec:inverseproblems}

The area of inverse problems, predominantly in imaging, was presumably the first class of mathematical methods
embracing deep learning with overwhelming success. Let us consider a forward operator $K\colon \cY \to \cX$ with $\cX, \cY$ being Hilbert spaces and the associated 
inverse problem of finding $y\in \cY$ such that $Ky=x$ for given features $x\in \cX$. The classical model-based approach to regularization aims to approximate $K$ by invertible operators, and is hence strongly based on functional analytic principles. Today, such approaches 
take well-posedness of the approximation, convergence properties, as well as the structure of regularized solutions
into account. The last item allows to incorporate prior information of the original solution such as regularity, 
sharpness of edges, or---in the case of sparse regularization~\cite{jin2017sparsity}---a sparse coefficient 
sequence with respect to a prescribed representation system. Such approaches are typically realized in a 
variational setting and hence aim to minimize functionals of the form
\begin{equation} \label{eq:IPsolver}
    \Vert Ky - x\Vert^2 + \alpha R(y), 
\end{equation}
where $\alpha\in(0,\infty)$ is a regularization parameter, $R\colon \cY\to [0,\infty)$ a regularization term, and $\|\cdot\|$ denotes the norm on $\cY$. As said, the regularization term aims to 
model structural information about the desired solution. However, one main hurdle in this approach is the problem that
typically solution classes such as images from computed tomography cannot be modeled accurately enough to, for instance,
allow reconstruction under the constraint of a significant amount of missing features. 

This has opened the door to data-driven approaches, and recently, deep NNs. Solvers of inverse problems 
which are based on deep learning techniques can be roughly categorized into three classes:
\begin{enumerate}
    \item \textit{Supervised approaches:} The most straightforward approach is to train a NN $\Phi(\cdot,\theta)\colon \cX\to \cY$ end-to-end, i.e., to completely learn the map from data $x$ to the solution $y$. More advanced approaches in this direction incorporate information about the operator $K$ into the NN such as in~\cite{adler2017solving,gilton2019neumann,monga2021algorithm}.  
    Yet another type of approaches aims to combine deep NNs with classical model-based approaches. The first suggestion in this realm was to start by applying a standard solver, followed by a deep NN $\Phi(\cdot,\theta)\colon\cY\to\cY$ which serves as a denoiser for specific reconstruction artifacts, e.g.,~\cite{jin2017deep}. This was followed by more sophisticated methods such as plug-and-play frameworks for coupling inversion and denoising~\cite{romano2017little}.
    \item \textit{Semi-supervised approaches:} These type of approaches aim to encode the regularization by a deep NN $\Phi(\cdot,\theta)\colon \cY \to [0,\infty)$. The underlying idea is often to require stronger regularization on solutions $y^{(i)}$ that are more prone to artifacts or other effects of the instability of the problem. On solutions where typically few artifacts are observed less regularization can be used. Therefore, the learning algorithm only requires a set of labels $(y^{(i)})_{i=1}^m$ as well as a method to assess how hard the inverse problem for this label would be. In this sense, the algorithm can be considered semi-supervised. This idea was followed, for example, in~\cite{lunz2018adversarial,li2020nett}. Taking a Bayesian viewpoint, one can also learn prior distributions as deep NNs, which was done in~\cite{barbano2020quantifying}. 
    
    \item \textit{Unsupervised approaches:} One highlight of what we might coin unsupervised approaches in our problem setting is the introduction of deep image priors in~\cite{dittmer2020regularization,ulyanov2018deep}. The key idea is to parametrize the solutions $y$ as the output of a NN $\Phi(\xi,\cdot)\colon \cP \to \cY$ with parameters in a suitable space $\cP$, applied to a fixed input $\xi$. Then, for given features $x$, one tries to solve $\min_{\theta\in \cP}\Vert K \Phi(\xi,\theta) - x \Vert^2$ in order to obtain parameters $\hat{\theta}\in\cP$ that yield a solution candidate $y=\Phi(\xi,\hat{\theta})$. Here often early stopping is applied in the training of the network parameters.  
  \end{enumerate}

As can be seen, one key conceptual question is how to \enquote{take the best out of both worlds}, in the sense of optimally
combining classical (model-based) methods---in particular the forward operator $K$---with deep learning. This is 
certainly sensitively linked to all characteristics of the particular application at hand, such as availability and 
accuracy of training data, properties of the forward operator, or requirements for the solution. And each of the three  
classes of hybrid solvers follows a different strategy. 

Let us now discuss advantages and disadvantages of methods from the three categories with a particular focus on a 
mathematical foundation. \emph{Supervised} approaches suffer on the one hand from the problem that often ground-truth data
is not available or only in a very distorted form, leading to the fact that synthetic data constitutes a significant
part of the training data. Thus the learned NN will mainly perform as well as the algorithm which generated the 
data, but not significantly improve it---only from an efficiency viewpoint. On the other hand, the inversion is
often highly ill-posed, i.e., the inversion map has a large Lipschitz constant, which negatively affects the 
generalization ability of the NN. Improved approaches incorporate knowledge about the forward operator $K$ 
as discussed, which helps to circumvent this issue. 

One significant advantage of \emph{semi-supervised} approaches is that the underlying mathematical model of the inverse problem is merely augmented by the neural network-based regularization. Assuming that the learned regularizer satisfies natural assumptions, convergence proofs or stability estimates for the resulting regularized methods are still available. 

Finally, \emph{unsupervised} approaches have the advantage that the regularization is then fully due to the specific 
architecture of the deep NN. This makes these methods slightly easier to understand theoretically,
although, for instance, the deep prior approach in its full generality is still lacking a profound mathematical
analysis.

\subsection{PDE-based models}
\label{subsect:pdebasedmodels}
Besides applications in image processing and artificial intelligence, deep learning methods have recently strongly impacted the field of numerical analysis. In particular, regarding the numerical solution of high-dimensional PDEs. These PDEs are widely used as a model for complex processes and their numerical solution presents one of the biggest challenges in scientific computing. We mention three exemplary problem classes:
\begin{enumerate}
    \item \textit{Black--Scholes model:} The Nobel award-winning theory of Fischer Black, Robert Merton, and Myron Scholes proposes a linear PDE model for the determination of a fair price of a (complex) financial derivative. The dimensionality of the model corresponds to the number of financial assets which is typically quite large. The classical linear model, which can be solved efficiently via Monte Carlo methods is quite limited. In order to take into account more realistic phenomena such as default risk, the PDE that models a fair price becomes nonlinear, and much more challenging to solve. In particular (with the notable exception of Multilevel Picard algorithms~\cite{weinan2019multilevel}) no general algorithm exists that provably scales well with the dimension. 

 \item \textit{Schrödinger equation:} The electronic Schrödinger equation describes the stationary nonrelativistic behavior of a quantum mechanical electron system in the electric field generated by the nuclei of a molecule. Its numerical solution is required to obtain stable molecular configurations, compute vibrational spectra, or obtain forces governing molecular dynamics. If the number of electrons is large, this is again a high-dimensional problem and to date there exist no satisfactory algorithms for its solution: It is well known that different gold standard methods may produce completely different energy predictions, for example, when applied to large delocalized molecules, rendering these methods useless for those problems.

\item \textit{Hamilton--Jacobi--Bellman equation:} The Hamilton--Jacobi--Bellman (HJB) equation models the value function of (deterministic or stochastic) optimal control problems. The underlying dimensionality of the model corresponds to the dimension of the space of states to be controlled and tends to be rather high in realistic applications. The high dimensionality, together with the fact that HJB equations typically tend to be fully nonlinear with non-smooth solutions, renders the numerical solution of HJB equations extremely challenging and no general algorithms exist for this problem.
\end{enumerate}
Due to the favorable approximation results of NNs for high-dimensional functions (see especially Subsection~\ref{subsec:PDEapprox}), it might not come as a surprise that a NN ansatz has proven to be quite successful in solving the aforementioned PDE models. A pioneering work in this direction is~\cite{han2018solving} which uses the backwards SDE reformulation of semilinear parabolic PDEs to reformulate the evaluation of such a PDE at a specific point as an optimization problem that can be solved by the deep learning paradigm. The resulting algorithm proves quite successful in the high-dimensional regime and, for instance, enables the efficient modeling of complex financial derivatives including nonlinear effects such as default risk. Another approach specifically tailored to the numerical solution of HJB equations is~\cite{nakamura2021adaptive}. In this work, one uses the Pontryagin principle to generate samples of the PDE solution along solutions of the corresponding boundary value problem. Other numerical approaches include the \emph{Deep Ritz Method}~\cite{weinan2018deep}, where a Dirichlet energy is minimized over a set of NNs, or so-called \emph{Physics Informed Neural Networks}~\cite{raissi2019physics}, where typically the PDE residual is minimized along with some natural constraints, for instance, to enforce boundary conditions. 

Deep-learning-based methods arguably work best if they are combined with domain knowledge to inspire NN architecture choices. We would like to illustrate this interplay at the hand of a specific and extremely relevant example:
the electronic Schrödinger equation (under the Born--Oppenheimer approximation) which amounts to finding the smallest nonzero eigenvalue of the eigenvalue problem
\begin{equation}\label{eq:schro}
    \mathcal{H}_R \psi = \lambda_{\psi}\psi,\hspace{1
			cm} 
\end{equation}
for $\psi\colon \mathbb{R}^{3\times n} \to \mathbb{R}$, where the Hamiltonian
\begin{equation*}
		(\mathcal{H}_R \psi)(r) = -\sum_{i=1}^n\frac{1}{2}(\Delta_{r_i}\psi)(r)
		 - \left(\sum_{i=1}^n\sum_{j=1}^p\frac{Z_j}{\|{r}_i - {R}_j\|_2}-\sum_{i=1}^{p-1}\sum_{j=i+1}^p\frac{Z_iZ_j}{\|{R}_i - {R}_j\|_2} - \sum_{i=1}^{n-1}\sum_{j=i+1}^n\frac{1}{\|{r}_i - {r}_j\|_2}\right)\psi(r)
\end{equation*}   
describes the kinetic energy (first term) as well as Coulomb attraction force between electrons and nuclei (second and third term) and the Coulomb repulsion force between different electrons (third term).
Here, the coordinates $R=\begin{bmatrix} R_1 \dots R_p\end{bmatrix}\in\R^{3\times p}$ refer to the positions of the nuclei, $(Z_i)_{i=1}^p\in\N^p$ denote the atomic numbers of the nuclei, and the coordinates $r=\begin{bmatrix} r_1,\dots, r_n \end{bmatrix}\in \R^{3\times n}$ refer to the positions of the electrons. The associated eigenfunction $\psi$ describes the so-called \emph{wavefunction} which can be interpreted in the sense that $|\psi(r)|^2/\|\psi\|_{L^2}^2$ describes the joint probability density of the $n$ electrons to be located at $r$.
The smallest solution $\lambda_\psi$ of~\eqref{eq:schro} describes the \emph{ground state energy} associated with the nuclear coordinates $R$. It is of particular interest to know the ground state energy for all nuclear coordinates, the so-called \emph{potential energy surface} whose gradient determines the forces governing the dynamic motions of the nuclei. The numerical solution of~\eqref{eq:schro} is complicated by the \emph{Pauli principle} which states that the wave function $\psi$ must be antisymmetric in all coordinates representing electrons of equal spin. To state it, we need to clarify that every electron is not only defined by its location but also by its spin which may be positive or negative. Depending on whether two electrons have the same spin or not, their interaction changes massively. This is reflected by the Pauli principle that we already mentioned: Suppose that electrons $i$ and $j$ have equal spin, then the wave function must satisfy
\begin{equation}\label{eq:pauli}
    P_{i,j}\psi = - \psi, 
\end{equation}
where $P_{i,j}$ denotes the operator that swaps $r_i$ and $r_j$, i.e., $(P_{i,j}\psi)(r) = \psi(r_1,\dots , r_j , \dots , r_i, \dots , r_n)$.
In particular, no two electrons with the same spin can occupy the same location. The challenges associated with solving the Schr\"odinger equation inspired the following famous quote by Paul Dirac~\cite{dirac1929quantum}:
\begin{quote}
    \enquote{The fundamental laws necessary for the mathematical treatment of a large part of physics and the whole of chemistry are thus completely known, and the difficulty lies only in the fact that application of these laws leads to equations that are too complex to be solved.}
\end{quote}

We now describe how deep learning methods might help to mitigate this claim to a certain extent.
Let $X$ be a random variable with density $|\psi(r)|^2/\|\psi\|_{L^2}^2$. Using the Rayleigh--Ritz principle, finding the minimal nonzero eigenvalue of~\eqref{eq:schro} can be reformulated as minimizing the Rayleigh quotient \begin{equation}\label{eq:raleigh}\frac{\int_{\mathbb{R}^{3\times n}}\overline{\psi(r)}(\mathcal{H}_R\psi) (r)\, \mathrm{d}r}{\|\psi\|_{L^2}^2}=\E\left[\frac{(\mathcal{H}_R\psi) (X)}{\psi(X)}\right]\end{equation}
over all $\psi$'s satisfying the Pauli principle, see~\cite{szabo2012modern}. Since this represents a minimization problem it can in principle be solved via a NN ansatz 
by generating training data distributed according to $X$
using MCMC sampling\footnote{Observe that for such sampling methods one can just use the unnormalized density $|\psi(r)|^2$ and thus avoid the computation of the normalization $\|\psi\|_{L^2}^2$.}. Since the wave function $\psi$ will be parametrized as a NN, the minimization of~\eqref{eq:raleigh} will require the computation of the gradient of~\eqref{eq:raleigh} with respect to the NN parameters (the method in~\cite{pfau2020ab} even requires second order derivatives) which, at first sight, might seem to require the computation of third order derivatives. However, due to the Hermitian structure of the Hamiltonian one does not need to compute the derivative of the Laplacian of $\psi$, see, for example,~\cite[Equation (8)]{hermann2020deep}.

Compared to the other PDE problems we have discussed, an additional complication arises from the need to incorporate structural properties and invariances such as the Pauli principle. Furthermore, empirical evidence shows that it is also necessary to hard code the so-called \emph{cusp conditions} which describe the asymptotic behavior of nearby electrons and electrons close to a nucleus into the NN architecture. A first attempt in this direction has been made in~\cite{han2019solving} and significantly improved NN architectures have been developed in~\cite{hermann2020deep,pfau2020ab,scherbela2021} opening the possibility of accurate ab initio computations for previously intractable molecules. The mathematical properties of this exciting line of work remain largely unexplored. We briefly describe the main ideas behind the NN architecture of~\cite{hermann2020deep,scherbela2021}. Standard numerical approaches (notably the Multireference Hartree Fock Method, see~\cite{szabo2012modern}) use a low rank approach to minimize~\eqref{eq:raleigh}. Such a low rank approach would approximate $\psi$ by sums of products of \emph{one electron orbitals} $\prod_{i=1}^n\varphi_i(r_i)$ but clearly this does not satisfy the Pauli principle~\eqref{eq:pauli}. In order to ensure the Pauli principle, one constructs so-called 
\emph{Slater determinants} from one electron orbitals with equal spin. More precisely, suppose that the first $n_+$ electrons with coordinates $r_1,\dots, r_{n_+}$ have positive spin and the last $n-n_+$ electrons have negative spin. Then any function of the form
\begin{equation}\label{eq:slater}
 \det\left(\left(\varphi_i(r_j)\right)_{i,j=1}^{n_+}\right)\cdot \det\left(\left(\varphi_i(r_j)\right)_{i,j=n_+ + 1}^{n}\right)
\end{equation}
satisfies~\eqref{eq:pauli} and is typically called a Slater determinant. While the Pauli principle establishes an (non-classical) interaction between electrons of equal spin, the so-called \emph{exchange correlation}, electrons with opposite spins are uncorrelated in the representation~\eqref{eq:slater}. In particular,~\eqref{eq:slater} ignores interactions between electrons that arise through Coulomb forces, implying that no nontrivial wavefunction can be accurately represented by a single Slater determinant. To capture physical interactions between different electrons, one needs to use sums of Slater determinants as an ansatz. However, it turns out that the number of such determinants that are needed to guarantee a given accuracy scales very badly with the system size $n$ (to the best of our knowledge the best currently known approximation results are contained in~\cite{yserentant2010regularity}, where an $n$-independent error rate is shown, however the implicit constant in this rate depends at least exponentially on the system size $n$).

%

We would like to highlight the approach of~\cite{hermann2020deep} whose main idea is to use NNs to incorporate interactions into Slater determinants of the form~\eqref{eq:slater} using what is called the \emph{backflow trick}~\cite{rios2006inhomogeneous}.
The basic building blocks would now consist of functions of the form 
\begin{equation}\label{eq:backflow}
 \det\left(\left(\varphi_i(r_j)\Psi_{j}(r,\theta_j)\right)_{i,j=1}^{n_+}\right)\cdot \det\left(\left(\varphi_i(r_j)\Psi_{j}(r,\theta_j)\right)_{i,j=n_+ + 1}^{n}\right),
\end{equation}
where $\Psi_{k}(\cdot,\theta_k)$, $k\in [n]$, are NNs. 
If these are arbitrary NNs, it is easy to see that the Pauli principle~\eqref{eq:pauli} will not be satisfied. However, if we require the NNs to be symmetric, for example, in the sense that for $i,j,s\in [n_+]$ it holds that
\begin{equation}\label{eq:backsym}
P_{i,j}\Psi_{k}(\cdot,\theta_k) = \begin{cases} \Psi_{k}(\cdot,\theta_k),&\text{if } k\notin \{i,j\}, \\
                                            \Psi_{i}(\cdot,\theta_i),&\text{if }k=j, \\
                                            \Psi_{j}(\cdot,\theta_j),&\text{if }k=i,
                                            \end{cases}
\end{equation}
and analogous conditions hold for $i,j,k\in [n] \setminus [n_+]$,
the expression~\eqref{eq:backflow} does actually satisfy~\eqref{eq:pauli}. The construction of such symmetric NNs can be achieved by using a modification of the so-called \emph{SchNet Architecture}~\cite{schutt2017schnet} which can be considered as a specific residual NN. 

We describe a simplified construction which is inspired by~\cite{han2019solving} and used in a slightly more complex form in~\cite{scherbela2021}. We restrict ourselves to the case of positive spin (e.g., the first $n_+$ coordinates), the case of negative spin being handled in the same way. Let $\Upsilon(\cdot , \theta^+_{\mathrm{emb}})$ be a univariate NN (with possibly multivariate output) and denote
\begin{equation*}
 \mathrm{Emb}_k(r,\theta^+_{\mathrm{emb}})\coloneqq \sum_{i=1}^{n_+}\Upsilon(\|r_k - r_i\|_2,\theta_{\mathrm{emb}}^+), \quad k\in[n_+],
\end{equation*}  
the $k$-th \emph{embedding layer}. For $k\in[n_+]$, we can now define 
\begin{equation*}
\Psi_k\left(r,\theta_k \right) = \Psi_k\left(r,(\theta_{k,\mathrm{fc}},\theta^+_{\mathrm{emb}})\right) = \Gamma_k\left(\left(\mathrm{Emb}_k(r,\theta^+_{\mathrm{emb}}), (r_{n_++1},\dots , r_n)\right), \theta_{k,\mathrm{fc}} \right),
\end{equation*}
where $\Gamma_k(\cdot , \theta_{k,\mathrm{fc}})$ denotes a standard FC NN with input dimension equal to the output dimension of $\Psi^+$ plus the dimension of negative spin electrons. The networks $\Psi_k$, $k\in[n]\setminus [n_+]$, are defined analogously using different parameters $\theta_{\textrm{emb}}^-$ for the embeddings. It is straightforward to check that the NNs $\Psi_k$, $k\in[n]$, satisfy~\eqref{eq:backsym} so that the backflow determinants~\eqref{eq:backflow} satisfy the Pauli principle~\eqref{eq:pauli}.

In~\cite{hermann2020deep} the backflow determinants~\eqref{eq:backflow} are further augmented by a multiplicative correction term, the so-called \emph{Jastrow factor} which is also represented by a specific symmetric NN, as well as a correction term that ensures the validity of the cusp conditions. The results of~\cite{hermann2020deep} show that this ansatz (namely using linear combinations of backflow determinants~\eqref{eq:backflow} instead of plain Slater determinants~\eqref{eq:slater}) is vastly more efficient in terms of number of determinants needed to obtain chemical accuracy. The full architecture provides a general purpose NN architecture to represent complicated wave functions. A distinct advantage of this approach is that some parameters (for example, embedding layers) may be shared across different nuclear geometries $R\in\R^{3 \times p}$ which allows for the efficient computation of potential energy surfaces~\cite{scherbela2021}, see Figure~\ref{fig:DeepErwin}.
\ifbook
\else%
\begin{figure}[t]
    \centering
    \includegraphics[width = .4\textwidth]{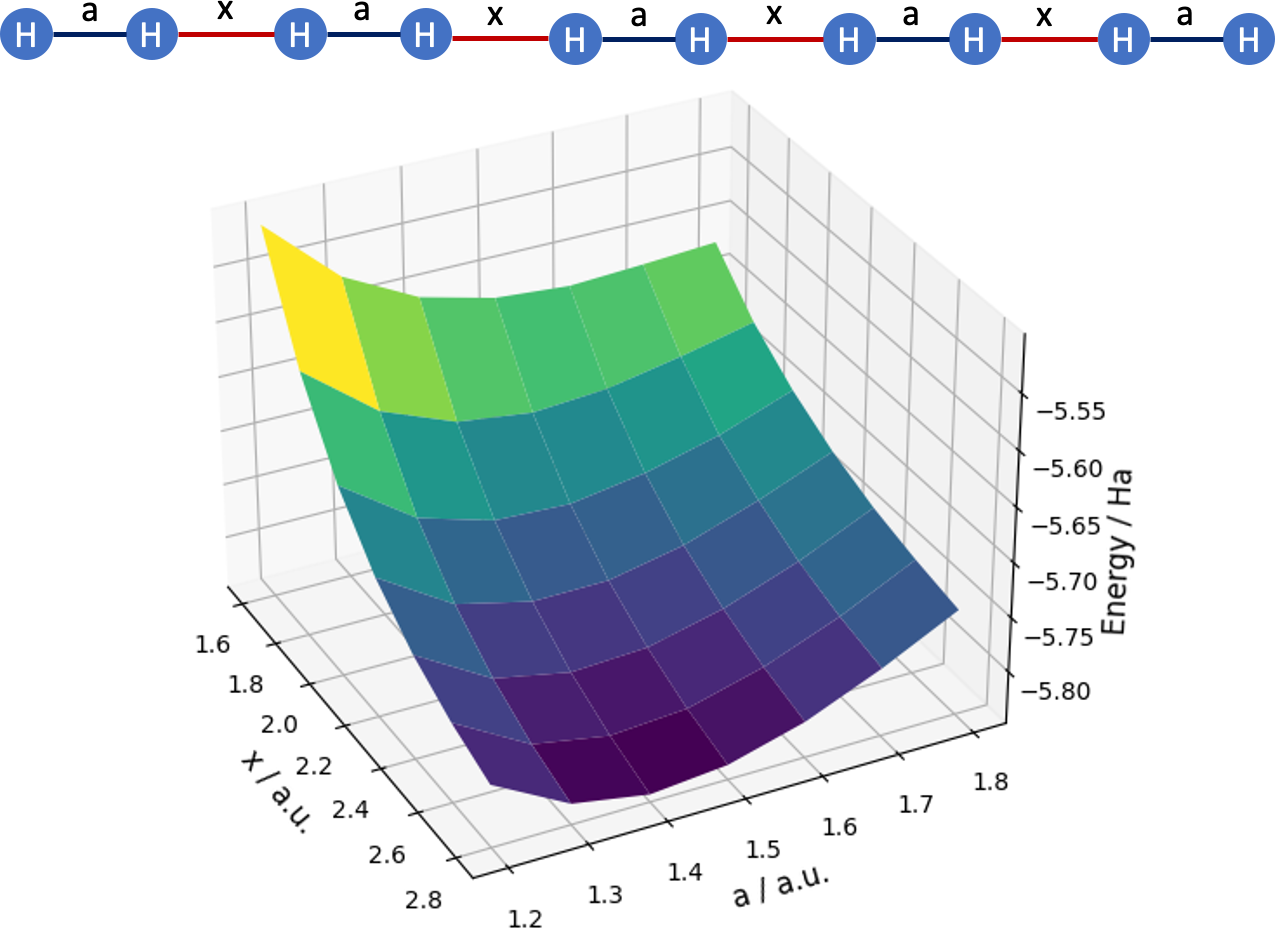}
    \includegraphics[width = .5\textwidth]{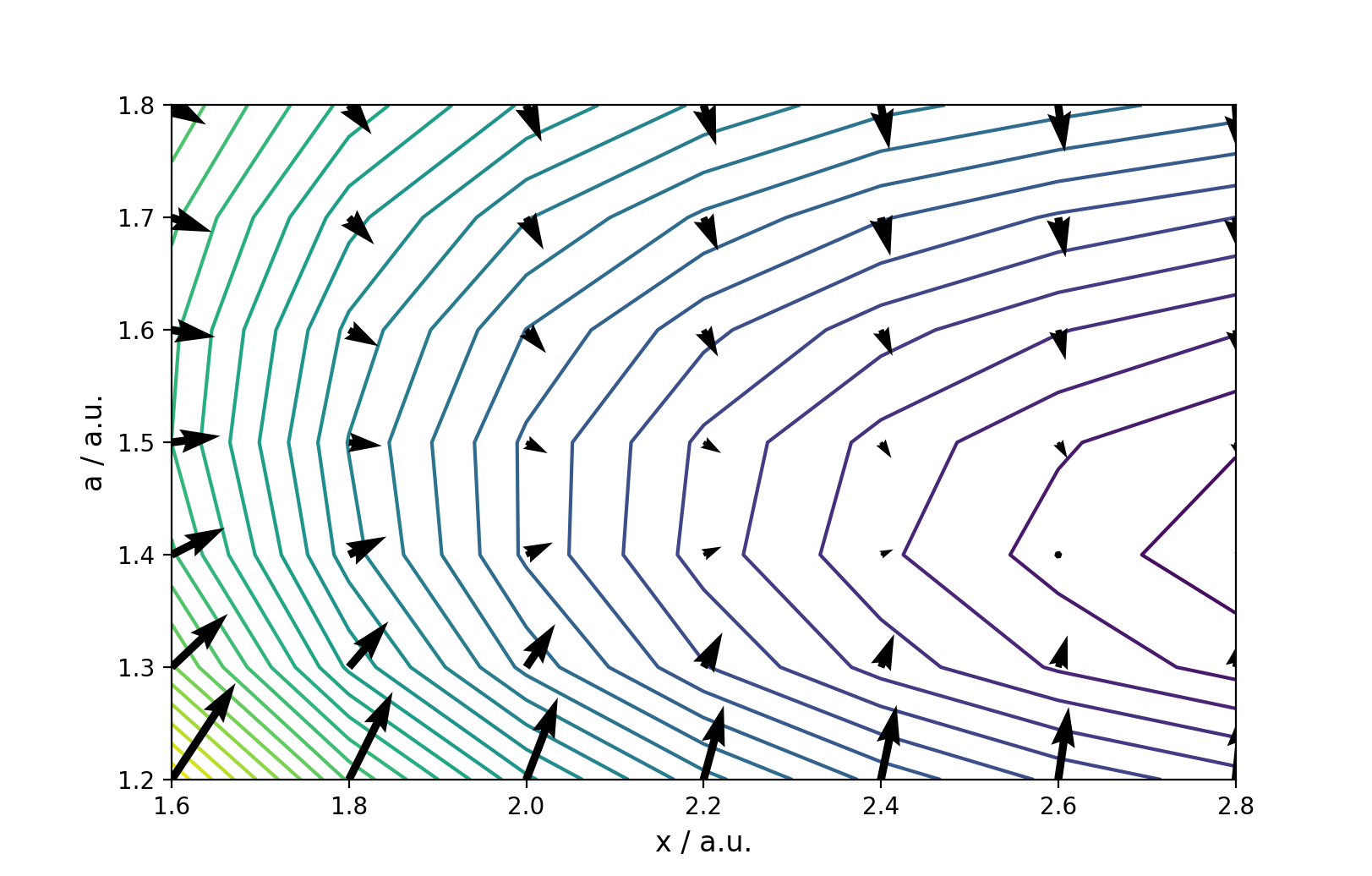}
    \caption{By sharing layers across different nuclear geometries one can efficiently compute different geometries in one single training step~\cite{scherbela2021}. Left: Potential energy surface of H10 chain computed by the deep-learning-based algorithm from~\cite{scherbela2021}. The lowest energy is achieved when pairs of H atoms enter into a covalent bond to form five H2 molecules. Right: The method of~\cite{scherbela2021} is capable of accurately computing forces between nuclei which allows for molecular dynamics simulations from first principles.}
    \label{fig:DeepErwin}
\end{figure}
\fi%
Finally, we would like to highlight the customized NN design that incorporates physical invariances, domain knowledge (for example, in the form of cusp conditions), and existing numerical methods, all of which are required for the method to reach its full potential.
\ifbook
\else%
\section*{Acknowledgment}
The research of JB was supported by the Austrian Science Fund (FWF) under grant I3403-N32. GK acknowledges support from DFG-SPP 1798 Grants KU 1446/21-2 and KU 1446/27-2, DFG-SFB/TR 109 Grant C09, BMBF Grant MaGriDo, and NSF-Simons Foundation Grant SIMONS 81420. The authors would like to thank Héctor Andrade Loarca, Dennis Elbrächter, Adalbert Fono, Pavol Harar, Lukas Liehr, Duc Anh Nguyen, Mariia Seleznova, and Frieder Simon for their helpful feedback on an early version of this article. In particular, Dennis Elbrächter was providing help for several theoretical results.
\fi%

\bibliography{reference}

\end{document}